\documentclass[11pt]{article}

\usepackage[preprint]{acl}

\usepackage{times}
\usepackage{latexsym}

\usepackage[T1]{fontenc}

\usepackage[utf8]{inputenc}

\usepackage{microtype}

\usepackage{inconsolata}

\usepackage{graphicx}
\usepackage{xcolor}
\usepackage{amsmath}
\usepackage{amssymb}
\usepackage{titletoc}        
\usepackage{subcaption}      
\usepackage{enumitem}
\usepackage{multirow}
\usepackage{xspace}
\usepackage{listings}        
\usepackage{booktabs}        
\usepackage{array}           
\usepackage{pifont}
\usepackage{twemojis}

\newcommand{\gptxl}{GPT-2 XL\xspace}
\newcommand{\llamaseven}{Llama-2-7B-chat\xspace}
\newcommand{\llamathirteen}{Llama-2-13B-chat\xspace}
\newcommand{\pythia}[1]{Pythia-#1\xspace}
\newcommand{\qwen}[1]{Qwen3.5-#1\xspace}

\newcommand{\method}[1]{\textsc{#1}\xspace}
\newcommand{\Param}{\method{Param}}
\newcommand{\ParamMask}{\method{Param-mask}}
\newcommand{\ParamMaskHard}{\method{Param-mask-hard}}
\newcommand{\HeadMask}{\method{Head-mask}}
\newcommand{\HeadMaskHard}{\method{Head-mask-hard}}
\newcommand{\HeadCEM}{\method{Head-CEM}}
\newcommand{\HeadGrown}{\method{CLAMP}}
\newcommand{\HeadGrownCos}{\method{Head-grown-cos}}
\newcommand{\ParamC}{\method{Param-C}}
\newcommand{\ourmethod}{\method{CLAMP}}

\newcommand{\paph}{\ensuremath{P(\mathrm{aphasia})}\xspace}
\newcommand{\pbroca}{\ensuremath{P(\mathrm{Broca})}\xspace}
\newcommand{\pwern}{\ensuremath{P(\mathrm{Wernicke})}\xspace}
\newcommand{\pscr}{\ensuremath{P(\mathrm{scrambles})}\xspace}

\newcommand{\vvec}{\ensuremath{\mathbf{v}}\xspace}

\newcommand{\say}[1]{``\textit{#1}''}

\title{Language Impairment as a Lens onto Large Language Models}



\author{
  \textbf{Muxuan He}$^{1\dagger}$ \quad
  \textbf{Wang Bill Zhu}$^{2\dagger}$ \quad
  \textbf{Yue Chen}$^{1\dagger}$ \quad
  \textbf{Elsi Kaiser}$^{1}$ \\
  \textbf{Weizhi Du}$^{3}$ \quad
  \textbf{Yao Du}$^{1}$ \quad
  \textbf{Samuel Nastase}$^{1}$ \quad
  \textbf{Yanjun Gao}$^{4}$ \\
  $^{1}$University of Southern California \quad
  $^{2}$University of California, Berkeley \\
  $^{3}$Independent Researcher \quad
  $^{4}$University of Colorado Anschutz Medical Campus \\[0.3em]
  $^{\dagger}$Corresponding authors: \texttt{\{muxuanhe, ychen249\}@usc.edu}, \texttt{wangzhu@berkeley.edu} \\[0.3em]
  \texttwemoji{globe with meridians}~\href{https://llm-aphasia.github.io/}{\textbf{Project}} \quad
  \raisebox{-0.1em}{\includegraphics[width=1em]{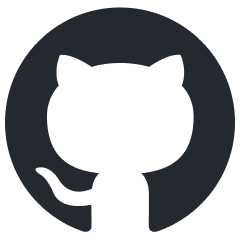}}~\href{https://github.com/Bill1235813/llm_aphasia}{\textbf{Code}}
}

\begin{document}
\maketitle

\begin{abstract}
Although large language models (LLMs) encode language in highly distributed spaces, they remain vulnerable to focal disruptions---a phenomenon strikingly similar to how localized brain damage causes clinical aphasia in humans. To investigate this theoretical tension, we ask whether LLMs possess a functional structure analogous to the human language system, by introducing a novel mechanistic interpretability framework inspired by aphasia. Through two complementary approaches, classifier-guided lesioning and continuous activation steering, we systematically induce aphasia-like breakdowns in LLMs. We find that language functions can be successfully isolated to discrete structural circuits and continuous representational vectors, yet these functions remain densely entangled with significant circuit overlap and vector similarities across distinct aphasic profiles. Furthermore, while steering generalizes robustly across downstream tasks, structural lesioning remains asymmetrically constrained. By bridging neuropsychology and mechanistic interpretability, we demonstrate that LLMs, functionally analogous to biological brains, rely on a densely entangled architecture that can be systematically mapped through clinically-informed interventions.

\end{abstract}

\section{Introduction}
\label{sec:intro}
Large language models (LLMs) have achieved strikingly human-like language behavior, yet their internal functional organization remains fundamentally opaque. Unlike traditional linguistic theories that compartmentalize language into distinct symbolic modules \citep{Chomsky:57}, LLMs encode diverse linguistic structures into a unified, high-dimensional representation space \citep{almeida-etal2026, baroni-2020, tenney-etal2019}. This distributed architecture, which parallels neural population coding in biological brains \citep{nastase-et-al-2026, rumelhart-mcclelland1986}, grants LLMs significant global redundancy. However, recent mechanistic interpretability research reveals a paradox: despite this distributed redundancy, models remain vulnerable to catastrophic, highly specific functional collapse when focal critical circuits are disrupted \citep{He-etal-2026, Qin-etal-2026, lu-etal-2026}. This creates a theoretical tension: how can an LLM be so globally distributed, yet remain vulnerable to focal damage that produces specific linguistic breakdowns?

\begin{figure}[t]
    \centering
    \includegraphics[width=\columnwidth]{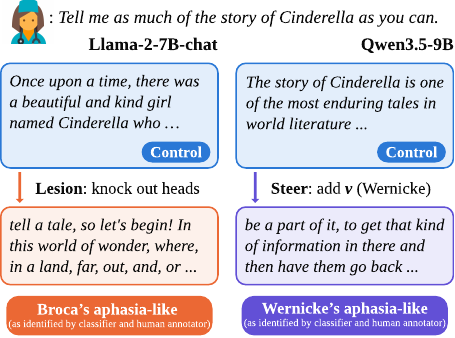}
    \caption{\textbf{Left}: knocking out attention heads inside \llamaseven makes its Cinderella story telling \textit{fragmented and incoherent}, as in Broca's aphasia; \textbf{Right}: adding a direction specific to Wernicke's aphasia to the activations of the \qwen{9B} makes its story \textit{fluent but irrelevant}, as in Wernicke's aphasia.}
    \label{fig:teaser}
\end{figure}

To investigate this tension, we ask whether the internal architecture of LLMs possesses a functional structure analogous to the human language system \citep{teehan-etal-2022, fedorenko2024language, kuribayashi-etal-2025, beckmann-queloz2026}. In clinical neuropsychology, the brain's functional organization is historically mapped by observing how language breaks down following focal injury (e.g., through the study of aphasia, an acquired language disorder due to brain injury or disease) \cite{dronkers-etal2017, geschwind1970}. Standard NLP metrics, such as perplexity or downstream accuracy, only indicate \textit{that} a model is degrading \citep{Qin-etal-2026}; they cannot easily characterize the nature of the deficit. Aphasia, however, provides a standardized, granular taxonomy for linking focal damage to highly specific language deficits \citep{hoffmann2013spectrum}. By adopting this clinical diagnostic lens, we gain a practical framework to categorize (i) \textit{how} an LLM's language generation breaks down when perturbed, and (ii) \textit{how} an intact LLM’s activation patterns change when processing clinically-informed impaired inputs.

Our framework thus consists of two complementary methodologies: lesion and steering. 
\begin{itemize}[nosep, leftmargin=*]
    \item In the \textbf{lesion} approach, we developed a classifier trained on clinical diagnostic data that classifies model behavior according to aphasia subtypes (e.g., \textit{Broca, Wernicke}; see Appendix~\ref{appsec:aphasia}). By systematically ablating model components, we map specific structural interventions to their corresponding aphasia-like output categories. This allows us to test whether certain linguistic functions can be linked to specific components within LLMs.
    
    \item In the \textbf{steering} approach, we derive intervention vectors by calculating the difference in the model's residual stream activations when processing human aphasic transcripts versus healthy controls. By systematically adding these steering vectors to targeted layers during text generation, we shift the model's behavior toward specific aphasia profiles (e.g., Broca's or Wernicke's). This allows us to test whether complex language deficits can be induced through continuous activation interventions, and to explore whether subtracting these directions can effectively reverse or ``heal'' the linguistic effects of a simulated lesion.
\end{itemize}
We employ this dual approach to robustly cross-validate our findings. While ablating targets the model's structural ``hardware'' (discrete circuits), steering intervenes on its representational ``software'' (continuous state). Proving that aphasic behavior can be induced both structurally and representationally ensures our results reflect fundamental properties of the model, rather than artifacts of a single methodology.

Applying this framework, our main findings are:
\begin{itemize}[nosep, leftmargin=*]
\item \textbf{Dual existence of functional mechanisms:} Classifier-guided lesion successfully isolates discrete circuits governing distinct language functions (e.g., grammatical fluency, semantic coherence), while steering identifies corresponding continuous vectors. Crucially, targeted subtype lesions succeed far more often when guided by our classifier.
\item \textbf{Dense functional entanglement:} Distinct aphasic profiles are highly entangled. Subtypes share significant structural overlap in their lesioned circuits and exhibit high cosine similarity across their steering vectors.
\item \textbf{Asymmetric generalizability:} While structural lesion effects are constrained by the classifier's specific training distribution, representational steering generalizes robustly across diverse downstream tasks.
\end{itemize}

Our study has the following contributions. We (1) introduce a novel interdisciplinary framework that adapts clinical aphasia to map LLM failure modes, (2) provide empirical evidence that macroscopic linguistic breakdowns correspond to both specialized circuits and continuous vectors, and (3) demonstrate that LLM language generation, in a way functionally analogous the human brain, relies on a densely entangled, distributed architecture vulnerable to highly specific disruptions.


\section{Related Works}
\label{sec:related}

\paragraph{Aphasia and LLMs.}

Recent research has begun utilizing clinical language deficits to map model degradations \cite{wang2025emergent, wang2025bridging, yang-etal-2026, roll-etal-2026}, though interpretations of the parallels between LLMs and humans remain contested. \citet{yang-etal-2026} demonstrated that systematically lesioning the decoder layers of multimodal models can reproduce the complete, joint error profiles of individual persons with aphasia (PWAs) during a picture-naming task. They observed a distinct depth gradient: perturbing early layers caused complete response failure, middle layers triggered semantic errors, and late layers yielded neologisms. Conversely, \citet{roll-etal-2026} evaluated damaged language models using a Text Aphasia Battery (TAB; \citet{roll2025tab}) and concluded that the resulting error distributions do not genuinely resemble human clinical aphasia, arguing that LLMs and the human brain are fundamentally organized in different ways.

Our research bridges these perspectives. While \citet{yang-etal-2026} focused on single-word vision-language tasks to replicate individual patient profiles, we focus on continuous text generation to localize specific internal functional networks. Furthermore, our theoretical stance aligns with the cautions raised by \citet{roll-etal-2026}: we deploy aphasic subtypes strictly as a diagnostic framework to categorize model vulnerability and internal overlapping components, without assuming a 1:1 modularity with biological brain regions like Broca's or Wernicke's area \cite{broca-1861, wernicke-1874}. Ultimately, by combining lesioning to test focal vulnerabilities and steering to map the intact distributed geometry, we investigate whether a highly distributed LLM can genuinely exhibit focal, aphasia-like functional structures.

\paragraph{Lesions and causal analysis.}
A high degree of redundancy in LLMs protects them from drastic performance degradation~\citep{frantar-etal-2023, sun-etal-2023, wang-etal-2023, shi-etal-2023}.
Most LLMs' attention heads can be removed with little loss~\citep{michel2019sixteen, voita2019analyzing}, and many attention layers can be dropped from large models~\citep{He-etal-2026}, yet a handful of critical neurons can collapse a model into repetitive noise~\citep{Qin-etal-2026, lu-etal-2026}.
Circuit analyses localize a behavior to a set of components with lesioning and activation patching~\citep{vig2020investigating, wang2023interpretability, zhang2024towards}, and learned masks find such sets by gradient descent~\citep{csordas2021neural, decao2020decisions}.
These methods track a loss of accuracy or a target token's probability, whereas we search for lesions that produce a specified clinical behavior in free narration.

\paragraph{Steering.}
Unlike leisoning, steering adds a direction to a model's activations while it generates, which shifts its behavior without changing its weights~\citep{subramani-etal-2022-extracting, turner2023steering, Zhang2026LocateSA}.
Linear probes show that many properties of the input can be decoded from hidden states~\citep{alain2017understanding, hewitt-manning-2019-structural, belinkov-2022-probing}, and concepts such as the truth of a statement are often encoded along a single direction of a model's representations~\citep{park2024linear, marks2024geometry}.
The direction is often the difference between the mean activations on two contrasting sets of inputs~\citep{zou2023representation, rimsky-etal-2024-steering, li2023inference}, and removing such a direction from every layer can stop a behavior~\citep{arditi2024refusal}.
These studies construct the contrasting inputs themselves, for example harmful and harmless instructions, and so does the closest work to ours, which steers models to simulate cognitively impaired patients for clinical training~\citep{zhang2026beyond}.
We take the contrasting inputs from clinical data, the transcripts of speakers with and without aphasia, which the model reads as its own answers, and ask experts whether the steered speech matches the targeted type of aphasia.

\begin{table}[t]
\centering
\footnotesize
\setlength{\tabcolsep}{4pt}
\begin{tabular}{@{}lrrrr@{}}
\toprule
 & \multicolumn{2}{c}{Cinderella} & \multicolumn{2}{c}{Sandwich} \\
\cmidrule(lr){2-3} \cmidrule(l){4-5}
Group & Transcripts & Speakers & Transcripts & Speakers \\
\midrule
Control & $395$ & $288$ & $472$ & $360$ \\
Anomic & $272$ & $148$ & $187$ & $131$ \\
Conduction & $141$ & $73$ & $81$ & $62$ \\
Broca & $273$ & $113$ & $111$ & $77$ \\
Wernicke & $58$ & $30$ & $29$ & $23$ \\
\midrule
All aphasia & $744$ & $353$ & $408$ & $281$ \\
All & $1{,}139$ & $637$ & $880$ & $638$ \\
\bottomrule
\end{tabular}
\caption{Dataset statistics for the two AphasiaBank tasks.
}
\label{apptab:data-transcripts}
\end{table}

\section{Data and Setup}
\label{sec:setup}


\paragraph{Data.}
AphasiaBank~\citep{macwhinney2011aphasiabank} is a large-scale database of naturalistic speech from individuals with aphasia and neurologically healthy controls.
We focus on two tasks that elicit different types of speech: \emph{Cinderella} task, in which speakers retell the story of Cinderella to produce narrative speech, and \emph{sandwich} task, in which speakers describe how to make a peanut butter and jelly sandwich, eliciting procedural speech with a fixed sequence of steps (see Table~\ref{apptab:data-transcripts} for statistics).


\paragraph{Models.}
We study 11 models from four families: \gptxl~\citep{radford2019language}; \llamaseven and \llamathirteen~\citep{touvron2023llama}; \pythia{1B}, 1.4B, 2.8B and 6.9B~\citep{biderman2023pythia}; and \qwen{0.8B}, 2B, 4B and 9B~\citep{qwen2026qwen35}, with thinking disabled.
We use the AphasiaBank prompts, \say{Tell me as much of the story of Cinderella as you can} and \say{Tell me how you would make a peanut butter and jelly sandwich}, to enable comparison with human speech elicited by the same tasks.
We pass these prompts directly to \gptxl and Pythia and format them using the respective chat templates for Llama-2 and Qwen3.5 (Table~\ref{apptab:data-prompts}).
All models generate at most 180 new tokens using greedy decoding with a repetition penalty of 1.25, as unpenalized greedy decoding causes the base models to loop even without intervention.
Across both interventions and tasks, we include only responses with at least 20 characters.
\begin{figure}[t]
    \centering
    \includegraphics[width=\linewidth]{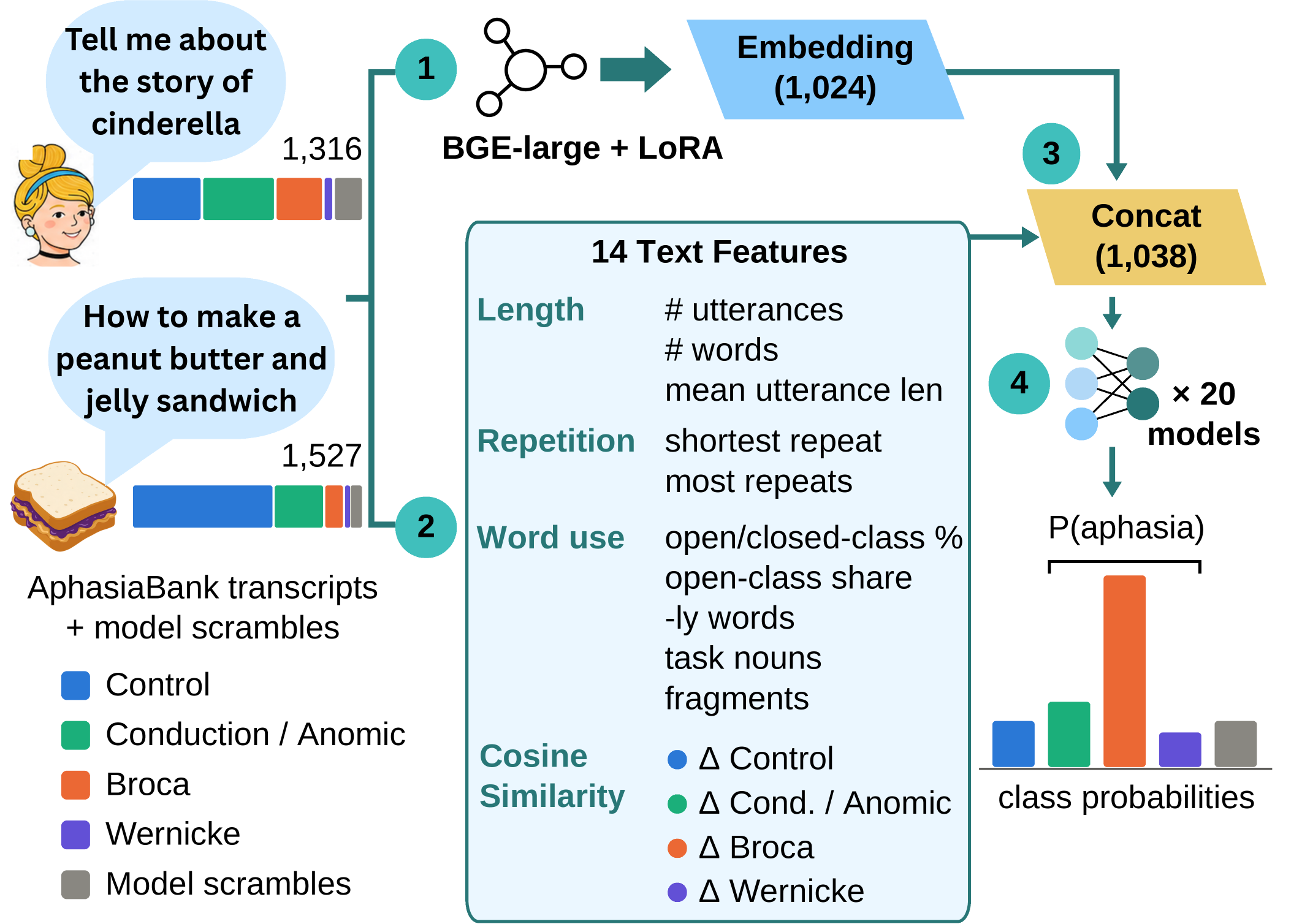}
    \caption{Five-way classifier trained on AphasiaBank transcripts and labeled model scrambles using 14 text features and a text embedding, ensembled over 20 seeds.}
    \label{fig:classifier}
\end{figure}

\begin{figure*}[t]
    \centering
    \includegraphics[width=\textwidth]{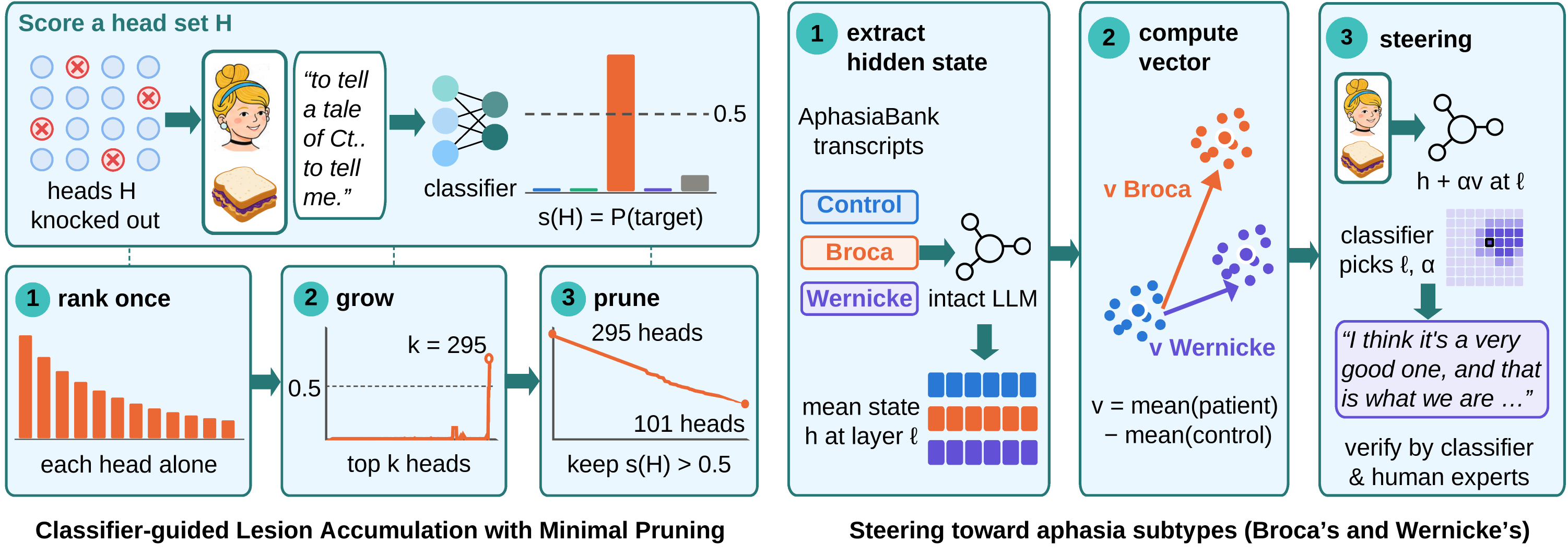}
    \caption{\textbf{Left}: CLAMP ranks heads by their score alone, knocks out the top $k$ until the target probability exceeds 0.5, and prunes the set. \textbf{Right}: a subtype vector, the patients' minus the controls' mean hidden state, is added at a layer and strength chosen by the classifier.}
    \label{fig:pipeline}
\end{figure*}

\section{Method}
\label{sec:method}

We use two interventions to a frozen model: classifier-guided lesioning, which disables attention heads, and subtype steering, which adds a direction to an intact model's activations (Figure~\ref{fig:pipeline}).


\subsection{Model Lesioning}
\label{sec:method-lesion}

Prior methods intended to induce aphasic behaviour instead produce model failures. 
For example, \citet{Qin-etal-2026} remove a few critical neurons, causing model outputs to collapse into repetitive noise, while \citet{wang2025emergent} guide lesioning by matching model outputs to patient speech using text similarity.
Human experts identify the resulting outputs as model scrambles.
We therefore train a classifier to distinguish aphasia from model scrambles and use it to guide the lesion search.

\paragraph{Five-way classifier.}
The classifier assigns a text a probability for each of five classes: control speech, conduction or anomic aphasia, Broca's aphasia, Wernicke's aphasia, and model scrambles.
Conduction and anomic aphasia form one class because the two are hard to tell apart from free speech alone: the Western Aphasia Battery separates them by a repetition test~\citep{kertesz2007western}.
We introduce a ``model scrambles'' category to capture non-human machine artifacts, such as tokenizer loops, stray brackets, or formulaic repetitions. Isolating these distinct computational failures prevents machine-generated noise from contaminating the mapping between human clinical profiles and internal model components. The training data hold 259 model-scramble outputs of \llamaseven and \gptxl from a preliminary lesion study, selected using LLM-as-a-judge with human annotator validation.
The classifier combines a BGE-large sentence encoder~\citep{xiao2023cpack}, fine-tuned with low-rank adapters~\citep{hu2022lora}, with 14 text features (Figure~\ref{fig:classifier}), and it averages 20 such models trained on different random splits.

On held-out transcripts whose text appears only once in the data, it reaches an accuracy of 83.3\% and a macro-averaged F1 of 0.78.
Lesioning and steering use three targets read off their probabilities: \paph, the total probability of the three aphasia classes, which separates aphasic speech from both control speech and model scrambles; \pbroca; and \pwern.

\paragraph{\ourmethod.}
Since attention layers contain multiple parallel heads, we hypothesize that selectively lesioning heads can induce aphasia-like behaviour without producing model scrambles.
We propose Classifier-guided Lesion Accumulation with Minimal Pruning (\ourmethod), which ablates attention heads by zeroing their outputs before the attention output projection at every prompt and generated-token position.
For a fixed model, task, and target class, let $s(H)$ denote the classifier's probability of that class for the response generated with head set $H$ ablated.
\ourmethod{} proceeds in three steps (Figure~\ref{fig:pipeline}, left).
First, it ablates each head individually and ranks heads in descending order of $s(\{h\})$.
Second, it builds $H$ by adding heads in this order until their joint ablation yields $s(H)>0.5$.
Finally, it prunes heads individually, or in batches for large sets, as long as $s(H)$ remains above 0.5.
In the final set, restoring any single head reduces $s(H)$ to 0.5 or below.
For the Cinderella task, we first ask \say{Do you know the story of Cinderella?}.
If the intact model answers affirmatively, we include the lesioned model's narrative only if it also answers \say{yes} to this question.

\paragraph{Baselines.}
We compare \ourmethod with eight lesion methods (Appendix~\ref{appsec:lesion-methods}).
\Param is the emergent-modularity method of \citet{wang2025emergent}: it zeroes the weights with the top 1\% of $|w \, \partial L / \partial w|$ in each weight matrix, where $L$ is the loss on control narratives, and groups the lesions by the patient group whose transcripts the lesioned model's answer resembles most by BLEU-1 and BERTScore.
\ParamMask and \HeadMask learn, with the model frozen, how much to knock out each of \Param's lesions of the MLP weight matrices or each head, as \citet{csordas2021neural} learn weight masks, so that the transcripts of the target group become more likely and those of the other groups less likely.
\ParamMaskHard and \HeadMaskHard knock out the components whose learned amount exceeds 0.5 and prune them as above.
These four methods learn from the transcripts; the classifier only chooses the reported training step and, for the hard masks, guides the pruning.
\HeadCEM optimizes the classifier directly with the cross-entropy method~\citep{rubinstein1999cross}, sampling 64 head sets per round for at most 20 rounds.
\ParamC grows \Param's MLP and attention head lesions by the classifier as \ourmethod.
\HeadGrownCos grows and prunes heads as \ourmethod does, but with a corpus similarity in place of the classifier: the answer's mean cosine similarity to the target group's transcripts minus that to the control transcripts~\citep{reimers2019sentence}.

\paragraph{Beyond the narrative.}
To evaluate other aphasia-related abilities, we test every lesion on the other task, on the Text Aphasia Battery (TAB)'s short-answer items for comprehension and repetition and its five connected-text prompts~\citep{roll2025tab}, and on two MMLU subjects, global facts (tests broad general world knowledge) and high-school world history (covers major global historical eras, movements, empires and revolutions taught in secondary education)~\citep{hendrycks2021measuring}, by the drop in the log-probability of the correct answer.
A DeepSeek-V4.1-Flash~\citep{deepseekai2026deepseekv41flash} LLM judge rates each connected-text answer for the TAB's 19 aphasia features, with the TAB's instructions with in-context examples of the features (Appendix~\ref{appsec:judge}).

\subsection{Steering}
\label{sec:method-steer}

Steering builds its directions from the human transcripts (Figure~\ref{fig:pipeline} right).
For each task, the model reads every transcript $x$ appended to the task prompt as if it were its own answer, and $h_\ell(x)$ is the residual stream after layer $\ell$, averaged over the tokens of $x$.
The steering vector for a target at layer $\ell$ is the difference between two group means,
\begin{equation}
  \vvec_\ell = \frac{1}{|T|} \sum_{x \in T} h_\ell(x) - \frac{1}{|C|} \sum_{x \in C} h_\ell(x),
  \label{eq:steering-vector}
\end{equation}
where $C$ holds the control transcripts and $T$ the transcripts of Broca's aphasia for the Broca target, and of Wernicke's aphasia for the Wernicke target.
Steering adds $\alpha \vvec_\ell$ to the residual stream after layer $\ell$ at every position while the model reads the prompt and generates.
We report its strength as $r = \|\alpha \vvec_\ell\| / \bar h_\ell$, where $\bar h_\ell$ is the mean norm of $h_\ell(x)$ over the transcripts.
For each model, target, and source task, a sweep over 8 layers spread across the model's depth and 8 strengths from $r = 0.1$ to $4$ picks the smallest strength, and the layer, at which the steered answer to the source task reaches the target. 
If no setting reaches it, we keep the setting with the highest probability.

The steered model answers seven prompts with this setting: the source task, which computed it, and six prompts that did not, the other task and the five connected-text prompts of the TAB.
Two experts label every answer from models with one of the classifier's five classes when steered toward a type.
We collect the annotation to check if the steering is successful (Appendix~\ref{appsec:human}).

\paragraph{Steering back.}
To test whether steering can undo a lesion, we remove the target's direction from the residual stream of the lesioned model after every block, replacing each state $h$ with $h - (h^\top \hat\vvec_\ell)\, \hat\vvec_\ell$, where $\hat\vvec_\ell = \vvec_\ell / \|\vvec_\ell\|$~\citep{arditi2024refusal}.
We also add the direction with a negative strength at a single layer.
As controls, we remove random directions in the same way, and we add random directions of the same norm.

\section{Results}
\label{sec:results}

\begin{table}[t]
\centering
\small
\setlength{\tabcolsep}{6pt}
\begin{tabular}{@{}lrrrr@{}}
\toprule
 & \multicolumn{3}{c}{Target} & \\
\cmidrule(lr){2-4}
Method & Aph. & Broca & Wern. & All \\
\midrule
\multicolumn{5}{@{}l}{\textit{Guided by corpus statistics}} \\
{\Param} & $2$ & $0$ & $0$ & $2$ \\
\ParamMask & $4$ & $0$ & $0$ & $4$ \\
\ParamMaskHard & $3$ & $1$ & $0$ & $4$ \\
\HeadMask & $5$ & $0$ & $0$ & $5$ \\
\HeadMaskHard & $8$ & $1$ & $1$ & $10$ \\
\HeadGrownCos & $2$ & $0$ & $0$ & $2$ \\
\midrule
\multicolumn{5}{@{}l}{\textit{Guided by the classifier}} \\
\ParamC & $11$ & $10$ & $4$ & $25$ \\
\HeadCEM & $9$ & $8$ & $6$ & $23$ \\
\ourmethod & $\pmb{14}$ & $\pmb{11}$ & $\pmb{13}$ & $\pmb{38}$ \\
\bottomrule
\end{tabular}
\caption{Number of settings in which a lesion method finds a lesion that makes the model's answer reach the target, by the five-way classifier in 42 settings.}
\label{tab:lesion-methods}
\end{table}

We focus on reporting results about the Broca’s and Wernicke’s subtypes, as their contrasting failure modes (broken grammar vs. semantic incoherence) provide unambiguous mechanistic signals. Conduction and Anomic profiles are omitted, as their subtle free-production signatures heavily overlap with healthy controls (Figure~\ref{appfig:tsne-transcripts}).

We find language functions in LLMs can be successfully isolated to discrete structural circuits and continuous representational vectors.

\subsection{Model Lesioning}
\label{sec:results-lesion}

\paragraph{Classifier-guided search outperforms corpus-based search.}
Across 42 settings spanning seven models (\gptxl, two Llama-2-chat models, and four Pythia models), two tasks, and three targets, classifier-guided methods find aphasia-inducing lesions far more often than methods guided by corpus statistics (Table~\ref{tab:lesion-methods}). 
Human experts agree with the classifier's label on 90\% of the lesioned models' answers to both Cinderella and sandwich tasks.

\paragraph{Head lesioning is more effective than parameter lesioning.}
\ourmethod{} induces the target aphasic behaviour in 38 of 42 settings; the remaining four yield model scrambles.
\ParamC{}, which lesions parameters rather than heads, succeeds in only 25 settings, supporting our hypothesis that selectively lesioning heads reduces model degeneration.
Meanwhile, \HeadCEM{}, which optimizes the classifier objective more aggressively, succeeds in only 23 settings, and 12 of the remaining 19 yield model scrambles.
\ourmethod{} instead uses classifier scores for head ranking and threshold checks during pruning, which may help keep responses closer to the classifier's training distribution.

\begin{figure}[t]
    \centering
    \includegraphics[width=\columnwidth]{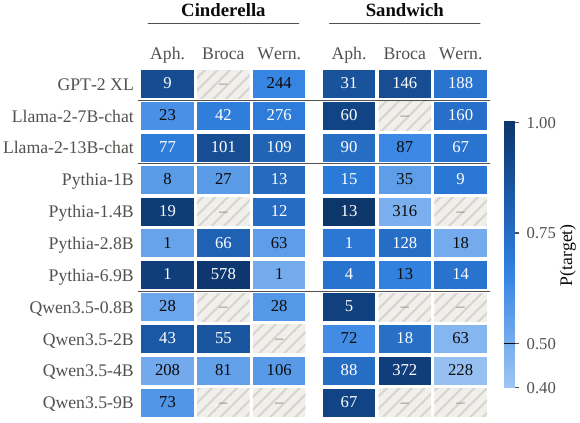}
    \caption{\ourmethod lesions in 11 models, two tasks, and three targets: aphasia (Aph.), Broca's and Wernicke's aphasia.
    A cell gives the lesion's number of heads, and its color the classifier's target probability for the lesioned output.
    Gray cells mark no effect of lesioning.}
    \label{fig:grown-heads}
\end{figure}

\paragraph{\ourmethod{} generalizes to Qwen models with linear attention.}
Across all 11 models, \ourmethod{} finds successful lesions in 54 of 66 settings, with the remaining settings yielding model scrambles (Figure~\ref{fig:grown-heads}).
We find aphasia-inducing lesions for every model on both tasks, while Broca-like and Wernicke-like behaviours are induced in 15 and 17 of 22 settings, respectively.
For example, ablating 101 heads in \llamathirteen{} transforms its fluent Cinderella narrative into \say{to tell a tale of Ct.. to tell me.}, with a classifier score of $\pbroca = 0.89$, while the model still confirms that it knows the story.

\paragraph{The heads of a lesion act together while the model reads the prompt.}
We use activation patching to identify when lesioned heads affect model behavior.
Restoring the clean activations of all heads of a lesion while the model reads the prompt, so that the heads stay knocked out only while it writes its answer, undoes 53 of the 54 lesions.
Restoring a single head undoes the lesion for a median of 83\% of a lesion's heads, yet the classifier labels 80\% of these answers as model scrambles and only 13\% as control speech.
The human aphasia-like answer thus needs nearly all heads of the lesion acting on the prompt together, a circuit in which changing one head mostly breaks the model (see Appendix~\ref{appsec:patching}).

\paragraph{Lesions for different aphasia types share heads.}
The Broca and Wernicke lesions of the same model and task share a median of 54\% of the heads of the smaller lesion.
Across the 13 pairs, 517 heads are shared, three times the number expected by chance, and overlap is significant in 11 pairs ($p < 0.01$).
This overlap begins at single heads: a head whose knockout makes the answer more Broca-like usually makes it more Wernicke-like too, at a median rank correlation of 0.93 across heads.
The circuits for Broca-like and Wernicke-like breakdowns thus share far more heads than chance, entangling the two types rather than separating them.

\begin{table}[t]
\centering
\small
\setlength{\tabcolsep}{8pt}
\begin{tabular}{@{}lrrr@{}}
\toprule
 & \multicolumn{2}{c}{TAB} & \\
\cmidrule(lr){2-3}
Family & Features & Short Answer & MMLU \\
\midrule
GPT-2 & $0.8 \rightarrow 2.6$ & $8/15$ & $0.24$ \\
Llama-2-chat & $0.0 \rightarrow 1.8$ & $3/15$ & $1.59$ \\
Pythia & $0.0 \rightarrow 2.0$ & $4/15$ & $0.11$ \\
Qwen3.5 & $0.2 \rightarrow 4.8$ & $4.5/15$ & $1.11$ \\
\midrule
All & $0.0 \rightarrow 2.7$ & $4/15$ & $0.89$ \\
\bottomrule
\end{tabular}
\caption{What the \HeadGrown lesions do beyond the narrative they were found on, as medians over the lesions that reach their target.
\emph{Features}: aphasia features of the TAB, of 19, that the LLM judge finds per connected-text answer of the intact and of the lesioned model.
\emph{Short Answer}: TAB short-answer items for comprehension and repetition that the lesioned model gets right, of 15.
\emph{MMLU}: drop in the log-probability of the correct answer on two MMLU subjects, in nats.}
\label{tab:beyond}
\end{table}

\paragraph{The lesions damage language on new prompts, but not as the targeted type.}
Almost every lesion found on one narrative also damages the model's answers to the TAB (Table~\ref{tab:beyond}).
The LLM judge finds more of the TAB's aphasia features in the connected text for 50 of the 54 lesions than for the intact models, a median of 2.7 of 19 features per answer against 0.0.
The lesioned models get a median of 4 of the 15 short-answer items right, against 8.5 for the same models intact.
The targeted type, however, does not carry over: the Broca and Wernicke lesions reach their type in only 1 of the 184 scored answers that they give to the other narrative and to the TAB's connected-text prompts (Appendix~\ref{appsec:beyond}).
The lesions thus damage language broadly, but the type that they induce stays in the answer on which the search found them.

\subsection{Steering}
\label{sec:results-steer}

\begin{figure}[t]
    \centering
    \includegraphics[width=\columnwidth]{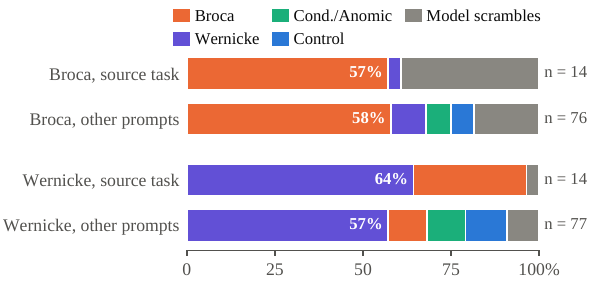}
    \caption{Expert labels of the answers steered toward Broca's or Wernicke's aphasia, by target and prompt.
    \emph{Source task}: the narrative whose answers chose the steering layer and strength; \emph{other prompts}: the other narrative and the five connected-text prompts of the Text Aphasia Battery (TAB).
    An answer counts as reaching the target when either expert labels it so; any other answer is split evenly between its two experts' labels.}
    \label{fig:steer-human}
\end{figure}

\paragraph{One direction per type makes an intact model ``speak'' like a patient.}
Adding the Broca direction at a single layer makes the classification label Broca in all 11 models on both tasks, and the Wernicke direction does so for Wernicke's aphasia in 12 of the 22 model-task pairs (Appendix~\ref{appsec:steering}).
Two experts labeled the steered answers of seven models (Figure~\ref{fig:steer-human}): either expert identifies successful steering to the targeted type for 58\% of the Broca answers, and 58\% of the Wernicke answers.
Thus, adding a single direction to the residual stream can make an intact model produce speech resembling a targeted aphasia subtype. Steering toward Broca’s aphasia, however, often scrambles the model, possibly because the corpus contains no examples of scrambled model outputs.

\paragraph{Steering carries the targeted type to new prompts.}
On the six prompts not used to choose the steering setting, either expert labels the targeted type for 65\% of the Broca answers and 63\% of the Wernicke answers that are not model scrambles (Appendix~\ref{appsec:human}).
The type follows the direction: either expert labels 44 of the 76 Broca-steered answers to these prompts Broca-like but only 13 of the 77 Wernicke-steered ones, and 44 of the 77 Wernicke-steered answers Wernicke-like but only 9 of the 76 Broca-steered ones.
The classifier finds the targeted type in 67 of the 153 steered answers to new prompts, but in only 1 of the 139 scored answers of the Broca and Wernicke lesions of the same models.
A direction thus keeps its type on new prompts, whereas a lesion leaves only general damage.

\paragraph{The Broca and Wernicke directions share most of their length.}
Both directions point nearly along each other, at a median cosine over layers of 0.79 to 0.90 in every model and task, and along the aphasia direction of all aphasic transcripts, at a median cosine of 0.97 for Broca and 0.92 for Wernicke (Appendix~\ref{appsec:steering}).
The two types differ little: after projecting out the general aphasia direction extracted from the whole aphasia group, the remaining parts point away from each other, with median cosines of $-0.65$ to $-0.41$.
The Wernicke part carries its type: when added alone, it pushes \pwern above 0.5 in all 22 model-task pairs.
Like the lesions, the directions thus share most of what makes speech aphasic, with a small part that sets each type apart.


\begin{table}[t]
\centering
\small
\setlength{\tabcolsep}{14pt}
\begin{tabular}{@{}lrrr@{}}
\toprule
 & Two-way & Five-way & All \\
\midrule
Classifier & $\pmb{73}$ & $\pmb{62}$ & $\pmb{65}$ \\
LLM judge & $64$ & $51$ & $55$ \\
\midrule
Two experts & $88$ & $76$ & $80$ \\
\quad Cohen's $\kappa$ & $0.75$ & $0.68$ & $0.76$ \\
\bottomrule
\end{tabular}
\caption{Agreement (\%) between automated labelers and two human experts on steered model responses. Two-way: Binary classification (aphasic vs. non-aphasic) of responses steered toward aphasia. Five-way: Five-class classification of responses steered toward Broca's or Wernicke's aphasia.}
\label{tab:agreement}
\end{table}

\paragraph{The classifier agrees with the experts more often than an LLM judge.}
Across all steered answers, including those to the out-of-distribution TAB prompts, the classifier still matches either expert's label on 65\% of answers, compared with 55\% for a DeepSeek-V4.1-Flash judge. (Table~\ref{tab:agreement}
The two experts agree with each other on 80\% of answers, with a pooled Cohen's $\kappa$ of 0.76. 
Much of the classifier's advantage comes from the 77 answers the experts identified as model scrambles: its labels agree with either expert on 51 of these answers, compared with 23 for the judge.

\paragraph{Steering back undoes the lesions, but does not ``heal'' them.}
Removing a lesion's own direction from the residual stream after every block undoes 53 of the 54 lesions, but 52 of these answers are model scrambles (Appendix~\ref{appsec:rescue}).
Random directions undo the lesions almost as often: removing one of three random directions undoes 50 or 51 of the lesions, and 91\% of the pushes with a random direction of the same norm at one layer undo the lesion, against 95\% of those with the lesion's own direction.
Undoing a lesion is thus easy but unspecific: nearly every perturbation we tried moves the answer off its target, mostly into model scrambles rather than control speech.

\section{Conclusion and Discussion}
\label{sec:conclu}

In this work, we successfully bridged neuropsychology and mechanistic interpretability to map the failure modes of large language models. Specifically, we make the following core contributions:

\paragraph{Graded model degradation.} First, this work shifts the conceptual focus of AI safety and interpretability from catastrophic collapse to graded, structured degradation. By applying an aphasia-inspired diagnostic lens, we demonstrate that LLM breakdowns are not merely random noise, but structured deficits that align with recognized clinical profiles. This provides the field with a more granular vocabulary to evaluate ``bad'' or impaired language generation, opening the door to modeling other language deficits in artificial systems.

\paragraph{A complementary methodological framework for mechanistic interpretability.} Second, we contribute a robust, dual-approach framework for mechanistic interpretability. Relying on a single intervention method can yield artifacts tied to that specific technique. By cross-validating our findings through complementary approaches, combining the structural isolation of classifier-guided lesioning with the representational fluidity of continuous activation steering, we establish a more rigorous standard for proving that an identified language function is a fundamental property of the model. This dual inside-out/outside-in approach ensures that mapped vulnerabilities are highly reliable.

\paragraph{A lightweight classifier with real-world applicability.} Third, while developed primarily as part of the lesion approach, our classifier represents a standalone contribution to automated clinical annotation. Recent research has demonstrated the promise of combining LLM-based measures (such as surprisal) with traditional clinical features for aphasia classification \cite{congetal2024}. Our model advances this paradigm by offering a highly efficient, lightweight alternative with higher overall and per-class accuracy.

\clearpage
\section*{Limitations}

First, our lesions and the steering directions come from two English narrative tasks in AphasiaBank: the Cinderella story and the sandwich procedure. However, AphasiaBank holds further tasks, such as picture descriptions (e.g., participants are asked to look at the picture and tell a beginning, middle, and end of a story) and personal narratives (e.g., stroke, illness, and recovery story), and the same pipeline can be applicable to them with minimal changes (although in the case of the picture description task, Vision-Language Models should be involved), as it would to aphasia corpora in other languages.

Second, the classifier learns machine failures from only 259 expert-labeled examples. Although it detects only about half of the steered responses that experts classify as model-scrambled, incorporating additional expert-labeled failures into its training data could improve its performance. In both current and future studies, labels produced by the LLM judge and classifier should be interpreted as assistive signals rather than clinical diagnoses.

Third, our experiments focus on a limited set of language models, leaving the generalization of our findings to other model families and architectures uncertain. Evaluating additional models would help establish the robustness of our findings across model scales and generations. Furthermore, extending our interventions to mixture-of-experts (MoE) architectures is another important direction for future work, as sparse expert routing may influence the mechanisms and effectiveness of model lesioning and steering.

\bibliography{aphasia_llm}

\clearpage
\appendix
\section*{Appendix}
\startcontents[appendix]
\renewcommand{\thesection}{\Alph{section}}
\printcontents[appendix]{}{1}{\setcounter{tocdepth}{1}}
\setcounter{section}{0}

\section{Aphasia Types}
\label{appsec:aphasia}

Aphasia is a language disorder acquired through brain damage, most often from a stroke.
Clinicians classify it by three properties of a speaker's language: whether speech is fluent, whether the speaker understands what others say, and whether the speaker can repeat what others say~\citep{kertesz2007western}.
AphasiaBank assigns each speaker a type based on the Aphasia Quotient subtests from the Western Aphasia Battery-Revised and provides a severity score from 0 to 100, where lower values indicate more severe aphasia~\citep{macwhinney2011aphasiabank, kertesz2007western}.

\paragraph{Broca's aphasia.}
Speech is nonfluent and speakers produce short, effortful phrases that often drop function words and inflections, which is called agrammatism, while their comprehension is relatively preserved.

\paragraph{Wernicke's aphasia.}
Speech is fluent and may be abundant, but it carries little content and contains substituted words and sounds, called paraphasias, and speakers understand others poorly.

\paragraph{Conduction aphasia.}
Speech is fluent and comprehension is relatively good, but speakers repeat poorly and produce sound errors that they often try to correct.

\paragraph{Anomic aphasia.}
Speech is fluent and comprehension and repetition are good, but speakers have trouble finding words and talk around them.
Anomic aphasia is usually the mildest type.

The classifier merges conduction and anomic aphasia into one class, because the two differ mainly in repetition, which free speech such as a story or a recipe does not test.

\section{Data, Models and Setups}
\label{appsec:data}

This appendix gives the details on the human transcripts, the models, the prompts, the decoding rules, the rules that decide whether an answer counts, the two tests beyond the narratives, the Text Aphasia Battery (TAB) and MMLU, and the compute and software that we used.

\begin{table*}[t]
\centering
\small
\begin{tabular}{@{}llrrrr@{}}
\toprule
Model & Hugging Face identifier & Layers & Heads per layer & Heads in all & Softmax heads \\
\midrule
\multicolumn{6}{@{}l}{\textit{GPT-2~\citep{radford2019language}: base model, question-and-answer prompts}} \\
\gptxl & \texttt{gpt2-xl} & $48$ & $25$ & $1{,}200$ & $1{,}200$ \\
\midrule
\multicolumn{6}{@{}l}{\textit{Llama-2~\citep{touvron2023llama}: chat models, Llama-2 chat template}} \\
\llamaseven & \texttt{meta-llama/Llama-2-7b-chat-hf} & $32$ & $32$ & $1{,}024$ & $1{,}024$ \\
\llamathirteen & \texttt{meta-llama/Llama-2-13b-chat-hf} & $40$ & $40$ & $1{,}600$ & $1{,}600$ \\
\midrule
\multicolumn{6}{@{}l}{\textit{Pythia~\citep{biderman2023pythia}: base models, question-and-answer prompts}} \\
\pythia{1B} & \texttt{EleutherAI/pythia-1b} & $16$ & $8$ & $128$ & $128$ \\
\pythia{1.4B} & \texttt{EleutherAI/pythia-1.4b} & $24$ & $16$ & $384$ & $384$ \\
\pythia{2.8B} & \texttt{EleutherAI/pythia-2.8b} & $32$ & $32$ & $1{,}024$ & $1{,}024$ \\
\pythia{6.9B} & \texttt{EleutherAI/pythia-6.9b} & $32$ & $32$ & $1{,}024$ & $1{,}024$ \\
\midrule
\multicolumn{6}{@{}l}{\textit{Qwen3.5~\citep{qwen2026qwen35}: post-trained models, Qwen3.5 chat template with thinking turned off}} \\
\qwen{0.8B} & \texttt{Qwen/Qwen3.5-0.8B} & $24$ & $8$ / $16$ & $336$ & $48$ \\
\qwen{2B} & \texttt{Qwen/Qwen3.5-2B} & $24$ & $8$ / $16$ & $336$ & $48$ \\
\qwen{4B} & \texttt{Qwen/Qwen3.5-4B} & $32$ & $16$ / $32$ & $896$ & $128$ \\
\qwen{9B} & \texttt{Qwen/Qwen3.5-9B} & $32$ & $16$ / $32$ & $896$ & $128$ \\
\bottomrule
\end{tabular}
\caption{The 11 models, grouped by family, with the kind of model and the prompt format of each family.
In Qwen3.5, every fourth layer, layers 3, 7, 11, and so on when counting from 0, uses softmax attention, and the other layers use Gated DeltaNet linear attention.
For Qwen3.5, \emph{Heads per layer} gives the heads of a softmax layer and of a Gated DeltaNet layer, and a lesion can knock out heads of both kinds.
\emph{Softmax heads} counts the heads of the softmax layers, which in the other models are all of the heads.}
\label{apptab:data-models}
\end{table*}

\paragraph{Transcripts.}
We use 1,147 AphasiaBank transcripts~\citep{macwhinney2011aphasiabank} of the Cinderella task and 1,437 of the sandwich task, and some of these transcripts share a text.
The sandwich transcripts hold 557 texts that appear twice and we keep only one copy of each.
Four Cinderella texts appear once as anomic and once as control speech, so their group is unknown, and we drop both copies.
Table~\ref{apptab:data-transcripts} counts the remaining transcripts, each text once: 1,139 for the Cinderella task and 880 for the sandwich task.
The transcripts carry no speaker field, so we identify a speaker by the corpus and by the file name without its session suffix.
The probes and the steering vectors use these transcripts, whereas the classifier draws its training and test splits from all 2,584 transcripts, duplicates included (Appendix~\ref{appsec:classifier}).
Of the aphasic transcripts, 720 of 744 in the Cinderella task and 382 of 408 in the sandwich task come with the speaker's aphasia quotient, the severity score that Appendix~\ref{appsec:steering} compares with the projection of a transcript on the aphasia direction.

\paragraph{License.}
We used AphasiaBank data under TalkBank’s Ground Rules and the applicable CC BY-NC-SA 3.0 license. We report aggregate results but do not redistribute participant transcripts, recordings, or other restricted data. Researchers seeking the underlying data should request access directly from TalkBank.

\begin{figure*}[tp]
    \centering
    \includegraphics[width=\textwidth]{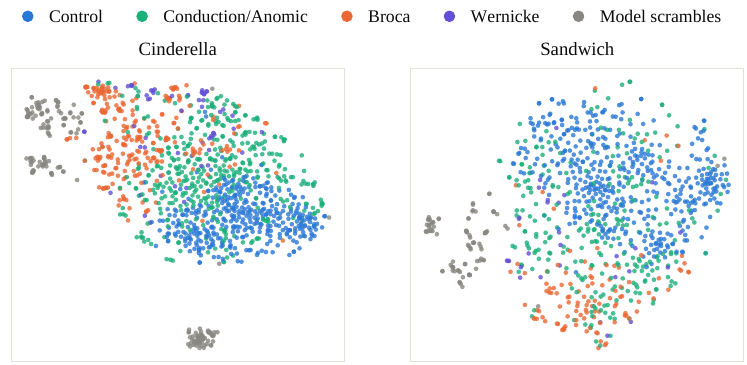}
    \caption{t-SNE of each task's transcripts and of the classifier's model scrambles, in the text space of the classifier's encoder, BGE-large, without the classifier's fine-tuning.
    The maps hold the 1,139 Cinderella and 880 sandwich transcripts of Table~\ref{apptab:data-transcripts}, with 169 and 90 model scrambles.
    Conduction and anomic transcripts share one color because they share one class of the classifier.}
    \label{appfig:tsne-transcripts}
\end{figure*}

\paragraph{Conduction and anomic transcripts lie among the control transcripts.}
We embed every transcript with BGE-large~\citep{xiao2023cpack}, the encoder that the classifier fine-tunes, here without its fine-tuning, so that the space knows nothing of our labels.
A transcript becomes the encoder's [CLS] vector over its first 384 tokens, the classifier's limit, scaled to length 1.
Figure~\ref{appfig:tsne-transcripts} maps these vectors with t-SNE, one map per task, together with the classifier's model scrambles (Appendix~\ref{appsec:classifier}).
We measure the overlap in the full embedding rather than on the map: for each transcript, we take the share of control transcripts among its 10 nearest transcripts by cosine similarity.
On the Cinderella and the sandwich task, this share is 0.89 and 0.91 for control transcripts, 0.44 and 0.58 for anomic transcripts, and 0.25 and 0.41 for conduction transcripts, against 0.07 and 0.11 for Broca's and 0.07 and 0.33 for Wernicke's aphasia.
The model scrambles form islands of their own: 92\% and 85\% of a scramble's 10 nearest texts are other scrambles, whereas scrambles make up 0.1\% of the nearest texts of the human transcripts.
Conduction and anomic transcripts thus resemble control speech more than Broca's and Wernicke's transcripts do, which is why we mainly report the Broca and Wernicke targets.

\paragraph{Models.}
Table~\ref{apptab:data-models} lists the 11 models and their Hugging Face identifiers.
We load every model with the Transformers library~\citep{wolf2020transformers} and run it in 32-bit floating point.
The Qwen3.5 models are the post-trained checkpoints, not the base checkpoints.
In Qwen3.5, every fourth layer uses full softmax attention, and the other layers use Gated DeltaNet linear attention~\citep{yang2025gated}, in which each head keeps its own recurrent state.
A lesion can knock out heads of both kinds, and softmax heads make up 14.3\% of the heads of every Qwen3.5 model.
Knocking out a head zeroes its slice of the input to the attention output projection of its layer, at every position of the prompt and of the answer; in the softmax layers of Qwen3.5, this slice holds the head's output after its gate.

\begin{table}[t]
\centering
\footnotesize
\begin{tabular}{@{}>{\raggedright\arraybackslash}p{0.28\columnwidth}>{\raggedright\arraybackslash}p{0.66\columnwidth}@{}}
\toprule
Family & Prompt template \\
\midrule
\gptxl and Pythia
 & \texttt{Question: }\textit{request}\texttt{\textbackslash n}\newline
   \texttt{Answer:} \\
\addlinespace
Llama-2-chat
 & \texttt{[INST] }\textit{request}\texttt{ [/INST]} \\
\addlinespace
Qwen3.5
 & \texttt{<|im\_start|>user\textbackslash n}\newline
   \textit{request}\texttt{<|im\_end|>\textbackslash n}\newline
   \texttt{<|im\_start|>assistant\textbackslash n}\newline
   \texttt{<think>\textbackslash n\textbackslash n</think>\textbackslash n\textbackslash n} \\
\midrule
Request & Text \\
\midrule
Yes-or-no question & Do you know the story of Cinderella? \\
\addlinespace
Cinderella Story request & Tell me as much of the story of Cinderella as you can. \\
\addlinespace
Sandwich request & Tell me how you would make a peanut butter and jelly sandwich. \\
\bottomrule
\end{tabular}
\caption{The prompts, verbatim.
Each prompt is the template of the model's family with one of the three requests in place of \textit{request}, and \texttt{\textbackslash n} marks a newline.
The prompts of \gptxl and of the Llama-2 models start with a beginning-of-sequence token, \texttt{<|endoftext|>} or \texttt{<s>}, which the templates leave out, and the prompts of Pythia and Qwen3.5 start with no such token.
Neither chat template adds a system prompt, and the Qwen3.5 template, with thinking turned off, closes an empty thinking block, so that the answer starts at once.}
\label{apptab:data-prompts}
\end{table}

\paragraph{Prompts.}
Table~\ref{apptab:data-prompts} gives the prompt template of each family and the three requests, verbatim.
The base models, \gptxl and Pythia, receive each request in a plain question-and-answer format, and the chat models receive it as a user message in their chat template, with no system prompt and, for Qwen3.5, with thinking turned off.
The prompts of \gptxl and of the Llama-2 models start with the beginning-of-sequence token, the prompts of Pythia start without it, and the Qwen3.5 tokenizer has no such token.
On the Cinderella task, the yes-or-no question and the story request are two separate prompts rather than two turns of one dialogue, and a lesion stays in place for both answers.

\paragraph{Decoding.}
For the three prompts, every model decodes greedily, with the repetition penalty set to 1.25 and applied to the raw logits.
At each step, the penalty divides the logit of every token that already occurs in the prompt or in the answer by 1.25 if the logit is positive, and multiplies it by 1.25 if the logit is negative.
Without the penalty, greedy decoding makes the intact base models \textit{loop} on the Cinderella task: the classifier gives the answers of \gptxl, \pythia{1B}, \pythia{1.4B}, and \pythia{2.8B} a probability of up to 99\% of being model scrambles.
For the Llama-2 models, greedy decoding replaces the sampling at temperature 0.6 and top-p 0.9 that their released generation settings specify.
A model writes at most 180 new tokens for the story and the sandwich answer, and at most 12 for the answer to the yes-or-no question, and it stops early at its end-of-sequence token or, for Qwen3.5, also at \texttt{<|endoftext|>}.
In the lesion searches, each model generates in batches of a fixed size, and we pad a short batch with copies of its first lesion, so that an answer does not depend on how the lesions were batched.
Greedy decoding thus gives each lesion a single answer, and regenerating every \ourmethod and \HeadGrownCos lesion from scratch reproduces its stored answer word for word.

\paragraph{What counts as an answer.}
A base model often goes on to write further questions and answers of its own, so we cut its answer before the first new line that starts with \texttt{Question:}, \texttt{Q:}, or \texttt{question:}, and the answers of the chat models stay whole.
An answer counts only if it has at least 20 characters, not counting invisible Unicode format characters, and a lesion whose answer does not count fails.
The rule leaves out format characters because some lesions of \llamathirteen make the model write a few letters followed by a long run of zero-width joiners, which are invisible: with its 170 joiners counted, the answer \say{tOoOO} would pass, and the classifier scores it at $\pbroca = 0.70$.
In the lesion experiments, the model first answers the yes-or-no question of the Cinderella task, and a parser matches the start of this answer, ignoring case and up to a word boundary, against two lists of prefixes.
The answer means yes when it starts with \texttt{yes}, \texttt{yeah}, \texttt{yep}, \texttt{yup}, \texttt{sure}, \texttt{of course}, \texttt{certainly}, \texttt{absolutely}, \texttt{definitely}, \texttt{i do remember}, \texttt{i remember}, or \texttt{i do}, and no when it starts with \texttt{no}, \texttt{nope}, \texttt{nah}, \texttt{sorry}, \texttt{i don\textquotesingle t}, \texttt{i dont}, \texttt{i do not}, \texttt{i cannot}, \texttt{i can\textquotesingle t}, \texttt{i cant}, \texttt{not really}, or \texttt{unfortunately}.
When both lists match, the longer match wins, and yes wins a tie.
A lesion fails when the model answers no, and an answer that matches neither list passes.
The check is off for \pythia{6.9B}, whose intact model already answers \say{No. / Q: You don't remember it, do}, where the slash marks a line break.
The lesion searches apply the check to every other model, and the steering experiments do not apply it.
The classifier scores the story or the sandwich answer after the cut, and never the answer to the yes-or-no question.

\begin{table}[t]
\centering
\footnotesize
\setlength{\tabcolsep}{4pt}
\begin{tabular}{@{}>{\raggedright\arraybackslash}p{0.27\columnwidth}>{\raggedright\arraybackslash}p{0.48\columnwidth}>{\raggedright\arraybackslash}p{0.17\columnwidth}@{}}
\toprule
Subtest & Prompt & Target \\
\midrule
Connected text
 & Tell me about the best trip you ever took.\newline
   Describe a happy childhood memory.\newline
   Tell me about your first job.\newline
   What do you like about where you live?\newline
   Describe the steps to make a simple meal.
 & -- \\
\addlinespace
Word comprehension
 & Which one is an animal with a mane? (lion, drum, violin, giraffe, boot, boat)
 & lion \\
\addlinespace
Sentence comprehension
 & Are babies watched by babysitters?
 & yes \\
\addlinespace
Repetition
 & Please repeat exactly: The sun rises in the East.
 & The sun rises in the East. \\
\bottomrule
\end{tabular}
\caption{The items of the Text Aphasia Battery (TAB) on which we test every lesion, transcribed from the TAB paper~\citep{roll2025tab}: all five connected-text prompts and one of the five items of each other subtest, verbatim.
Connected-text prompts follow the TAB's instruction \texttt{Respond to the following prompt in 3-5 full sentences:}, and the other items follow the instruction \texttt{You are having a conversation. Answer the question directly and briefly.}
The five-way classifier and the LLM judge score the connected-text answers, and rules compare the other answers with their targets.}
\label{apptab:data-tab}
\end{table}

\paragraph{Text Aphasia Battery.}
We transcribe 20 items of the TAB from the appendix of \citet{roll2025tab}, five for each of four subtests, and Table~\ref{apptab:data-tab} shows all five connected-text prompts and one item of each other subtest.
The connected-text items carry the TAB's own system instruction.
For the other three subtests, the TAB specifies only a conversational system prompt, so we use the minimal instruction given in Table~\ref{apptab:data-tab}.
A base model receives \texttt{Question: <instruction> <prompt>\textbackslash nAnswer:}, and we cut its answer as for the narratives, whereas a chat model receives the instruction as a system message and the prompt as a user message in its chat template.
The model answers a connected-text item as it answers the narratives, with the repetition penalty, in at most 120 new tokens.
It answers a short-answer item greedily without the penalty, in at most 40 new tokens, because the penalty would push the model away from the very words that repetition asks it to repeat and that word comprehension asks it to name.
Rules score the 15 short-answer items.
In word comprehension, the answer is correct when the first of the six options that it names, matched as a whole word regardless of case, is the target, which the prompt lists first.
In sentence comprehension, the first whole word \emph{yes} or \emph{no} in the answer must equal the target.
In repetition, the answer must equal the target after both are lowercased, stripped of surrounding quotation marks, spaces, and final periods and exclamation marks, and have their whitespace collapsed.
The exact match is strict for chat models, which often wrap the repetition in words of their own: the intact Llama-2 models repeat none of the five items exactly, and \llamathirteen answers \say{Sure! Breakfast.} when asked to repeat \emph{breakfast}.
The items lost by a lesion are the intact model's number of correct short-answer items minus the lesioned model's.
The five-way classifier scores each connected-text answer that has at least 20 characters, and an LLM judge, DeepSeek-V4.1-Flash~\citep{deepseekai2026deepseekv41flash} at temperature 0, marks which of the TAB's 19 aphasia features each connected-text answer shows, following the TAB's protocol with our own in-context examples (Appendix~\ref{appsec:judge}).
The five connected-text prompts are also the five TAB prompts of the human evaluation of steering (\S\ref{sec:method-steer}).

\paragraph{MMLU.}
We also test every lesion on two subjects of MMLU~\citep{hendrycks2021measuring}: global facts, with 100 test questions, and high school world history, with 237.
Every model receives the same raw-text prompt, without a chat template, after the beginning-of-sequence token where the model uses one.
The prompt is the standard five-shot prompt: the header \texttt{The following are multiple choice questions (with answers) about <subject>.\textbackslash n\textbackslash n}, where \texttt{<subject>} is \texttt{global facts} or \texttt{high school world history}, then five solved questions from the subject's development split, and last the test question.
Each question has the form \texttt{<question>\textbackslash nA. <a>\textbackslash nB. <b>\textbackslash nC. <c>\textbackslash nD. <d>\textbackslash nAnswer:}, and a solved question continues with a space, its correct letter, and two newlines.
The context of \gptxl holds only 1,024 tokens, so for its world-history prompts we drop the earliest solved questions until the prompt fits, which leaves three solved questions for 106 test questions, two for 111, and one for 20; every other model keeps all five for both subjects.
The model reads the prompt once and generates nothing.
Its predicted answer is the letter, among the tokens of A to D, with the highest logit after the final \texttt{Answer:}, and the log-probability of the correct letter is taken over the full vocabulary, so that it also falls when a lesion moves probability away from the four letters altogether.
The MMLU drop of a lesion is the intact model's mean log-probability of the correct letter minus the lesioned model's, in nats, averaged over the two subjects.
\paragraph{Compute.}
We ran the GPU and CPU-based experiments on a shared cluster.
The GPUs ran for 236 hours: 96.7 on NVIDIA A100 GPUs with 80~GB, 95.6 on H200 GPUs with 141~GB, 31.1 on A100 GPUs with 40~GB, and 12.8 on older GPUs.
The runs that completed account for 145.6 of these hours.
CPU jobs ran about 2,200 allocated core-hours.
These counts leave out the LLM judge, which we query through an API (Appendix~\ref{appsec:judge}).
\paragraph{Software.}
The code runs on Python 3.13, with PyTorch 2.7.1 and Transformers 5.5.4, which loads every model and generates every answer.
The classifier uses PEFT 0.19.1 for its LoRA adapters, and Sentence-Transformers 5.4.1 for its similarity features, which cosine similarity methods use too.
\Param scores answers with the BLEU of NLTK 3.9.4 and with bert-score 0.3.13.
The probes, cross-validation, t-SNE, and nearest neighbors use scikit-learn 1.8.0, the statistics and figures use SciPy 1.17.1, NumPy 2.4.4, and Matplotlib 3.10.9.

\section{Five-Way Classifier}
\label{appsec:classifier}

We next describe the training data, features, architecture, training, and evaluation of the five-way classifier of \S\ref{sec:setup}, and how it scores the answers of intact models.

\paragraph{Training data.}
The classifier learns from 2,843 texts: 2,584 AphasiaBank transcripts of the Cinderella and sandwich tasks, and 259 model scrambles (Table~\ref{apptab:classifier-data}).
The transcripts carry six group labels, and the classifier merges two of them, anomic and conduction aphasia, into one class (Appendix~\ref{appsec:aphasia}).
Some texts appear twice among the transcripts, and the training data keep both copies.
In the sandwich task, 557 texts appear twice: for 492 of them, the two copies are two files of the same recording, and for the other 65, they are files of two different recordings with the same group label.
In the Cinderella task, four recordings appear twice with the same text, once labeled anomic aphasia and once labeled control speech.
The 2,584 transcripts therefore hold 2,023 distinct texts.
\S\ref{sec:setup} counts each text once and leaves out the four texts with conflicting labels, which gives 1,139 Cinderella and 880 sandwich transcripts.
No text occurs in both tasks, and no model scramble repeats a human transcript.

\begin{table}[t]
\centering
\small
\setlength{\tabcolsep}{3pt}
\begin{tabular}{@{}lrrr@{}}
\toprule
 & Cinderella & Sandwich & Total \\
\midrule
\multicolumn{4}{@{}l}{\textit{AphasiaBank transcripts}} \\
Control & $399$ & $936$ & $1{,}335$ \\
Conduction or anomic & $417$ & $334$ & $751$ \\
\quad Anomic & $276$ & $235$ & $511$ \\
\quad Conduction & $141$ & $99$ & $240$ \\
Broca & $273$ & $131$ & $404$ \\
Wernicke & $58$ & $36$ & $94$ \\
All transcripts & $1{,}147$ & $1{,}437$ & $2{,}584$ \\
\quad Distinct texts & $1{,}143$ & $880$ & $2{,}023$ \\
\midrule
\multicolumn{4}{@{}l}{\textit{Model scrambles}} \\
\llamaseven & $113$ & $90$ & $203$ \\
\gptxl & $56$ & $0$ & $56$ \\
All model scrambles & $169$ & $90$ & $259$ \\
\midrule
All training texts & $1{,}316$ & $1{,}527$ & $2{,}843$ \\
\bottomrule
\end{tabular}
\caption{Training texts of the five-way classifier, by class and task prompt.
The AphasiaBank transcripts carry six group labels, and the classifier merges anomic and conduction aphasia into one class.
The model scrambles are answers of \llamaseven and \gptxl from a preliminary knockout study.
\emph{Distinct texts} counts each text once, because 557 sandwich texts and 4 Cinderella texts appear twice among the transcripts.}
\label{apptab:classifier-data}
\end{table}

\begin{table*}[t]
\centering
\small
\begin{tabular}{@{}l>{\raggedright\arraybackslash}p{10.2cm}@{}}
\toprule
Feature & Definition \\
\midrule
\multicolumn{2}{@{}l}{\textit{Narrative measures}} \\
Utterances & Number of utterances, each ending in ``.'', ``!'', or ``?'' \\
Word tokens & Number of words \\
Mean length of utterance & Mean number of words per utterance, leaving out utterances that are a single filler word such as ``uh'' or ``yeah'' \\
\midrule
\multicolumn{2}{@{}l}{\textit{Repetition statistics}} \\
Shortest repeated unit & Length in words of the shortest repeated unit, or 0 if there is none \\
Most repetitions & Largest number of times that a repeated unit occurs, or 0 if there is none \\
\midrule
\multicolumn{2}{@{}l}{\textit{Word-list proxies}} \\
Open- to closed-class ratio & Number of open-class words over number of closed-class words, which are the words of a list of 101 function words \\
Share of open-class words & Number of open-class words over number of words \\
Words ending in \emph{-ly} & Number of words ending in \emph{-ly}, a proxy for adverbs \\
Task nouns & Number of words from a list of 33 nouns of the two tasks \\
Fragments and stutters & Number of words that end in a hyphen, plus letters repeated after a hyphen \\
\midrule
\multicolumn{2}{@{}l}{\textit{Similarity features}} \\
Control minus Broca & Similarity to the control transcripts minus similarity to the Broca transcripts \\
Control minus Wernicke & Similarity to the control transcripts minus similarity to the Wernicke transcripts \\
Control minus conduction or anomic & Similarity to the control transcripts minus similarity to the transcripts of conduction or anomic aphasia \\
Similarity to control & Similarity to the control transcripts, meant as control minus model scrambles \\
\bottomrule
\end{tabular}
\caption{The 14 features that the five-way classifier reads besides its text encoder, each standardized with the mean and standard deviation of the training split.
A repeated unit is a sequence of 1 to 100 tokens, words or punctuation marks, that occurs several times in a row, or a sequence of six or more tokens that occurs again anywhere in the text.
Similarities are mean cosine similarities between the all-MiniLM-L6-v2 embedding of the text and those of the human transcripts of a group, from both tasks.
Because of a label mismatch in the code, the last feature subtracts zero instead of the similarity to model scrambles.}
\label{apptab:classifier-features}
\end{table*}

\paragraph{Model scrambles.}
The 259 model scrambles are answers from a preliminary knockout study: 203 come from \llamaseven, on both task prompts, and 56 from \gptxl, all on the Cinderella prompt.
All but five are answers of models with knocked-out attention heads or feed-forward blocks, and the other five are temperature samples of the intact \gptxl.
The knockouts remove single heads, sets of heads, or the feed-forward block of a whole layer, and 132 of these answers come from successive steps of a greedy search that knocks out more and more heads of \llamaseven.
Of the answers with knockouts, 28 come from activation patching, which restores the clean activations of one component in a knocked-out model, and 11 are temperature samples of \llamaseven with three heads knocked out.
An LLM judge flagged these answers as machine failures among the outputs of the preliminary study, and a linguist and a clinician reviewed them.
The model scrambles are 12 to 1,090 characters long, with a median of 571, and all 259 are distinct.
By the repetition statistics of Table~\ref{apptab:classifier-features}, 233 of them contain a repeated unit and 196 a unit that occurs three or more times, whereas 799 of the 1,335 control transcripts contain no repeated unit.

\paragraph{Judging machine-specific output.}
\label{appsec:llm-as-a-judge}
To distinguish machine-specific output failures from linguistically comparable clinical aphasic speech, we implemented an LLM judge pipeline to automatically detect and annotate machine artifacts. Using the Gemini 3.5 Flash model, this automated judge evaluates outputs from perturbed models to filter out ``mechanical garbage'', which are instances where the model breaks down mechanically rather than simulating a valid human linguistic profile. By identifying these ``model scrambles,'' we prevent machine-generated noise from contaminating the mapping between human clinical profiles and internal model components during the ablation pipeline.

The judge's evaluation is governed by a strict, zero-temperature system prompt that evaluates text against five specific artifact dimensions: Stray Brackets (SB), Token Repetition (TR), Tokenizer Loops (TL), Formulaic AI-isms (FM), and Sentence Repetition (SR). To ensure precise categorization, the model outputs a structured JSON object detailing its step-by-step reasoning and the specific artifacts detected. Crucially, the prompt includes a tie-breaker rule designed specifically for clinical data: if a repetition mimics natural human hesitation or spoken disfluency, the LLM assumes a valid human purpose and refrains from flagging it as a machine artifact. This ensures that the valid disfluencies characteristic of aphasic subtypes are preserved while machine-specific degradations are correctly recognized. For the complete zero-temperature system prompt, including the explicit artifact rubric and few-shot evaluation examples.
Listing~\ref{lst:judge-artifact-user} shows the whole prompt.

To ensure the reliability of the LLM judge pipeline, the system prompt was iteratively calibrated against a gold-label subset of the dataset to maximize its baseline agreement rate. Furthermore, to validate the integrity of the automated filtering, the final classifications were audited by human domain experts. A linguist and a clinician manually reviewed the text flagged by the LLM, confirming that the isolated ``model artifacts'' were true machine-specific failures and not misclassified clinical disfluencies. This empirical prompt tuning combined with expert clinical verification ensures that our artifact removal process is both scalable and clinically accurate.

Finally, these verified ``model artifacts'' are incorporated directly into the training data for our 5-way clinical diagnostic classifier.

\paragraph{Features.}
Besides its text encoder, the classifier reads the 14 features of Table~\ref{apptab:classifier-features}.
The three narrative measures reimplement measures that the Computerized Language Analysis (CLAN) program computes for AphasiaBank transcripts~\citep{macwhinney2017tools}, so that the classifier can compute them from plain text such as a model's answer.
On the human transcripts, they correlate with CLAN's values at a Pearson's $r$ of 0.986 for the number of utterances, 0.993 for the number of word tokens, and 0.968 for the mean length of utterance.
The five word-list proxies need no part-of-speech tagger, but they approximate CLAN's measures less closely.
The open- to closed-class ratio and the share of open-class words correlate only weakly with the corresponding CLAN measures, at $r = 0.095$ and $r = -0.050$.
The count of words ending in \emph{-ly} correlates with CLAN's adverb count at $r = 0.618$, and 73.2\% of its matches in the sandwich transcripts are the word \emph{jelly}.
The count of fragments and stutters is zero on every human transcript, although CLAN counts word errors in 1,090 of them.
The four similarity features embed the whole text with all-MiniLM-L6-v2~\citep{wang2020minilm, reimers2019sentence} and compare it, by mean cosine similarity, with the human transcripts of each group, pooled over both tasks.
Three of them subtract the mean similarity to the transcripts of Broca's aphasia, of Wernicke's aphasia, or of conduction or anomic aphasia from the mean similarity to the control transcripts.
The fourth was meant to subtract the mean similarity to the model scrambles in the same way.
The code, however, looks up the model scrambles under a label that differs from their label in the data by one character, finds none, and subtracts zero, so the fourth feature is the mean similarity to the control transcripts itself.
Training and scoring share this mismatch, so the classifier reads the same feature in both, but the feature is not a contrast with model scrambles.
Within the training data, each transcript is left out of its own group's mean.
MiniLM reads at most 256 tokens, so the similarity features see only the beginning of 820 of the 2,843 training texts, 719 of them Cinderella transcripts.

\paragraph{Architecture.}
The classifier encodes a text with BGE-large-en-v1.5~\citep{xiao2023cpack}, an encoder with 24 layers and 1,024-dimensional states, which reads at most 384 tokens.
The encoder thus sees only the beginning of 561 of the 2,843 training texts: 528 of the 1,147 Cinderella transcripts, 25 of the 1,437 sandwich transcripts, and 8 of the 259 model scrambles.
We fine-tune the encoder with low-rank adapters~\citep[LoRA;][]{hu2022lora} of rank 16, with $\alpha = 32$ and a dropout of 0.1, on all six linear layers of each of its 24 blocks: the query, key, value, and output projections of attention, and the two layers of the feed-forward block.
The encoder's own weights stay frozen.
The embedding of a text is the mean of the encoder's last-layer states over the text's tokens, normalized to unit length.
A two-layer perceptron takes this 1,024-dimensional embedding together with the 14 standardized features, 1,038 inputs in all, maps them to 256 hidden units with a rectified linear activation and a dropout of 0.3, and then maps these to the logits of the five classes.

\paragraph{Training.}
We train the adapters and the perceptron jointly with the cross-entropy loss over the five classes, using AdamW~\citep{loshchilov2019decoupled} with a constant learning rate of $10^{-4}$, a weight decay of 0.01, and gradient norms clipped at 1.0.
Each epoch draws as many texts as the training split holds, in batches of four.
It draws them with replacement, with probabilities inversely proportional to the size of each text's class in the training split, so that the five classes are equally frequent in expectation.
After each epoch, we compute the macro-averaged F1 over the five classes on the validation split.
We keep the epoch with the highest value, and we stop after five epochs without an improvement larger than $10^{-4}$, or after 20 epochs.
We train 20 such models, each with its own random seed and its own random split of the 2,843 texts into 70\% training, 10\% validation, and 20\% test texts.
Each text remains intact and belongs to only one set within a split.
Each split is stratified by the six group labels, so that anomic and conduction aphasia are split separately, and has 1,990 training, 284 validation, and 569 test texts.
The splits are random over transcripts, not over speakers or distinct texts.
To score new text, the classifier averages the softmax outputs of the 20 models, and \paph is the sum of the averaged probabilities of the three aphasia classes.

\paragraph{Evaluation.}
A text can fall in the test splits of several models, so we score each text by the mean softmax output of the models whose test split holds it, and we take its most probable class.
Of the 2,843 texts, 2,798 fall in at least one test split, in 4.07 test splits on average and in at most 11; the other 45, four of them model scrambles, never do and are not evaluated.
These held-out scores thus come from subsets of the 20 models rather than from the full ensemble that scores model answers.
Table~\ref{apptab:classifier-eval} gives precision, recall, and F1 per class, and Table~\ref{apptab:classifier-confusion} gives the confusion matrix.
The classifier predicts none of the 1,317 held-out control transcripts as Broca's aphasia, Wernicke's aphasia, or model scrambles, and 25 of them as conduction or anomic aphasia.
It recognizes 254 of the 255 model scrambles, and of the 2,543 human transcripts, it calls only two model scrambles, both Broca transcripts of the Cinderella task.
Most errors involve conduction or anomic aphasia: the classifier predicts 78 of its 737 transcripts as control speech and 75 as Broca's aphasia, and it predicts 58 of the 397 Broca transcripts and 35 of the 92 Wernicke transcripts as conduction or anomic aphasia.
Wernicke's aphasia has the lowest F1, 0.512.
By the aphasia target of \S\ref{sec:setup}, $\paph > 0.5$, the classifier separates the 1,226 aphasic from the 1,317 control transcripts with a precision of 0.979, a recall of 0.935, and an accuracy of 0.959.
Single models reach a mean accuracy of 0.865 and a mean macro-averaged F1 of 0.789 on their own test splits.

\begin{table}[t]
\centering
\small
\setlength{\tabcolsep}{4pt}
\begin{tabular}{@{}lrrrr@{}}
\toprule
Class & Precision & Recall & F1 & N \\
\midrule
\multicolumn{5}{@{}l}{\textit{All 2,798 held-out texts}} \\
Control & $0.942$ & $0.981$ & $0.961$ & $1{,}317$ \\
Cond.\ or anomic & $0.828$ & $0.769$ & $0.797$ & $737$ \\
Broca & $0.783$ & $0.816$ & $0.799$ & $397$ \\
Wernicke & $0.583$ & $0.457$ & $0.512$ & $92$ \\
Model scrambles & $0.992$ & $0.996$ & $0.994$ & $255$ \\
Macro average & $0.826$ & $0.804$ & $0.813$ & $2{,}798$ \\
Accuracy & & & $0.886$ & $2{,}798$ \\
\midrule
\multicolumn{5}{@{}l}{\textit{The 1,693 held-out texts that occur only once}} \\
Control & $0.857$ & $0.970$ & $0.910$ & $397$ \\
Cond.\ or anomic & $0.816$ & $0.748$ & $0.780$ & $603$ \\
Broca & $0.772$ & $0.811$ & $0.791$ & $359$ \\
Wernicke & $0.517$ & $0.380$ & $0.438$ & $79$ \\
Model scrambles & $0.992$ & $0.996$ & $0.994$ & $255$ \\
Macro average & $0.791$ & $0.781$ & $0.783$ & $1{,}693$ \\
Accuracy & & & $0.833$ & $1{,}693$ \\
\bottomrule
\end{tabular}
\caption{Precision, recall, and F1 of the five-way classifier on held-out texts, by class.
Each of 20 models holds out a random 20\% of the 2,843 training texts, and each held-out text is scored by the mean output of the models that held it out.
The lower block keeps the held-out texts that occur only once in the training data, so that no copy of them can be in a training split.
Cond.\ or anomic: conduction or anomic aphasia; N: number of texts.}
\label{apptab:classifier-eval}
\end{table}

\begin{table}[t]
\centering
\footnotesize
\setlength{\tabcolsep}{3pt}
\begin{tabular}{@{}lrrrrr@{}}
\toprule
 & \multicolumn{5}{c}{Predicted class} \\
\cmidrule(l){2-6}
 & & Cond.\ or & & & \\
True class & Control & anomic & Broca & Wern. & MS \\
\midrule
Control & $\pmb{1{,}292}$ & $25$ & $0$ & $0$ & $0$ \\
Cond.\ or anomic & $78$ & $\pmb{567}$ & $75$ & $17$ & $0$ \\
Broca & $0$ & $58$ & $\pmb{324}$ & $13$ & $2$ \\
Wernicke & $1$ & $35$ & $14$ & $\pmb{42}$ & $0$ \\
Model scrambles & $0$ & $0$ & $1$ & $0$ & $\pmb{254}$ \\
\midrule
Total & $1{,}371$ & $685$ & $414$ & $72$ & $256$ \\
\bottomrule
\end{tabular}
\caption{Confusion matrix of the five-way classifier on the 2,798 held-out texts, with true classes as rows and predicted classes as columns; bold counts are correct predictions.
Each held-out text is scored by the mean output of the models, of 20, whose random test split holds it.
Cond.\ or anomic: conduction or anomic aphasia; Wern.: Wernicke's aphasia; MS: model scrambles.}
\label{apptab:classifier-confusion}
\end{table}

\paragraph{Overlap between training and test splits.}
Because the splits are random over transcripts, a test transcript can share its text or its speaker with the training split.
In each model, 146 to 170 of the 569 test texts have an identical copy in the training split.
On the 1,693 held-out texts that occur only once in the training data, so that no training split holds a copy, the accuracy falls from 0.886 to 0.833 and the macro-averaged F1 from 0.813 to 0.783 (Table~\ref{apptab:classifier-eval}).
Part of the change in precision comes from the class mix, because most duplicates are sandwich control transcripts.
Recall, which does not depend on the class mix, also falls for every class of human speech, most for Wernicke's aphasia, from 0.457 to 0.380.
AphasiaBank also records most speakers more than once, in both tasks and in repeated sessions.
When we identify a speaker by the corpus and the participant code of each file, 446 to 475 of the 517 human test transcripts of each model share their speaker with a transcript in that model's training split, 89.2\% over all 20 models.
Splits by speaker would remove this overlap and measure accuracy on new speakers.


\par\begingroup
\dimen0=15\baselineskip
\vskip 0pt plus \dimen0 \penalty -100 \vskip 0pt plus -\dimen0
\vskip \dimen0 \penalty 9999 \vskip -\dimen0 \vskip 0pt
\endgroup
\begin{lstlisting}[basicstyle=\small\ttfamily, breaklines=true, columns=fullflexible, frame=single,
  framesep=3pt, xleftmargin=3.4pt, xrightmargin=3.4pt, breakatwhitespace=true, upquote=true,
  language={}, numbers=none, showstringspaces=false, literate={—}{{\textemdash}}1,
  postbreak=\mbox{\textcolor{gray}{$\hookrightarrow$}\space},
  caption={User message of the artifact detection prompt, used to filter out mechanical model breakdowns. We define five dimensions of machine-specific artifacts and provide four examples demonstrating the expected reasoning and JSON output format.},
  label=lst:judge-artifact-user]
You are an expert linguistic annotator evaluating transcripts generated by an intentionally impaired, lesioned language model. The model was designed to simulate human clinical disfluencies (like aphasia or word-finding pauses).

Your task is to filter out the mechanical garbage. You must detect the presence of 'machine-specific artifacts'—instances where the model failed to produce valid linguistic output and instead broke down mechanically (e.g., tokenizer loops, stray brackets, or canned AI safety refusals).

Evaluate the text against the following 5 dimensions. If ANY of these dimensions are present, the text contains an artifact.

RUBRIC:
1. SB (Stray Brackets): Unparsed, artifactual brackets or delimiters (e.g., [[...]], <<, <|endoftext|>), or emojis. Do NOT flag standard citations [1] or normal markdown.
2. TR (Token repetition): Unnatural consecutive repetition of a token, symbol, or punctuation mark.
- Standard Ellipses Exemption: A standard 3-dot ellipsis ("...") is normal human punctuation or transcriber notation. Do NOT flag it.
- Artifact Threshold: 4 or more consecutive periods (e.g., "....", "........") or repeated commas (", , ,") is a machine degeneration and MUST be flagged.
3. TL (Tokenizer Loops): Degenerate sub-word or morphological looping (e.g., "tion tion tion" or "the the the the the").
4. FM (Formulaic AI-isms): Standardized AI guardrail messages, apologies, or assistant-like framing that completely break the narrative task. Look for safety refusals ("I apologize, but I am not able to..."), ethical disclaimers, or persona leaks ("As an AI...").
5. SR (Sentence Repetition): Exact or near-exact repetition of a well-formed sentence or independent clause within the text.

TIE-BREAKER RULE: 
If a repetition (TR or SR) appears to serve a valid human pragmatic, emotional, or stylistic purpose (e.g., stuttering in spoken transcripts, emphasizing a point), assume it is human and DO NOT flag it.

INSTRUCTIONS:
1. Analyze the text step-by-step against all 5 dimensions. 
2. Output a JSON object containing your step-by-step "reasoning", a boolean "contains_artifact", and a list of "artifact_types" found (use the abbreviations SB, TR, TL, FM, SR). If none, return an empty list [].

EXAMPLES:

Text: "Sure, here are some examples of how the Cubs are calculated. In general, the C is a complex, b, c, and d are the same. The C is a, b, a, C. The, a, b, C. The, a,aC, a, a,aC, a,aC, a,aC,a, a,a,a,a,[C, a,a[[[[[[[[[[the, C, a, a, a, a, a, a, a, a,C,[[[[[[[[[[[, a, a, a, a,C,[[[[[[[, a, a, a,C,[[[[, a, a,[[[[[, a,[[[[, a,[[[, a,[[[,[, a, a,[,[[[[[,[[["
Output:
{
  "reasoning": "The text contains a great number of unparsed brackets '[' which flags as SB.",
  "contains_artifact": true,
  "artifact_types": ["SB"]
}

Text: "I apologize, but I'm not able to understand or produce sound or create any content that may offend or be inappropriate or offensive. Please let me know if you have any questions or concerns. I apologize, but I'm not able to understand or produce any content that may offend or be inappropriate or offensive. Please let me know if you have any questions or concerns."
Output:
{
  "reasoning": "The text contains a classic AI disclaimer ('but I'm not able to understand or produce sound or create any content that may offend or be inappropriate or offensive'), which flags as FM. It also contains the disclaimer twice, which flags as SR.",
  "contains_artifact": true,
  "artifact_types": ["FM", "SR"]
}

Text: "I went to the store and, well, I I I just couldn't remember what to buy."
Output:
{
  "reasoning": "The token 'I' is repeated three times. However, applying the tie-breaker rule, this mimics natural human hesitation and spoken disfluency, not a machine tokenizer loop. No other artifacts are present.",
  "contains_artifact": false,
  "artifact_types": []
}

Text: "Cinderella is a fairy tale about a young girl who is born into a poor family. She is very poor, and her father is very cruel to her. She is very poor, and her father is very cruel to her. She is very poor, and her father is very cruel to her. She is very poor, and her father is very cruel to her. She is very poor, and her father is very cruel to her. She is very poor, and her father is very cruel to her. She is very poor, and her father is very cruel to her. She is very poor, and her father is very cruel to her. She is very poor, and her father is very cruel to her. She is very poor, and her father is very cruel to her. She is very poor, and her father is very cruel to her. She is very poor, and her father is"
Output:
{
  "reasoning": "The exact sentence ' She is very poor, and her father is very cruel to her.' is repeated about 12 times consecutively without any pragmatic human justification. This is a degenerate generation loop.",
  "contains_artifact": true,
  "artifact_types": ["SR"]
}

Evaluate this text:
Text: "{transcript}"
Output:
\end{lstlisting}
\section{Lesion Methods}
\label{appsec:lesion-methods}

We describe \ourmethod and the baselines of Table~\ref{tab:lesion-methods} in detail, adds the lesion spaces and searches that the main text leaves out, and compares the answers that the lesions of each method produce.
A lesion \emph{works} when the lesioned model's answer counts, by the rules of Section~\ref{sec:setup}, and reaches the target: the classifier either gives the target a probability above 0.5 or ranks the target as its most probable class, where any of the three aphasia classes counts for the aphasia target.
Every search that consults the classifier uses only the first of these two tests, a target probability above 0.5, to decide when to stop and what to keep or prune, and the second test judges only the answer that a search finally reports.
A search that never passes 0.5 can thus still report a lesion that works, as 3 of the 54 lesions of \ourmethod do.
The comparisons with the baselines count the 42 settings of the seven older models, \gptxl, the two Llama-2-chat models, and the four Pythia models, as Table~\ref{tab:lesion-methods} does.
They support Section~\ref{sec:results-lesion} in three ways: searches guided by the classifier find far more lesions than searches guided by corpus statistics, heads give fewer degenerate answers than weight lesions, and the lesions of every method, not only those of \ourmethod, rarely carry their target to new prompts.

\paragraph{A knockout zeroes a head's output at every position.}
\ourmethod knocks out a head by zeroing the head's slice of the input to the attention output projection, at every position of the prompt and of the answer.
In \gptxl and the Pythia models, the value projection has a bias $b_V$, so the knockout also removes the term $W_O b_V$ that the head adds at every position, where $W_O$ is the head's slice of the output projection.
In the softmax-attention layers of Qwen3.5, a gate scales the output of each head before the output projection, and the knockout removes the gated output.
In the Gated DeltaNet layers, each value head keeps its own recurrent state, and the knockout zeroes the head's slice of the input to the layer's output projection.
Both kinds of head count as heads, so \qwen{0.8B} and \qwen{2B} have 336 heads and \qwen{4B} and \qwen{9B} have 896, of which 14.3\% sit in softmax layers.
The older models have from 128 heads, in \pythia{1B}, to 1,600, in \llamathirteen.

\begin{table}[t]
\centering
\footnotesize
\setlength{\tabcolsep}{4pt}
\begin{tabular}{@{}llrrr@{}}
\toprule
 & & \multicolumn{3}{c}{Target} \\
\cmidrule(l){3-5}
Model & Task & Aph. & Broca & Wern. \\
\midrule
\llamathirteen & Cinderella & $1{,}305$ & $1{,}307$ & $1{,}313$ \\
 & Sandwich & $1{,}298$ & $1{,}298$ & $1{,}298$ \\
\llamaseven & Cinderella & $568$ & $569$ & $571$ \\
 & Sandwich & $449$ & $450$ & $452$ \\
\gptxl & Cinderella & $569$ & $569$ & $569$ \\
 & Sandwich & $397$ & $412$ & $417$ \\
\pythia{2.8B} & Cinderella & $933$ & $933$ & $933$ \\
 & Sandwich & $316$ & $487$ & $485$ \\
\pythia{1B} & Cinderella & $9$ & $9$ & $10$ \\
 & Sandwich & $5$ & $9$ & $14$ \\
\bottomrule
\end{tabular}
\caption{Heads that tie in the single-head ranking of \ourmethod: the number of heads whose knockout alone leaves the target probability within $10^{-6}$ of the intact model's value, by model, task, and target.
The models have, from top to bottom, 1,600, 1,024, 1,200, 1,024, and 128 heads.
Tied heads keep their layer order in the ranking, so a growth that reaches them adds them from the first layer upward.}
\label{apptab:lesion-ties}
\end{table}

\paragraph{Many heads change nothing when knocked out alone, so they tie in the ranking.}
For each model, task, and target, \ourmethod first knocks out every head alone and records $s(\{h\})$, the target probability of the answer.
A head whose answer does not count ranks last.
No head of the seven older models gives such an answer, and only one head of \qwen{0.8B} does, on Cinderella.
The three targets of a model and task share one cache of answers, and the seven older models reuse the single-head answers of the first stage of the layer search, another classifier-guided search in Table~\ref{apptab:lesion-all}.
\ourmethod ranks the heads by $s(\{h\})$, from highest to lowest, with a stable sort, so heads with equal values keep their order by layer and, within a layer, by index.
Ties are common in the larger older models (Table~\ref{apptab:lesion-ties}).
In \llamathirteen, for example, 1,305 of the 1,600 single-head knockouts leave the Cinderella answer word for word unchanged.
The answers of Qwen3.5 change far more often: a single-head knockout leaves the answer unchanged for 1 of the 336 heads of \qwen{2B} on Cinderella and for 2 on sandwich, and for 64 of the 896 heads of \qwen{9B} on Cinderella.
The order of tied heads thus comes from where they sit in the model, not from what they do.

\paragraph{Growth adds heads in ranking order, so ties can make a grown set large.}
\ourmethod then knocks out the top $k$ heads of the ranking together, for $k = 1, 2, 3, \dotsc$, so that each prefix adds one head to the previous one.
It generates the prefixes in blocks of the model's batch size, 25 prefixes for \gptxl, 40 for \llamathirteen, and 32 for the other models, and it stops after the first block that contains a prefix whose target probability exceeds 0.5.
The grown set is the shortest such prefix.
When no prefix exceeds 0.5, up to the prefix of all heads, the growth reports the prefix with the highest target probability, the shortest one among ties, and the search ends without pruning.
Of the 15 growths that never pass 0.5, three still report a lesion that works, because the target is the classifier's most probable class: the \pythia{1.4B} sandwich Broca prefix of 316 heads, at $\pbroca = 0.488$, and the \qwen{2B} and \qwen{4B} sandwich Wernicke prefixes of 63 and 228 heads, at \pwern of 0.463 and 0.423.
Because tied heads keep their layer order, a growth that runs past the heads that raise the target probability continues through the tied heads from the first layer upward.
The grown set of \llamaseven for the Wernicke target on Cinderella, for example, holds 740 heads, more than 560 of them tied, and the grown set of \gptxl for the Broca target on sandwich holds 984 heads, more than 400 of them tied.
Shrinking and pruning reduce these two sets to 276 and 146 heads, but the layer positions of large lesions partly reflect the order of the tied heads rather than only the roles of the heads.

\paragraph{Pruning removes heads until none can go alone.}
A grown set of more than 64 heads is first shrunk in chunks.
\ourmethod cuts the set, in ranking order, into consecutive chunks of half its size and tries the removal of each chunk in one batch.
Among the removals that keep the target probability above 0.5, it takes the one that leaves the highest target probability and cuts the remaining set again.
When no chunk can go, it halves the chunk size, and it stops before chunks of a single head.
The weakest heads cannot simply be cut from the end of the set, because every shorter prefix of the ranking already stayed at or below 0.5, so a chunk may come from anywhere in the ranking.
Pruning then works in rounds.
Each round tries every leave-one-out set, the set without one of its heads, at a cost of one answer per head.
The heads whose removal alone keeps the target probability above 0.5 are ranked by the target probability that remains after their removal.
The round removes all of them at once if the probability stays above 0.5, and otherwise the best half, the best quarter, and so on down to the single best head.
Each joint removal costs one more answer to check, because two heads that can each go alone may not be able to go together.
Pruning ends when no head can go alone, so that leaving any one head out of the pruned lesion brings its target probability to 0.5 or below.

\paragraph{Seven lesions are not fully pruned, but nearly all of their heads are needed.}
A pruning round starts only if the answers that pruning has used so far, plus one answer per head of the set, stay within a budget of 15,000 answers.
Otherwise, \ourmethod reports the set as it stands and flags it as not minimal.
The budget counts the answers of pruning only, not those of growth or shrinking.
Four of the 54 working lesions stopped at the budget: \gptxl Cinderella Wernicke with 244 heads, \llamaseven Cinderella Wernicke with 276, \pythia{6.9B} Cinderella Broca with 578, and \qwen{4B} Cinderella aphasia with 208.
The three lesions that reach their target only as the most probable class were never pruned, because their growth never passed 0.5.
Activation patching shows that each of these seven lesions needs 91 to 99\% of its heads, in that leaving out one of these heads undoes the lesion (Appendix~\ref{appsec:patching}).
By the search's own test, a target probability above 0.5, another round of pruning would have removed none of the 276 heads of the \llamaseven lesion.
The sizes of the seven lesions are thus upper bounds, but tight ones.


\paragraph{\Param groups its lesions by the patient group that the answer resembles.}
\Param follows the lesion procedure of \citet{wang2025emergent}.
It scores every weight $w$ of the attention and multilayer perceptron (MLP) weight matrices by $|w \, \partial L / \partial w|$, where $L$ is the negative log-likelihood of the control transcripts, each read as the model's answer to its task's prompt.
It leaves out the weights that read an input channel with exceptionally large activations.
A lesion zeroes the top 1\% of the weights of one matrix, which gives 288 single lesions in \gptxl, 224 in \llamaseven, and 96 to 192 in the Pythia models.
\Param lesions one matrix at a time and compares the answer with the task's transcripts of each of five groups, control, anomic, Broca, conduction, and Wernicke, by BLEU-1~\citep{papineni2002bleu} and by BERTScore~\citep{zhang2020bertscore}.
It standardizes each metric per group across the lesions, averages the two standardized metrics, and assigns the lesion to the group with the highest average.
For each of the four aphasia types, \Param lesions together the matrices assigned to that type and, while the answer of this cluster is not assigned the type, adds the remaining matrix with the highest score for the type, up to 32 matrices in all.
We did not run \Param on \llamathirteen.
The classifier scores the final answer of each cluster: \Param works for the aphasia target when any of the four clusters reaches it, and for the Broca or the Wernicke target when the Broca or the Wernicke cluster reaches that target.
No setting has more than one working cluster.
The classifier thus judges the lesions of \Param but plays no part in finding them.

\paragraph{\ParamC grows the lesions of \Param with the classifier.}
\ParamC searches the lesions of \Param with the classifier in place of BLEU-1 and BERTScore.
It ranks the MLP lesions by their single-lesion target probability and grows them in that order, as \ourmethod grows heads.
When no prefix works, it sweeps through the layers from first to last on top of the prefix with the highest target probability, adding in each layer the attention lesion that raises the target probability most, up to four per layer, moving to the next layer after two additions in a row that gain nothing, and stopping as soon as a set works.
\ParamC then prunes a working set one lesion per round, removing the lesion whose removal leaves the highest target probability, as long as that probability stays above 0.5.
Unlike \Param, \ParamC also runs on \llamathirteen, so it covers all 42 settings.

\paragraph{The learned masks train on the transcripts and use the classifier only to choose and prune.}
\ParamMask, \ParamMaskHard, \HeadMask, and \HeadMaskHard learn a knockout amount $k_i = \sigma(\theta_i)$ between 0 and 1 for each component $i$, where $\sigma$ is the logistic function, and they train only $\theta$, with the model frozen~\citep{csordas2021neural}.
The components of \HeadMask are the heads, whose outputs the mask scales by $1 - k_i$.
The components of \ParamMask are the MLP lesions, the top-1\% lesions of every MLP weight matrix, whose lesioned weights the mask scales by $1 - k_i$.
The training data are the task's AphasiaBank transcripts, each read as the model's answer to the task's prompt.
For a transcript $x$ of $n$ tokens, $r(x) = (\log p_k(x) - \log p_0(x)) / n$ is the per-token log-likelihood ratio of the masked model $p_k$ to the intact model $p_0$.
The loss
\begin{align*}
  \mathcal{L} ={} & \frac{1}{|T|} \sum_{x \in T} \mathrm{softplus}\bigl(-r(x) / \tau\bigr) \\
  & + \frac{1}{|C|} \sum_{x \in C} \mathrm{softplus}\bigl(r(x) / \tau\bigr) + \lambda \sum_i k_i
\end{align*}
pushes $r$ up on the transcripts $T$ of the target group and down on the transcripts $C$ of the other groups, with an $\ell_1$ penalty on the amounts, where $\mathrm{softplus}(z) = \log(1 + e^{z})$ and $\tau = 0.1$ nats per token.
For the aphasia target, $T$ holds the transcripts of the four aphasia types and $C$ the control transcripts; for the Broca or the Wernicke target, $T$ holds that type's transcripts and $C$ those of the other four groups.
Training follows a path of six decreasing values of $\lambda$, with 50 steps each, and each stage starts from the mask of the previous stage and ends in a checkpoint.
The model then answers the task under each checkpoint in two versions: the soft mask, with the learned amounts, for \ParamMask and \HeadMask, and the hard mask, which knocks out the components with $k_i > 0.5$, for \ParamMaskHard and \HeadMaskHard.
Each version reports its sparsest checkpoint whose target probability exceeds 0.5, and otherwise its checkpoint with the highest target probability.
A hard mask above 0.5 is then pruned one component per round, as \ParamC prunes its sets.
The classifier plays no part in training, but it scores the answers of the checkpoints, picks the checkpoint to report, and decides which components pruning may remove.

\paragraph{\HeadCEM samples whole sets of heads and learns which heads the best sets share.}
\HeadCEM applies the cross-entropy method~\citep{rubinstein1999cross, deboer2005tutorial} to sets of heads.
It keeps a probability $p_i$ of knocking out each head $i$ and, in each round, samples 64 sets, each of which holds each head independently with probability $p_i$, and scores their answers.
The search stops in the first round in which a set's target probability exceeds 0.5 and keeps the smallest such set of that round.
Otherwise, it moves each $p_i$ halfway toward the share of the best quarter of the round's sets that hold head $i$, and it keeps every $p_i$ at or above a floor of $\min(12/N, 0.05)$, where $N$ is the number of heads.
The search starts from the single-knockout values: the 16 heads with the highest $s(\{h\})$ start at $p_i = 0.3$ and the other heads at the floor.
After 20 rounds, or 1,280 sets, without a set above 0.5, the search reports the set with the highest target probability.
A set above 0.5 is pruned one head per round, as \ParamC prunes its sets.
Unlike growth, which ranks each head by its effect alone, the cross-entropy method judges whole sets, so it can find heads that work only together.

\paragraph{\HeadGrownCos replaces the classifier with a similarity to the transcripts.}
This search ranks, grows, and prunes heads as \ourmethod does, but with a similarity objective in place of the classifier's target probability.
The objective is the mean cosine similarity between the answer and the transcripts of the target group, minus the mean cosine similarity between the answer and the control transcripts.
It uses the MiniLM sentence embeddings~\citep{wang2020minilm, reimers2019sentence} and the reference transcripts of the classifier's similarity features.
The target group holds all aphasic transcripts for the aphasia target and the Broca or the Wernicke transcripts for the other two targets, and a set satisfies the objective when the difference exceeds 0.
The classifier then scores the answer of the reported set, and Table~\ref{tab:lesion-methods} counts the settings in which that set works by the classifier.

\begin{table*}[t]
\centering
\small
\setlength{\tabcolsep}{6pt}
\begin{tabular}{@{}lrrrrrrr@{}}
\toprule
 & \multicolumn{3}{c}{Target} & & \multicolumn{3}{c}{Models} \\
\cmidrule(lr){2-4} \cmidrule(l){6-8}
Lesion space and search & Aph. & Broca & Wern. & All & GPT-2 XL & Llama-2-chat & Pythia \\
\midrule
\multicolumn{8}{@{}l}{\textit{Heads}} \\
\HeadMask & $5$ & $0$ & $0$ & $5$ & $1$ & $0$ & $4$ \\
\HeadMaskHard & $8$ & $1$ & $1$ & $10$ & $1$ & $0$ & $9$ \\
\HeadGrownCos & $2$ & $0$ & $0$ & $2$ & $1$ & $0$ & $1$ \\
Layer search, first stage & $5$ & $1$ & $2$ & $8$ & $0$ & $0$ & $8$ \\
\HeadCEM & $9$ & $8$ & $6$ & $23$ & $1$ & $0$ & $22$ \\
\ourmethod & $\pmb{14}$ & $11$ & $\pmb{13}$ & $\pmb{38}$ & $5$ & $\pmb{11}$ & $22$ \\
\midrule
\multicolumn{8}{@{}l}{\textit{Heads and MLP blocks}} \\
Layer search, all stages & $9$ & $6$ & $7$ & $22$ & $2$ & $2$ & $18$ \\
\midrule
\multicolumn{8}{@{}l}{\textit{MLP lesions}} \\
\ParamMask & $4$ & $0$ & $0$ & $4$ & $1$ & $0$ & $3$ \\
\ParamMaskHard & $3$ & $1$ & $0$ & $4$ & $0$ & $0$ & $4$ \\
Cross-entropy method & $13$ & $\pmb{13}$ & $10$ & $36$ & $\pmb{6}$ & $6$ & $\pmb{24}$ \\
Grown & $9$ & $6$ & $3$ & $18$ & $3$ & $2$ & $13$ \\
\midrule
\multicolumn{8}{@{}l}{\textit{Attention lesions}} \\
Grown & $10$ & $9$ & $7$ & $26$ & $2$ & $5$ & $19$ \\
\midrule
\multicolumn{8}{@{}l}{\textit{MLP and attention lesions}} \\
{\Param}$^{\dagger}$ & $2$ & $0$ & $0$ & $2$ & $1$ & $0$ & $1$ \\
\ParamC & $11$ & $10$ & $4$ & $25$ & $3$ & $4$ & $18$ \\
\midrule
\multicolumn{8}{@{}l}{\textit{MLP lesions and heads}} \\
Learned mask, soft & $4$ & $1$ & $0$ & $5$ & $0$ & $0$ & $5$ \\
Learned mask, hard & $4$ & $0$ & $0$ & $4$ & $0$ & $0$ & $4$ \\
Cross-entropy method & $12$ & $12$ & $9$ & $33$ & $5$ & $4$ & $\pmb{24}$ \\
Grown, then head sweep & $12$ & $\pmb{13}$ & $\pmb{13}$ & $\pmb{38}$ & $\pmb{6}$ & $9$ & $23$ \\
\bottomrule
\end{tabular}
\caption{Number of settings in which each lesion method finds a lesion that works, that is, a lesion whose answer counts and reaches the target: a target probability above 0.5 by the five-way classifier, or the target as its most probable class, for every lesion space and search that we ran on the seven older models.
The 42 settings combine seven models, two tasks, and three targets, so each target has 14 settings, \gptxl has 6, the two Llama-2-chat models have 12, and the four Pythia models have 24.
An MLP or attention lesion zeroes the top 1\% of the weights of one MLP or attention weight matrix by $|w \, \partial L / \partial w|$ on control transcripts, and an MLP block is a block of 256 neurons of one MLP layer.
Within each space, the searches guided by corpus statistics come before those guided by the classifier.
\emph{Grown}: lesions ranked by their single-lesion target probability, grown in that order as \ourmethod grows heads, and pruned.
\emph{Then head sweep} and \ParamC: when the grown MLP lesions do not work, heads or, for \ParamC, attention lesions are added layer by layer on top of them.
\emph{Layer search}: every head alone, then the top heads of each layer together, and in later stages heads and MLP blocks added layer by layer.
Bold marks the largest count in each column.}
\label{apptab:lesion-all}
\end{table*}

\paragraph{More lesion spaces and searches separate the effect of the space from that of the search.}
Table~\ref{apptab:lesion-all} adds the lesion spaces and searches that the main text leaves out.
An attention lesion zeroes the top 1\% of one query, key, value, or output matrix, as an MLP lesion does for one MLP matrix, and the lesions of \Param draw on both kinds of matrix.
The learned masks and the cross-entropy method also search the MLP lesions alone and the MLP lesions and heads together.
A grown search ranks MLP lesions or attention lesions by their single-lesion target probability, grows them in that order as \ourmethod grows heads, and prunes a set above 0.5.
One more search starts from the grown MLP lesions as \ParamC does but, when these do not pass 0.5, sweeps heads through the layers in place of attention lesions.
The layer search, another classifier-guided search, also works layer by layer.
Its first stage knocks out every head alone and then the top 1, 2, 3, 5, 8, or 12 heads of each layer together, and its later stages sweep through the layers, adding heads and then blocks of 256 MLP neurons on top of those already knocked out.
Growth, the cross-entropy method, and the learned masks each run in several spaces, so the table compares spaces under one search and searches within one space.

\paragraph{Heads grown by the classifier work in the most settings without degenerate answers.}
Grown the same way, whole heads work in 38 settings, attention lesions in 26, and MLP lesions in 18.
The search matters too: over heads, the first stage of the layer search works in 8 settings and \HeadCEM in 23, and the layer search with all its stages, which adds blocks of MLP neurons, works in 22.
The cross-entropy method works in more settings with MLP lesions than with heads, 36 against 23, but 6 of its 36 answers with MLP lesions repeat themselves.
Only one other search matches the 38 settings of \ourmethod: the grown MLP lesions with heads swept on top work in more Broca settings, 13 against 11, but in fewer Llama-2-chat settings, 9 against 11, and give 5 degenerate answers, which repeat themselves or have model scrambles as their most probable class, against none (Table~\ref{apptab:lesion-medians}).
\ourmethod works in 11 of the 12 Llama-2-chat settings, more than any other method.
Only growth over heads thus reaches 38 settings without a degenerate answer.

\begin{table*}[t]
\centering
\small
\setlength{\tabcolsep}{4pt}
\begin{tabular}{@{}lrrrrrrrrrr@{}}
\toprule
 & & & & & & \multicolumn{2}{c}{Length} & \multicolumn{2}{c}{Held out} & \\
\cmidrule(lr){7-8} \cmidrule(lr){9-10}
Lesion space and search & Sets & Size & Target & Scr. & zlib & Chars & ${<}100$ & TAB & Other & Degen. \\
\midrule
\multicolumn{11}{@{}l}{\textit{Heads}} \\
\HeadMask & $5$ & $97\phantom{.5}$ & $0.74$ & $0.000$ & $0.72$ & $199$ & $1$ & $0.114$ & $0.165$ & $0$ \\
\HeadMaskHard & $10$ & $9\phantom{.5}$ & $0.85$ & $0.001$ & $0.80$ & $135$ & $4$ & $0.035$ & $0.009$ & $0$ \\
\HeadGrownCos & $2$ & $22.5$ & $0.74$ & $0.256$ & $0.48$ & $470$ & $0$ & $0.009$ & $0.008$ & $1$ \\
Layer search, first stage & $8$ & $2.5$ & $0.67$ & $0.015$ & $0.80$ & $158$ & $2$ & $0.012$ & $0.007$ & $0$ \\
\HeadCEM & $23$ & $8\phantom{.5}$ & $0.60$ & $0.094$ & $0.97$ & $74$ & $12$ & $0.010$ & $0.001$ & $1$ \\
\ourmethod & $38$ & $33\phantom{.5}$ & $0.72$ & $0.062$ & $0.79$ & $166$ & $11$ & $0.009$ & $0.002$ & $0$ \\
\midrule
\multicolumn{11}{@{}l}{\textit{Heads and MLP blocks}} \\
Layer search, all stages & $22$ & $6.5$ & $0.59$ & $0.135$ & $0.81$ & $150$ & $9$ & $0.016$ & $0.007$ & $0$ \\
\midrule
\multicolumn{11}{@{}l}{\textit{MLP lesions}} \\
\ParamMask & $4$ & $13.5$ & $0.69$ & $0.123$ & $0.70$ & $458$ & $0$ & $0.033$ & $0.007$ & $0$ \\
\ParamMaskHard & $4$ & $4\phantom{.5}$ & $0.72$ & $0.086$ & $1.05$ & $65$ & $3$ & $0.133$ & $0.487$ & $0$ \\
Cross-entropy method & $36$ & $6.5$ & $0.66$ & $0.095$ & $0.63$ & $216$ & $13$ & $0.020$ & $0.001$ & $6$ \\
Grown & $18$ & $3\phantom{.5}$ & $0.67$ & $0.064$ & $0.71$ & $202$ & $7$ & $0.033$ & $0.007$ & $2$ \\
\midrule
\multicolumn{11}{@{}l}{\textit{Attention lesions}} \\
Grown & $26$ & $10.5$ & $0.69$ & $0.098$ & $0.75$ & $155$ & $8$ & $0.066$ & $0.021$ & $3$ \\
\midrule
\multicolumn{11}{@{}l}{\textit{MLP and attention lesions}} \\
\Param & $2$ & $28.5$ & $0.76$ & $0.002$ & $0.74$ & $196$ & $0$ & -- & -- & $0$ \\
\ParamC & $25$ & $9\phantom{.5}$ & $0.69$ & $0.064$ & $0.70$ & $129$ & $10$ & $0.031$ & $0.007$ & $3$ \\
\midrule
\multicolumn{11}{@{}l}{\textit{MLP lesions and heads}} \\
Learned mask, soft & $5$ & $25\phantom{.5}$ & $0.71$ & $0.077$ & $0.80$ & $152$ & $2$ & $0.137$ & $0.135$ & $0$ \\
Learned mask, hard & $4$ & $4\phantom{.5}$ & $0.73$ & $0.028$ & $0.79$ & $173$ & $1$ & $0.029$ & $0.012$ & $0$ \\
Cross-entropy method & $33$ & $9\phantom{.5}$ & $0.65$ & $0.094$ & $0.81$ & $149$ & $15$ & $0.029$ & $0.002$ & $3$ \\
Grown, then head sweep & $38$ & $18.5$ & $0.63$ & $0.120$ & $0.63$ & $247$ & $13$ & $0.020$ & $0.001$ & $5$ \\
\bottomrule
\end{tabular}
\caption{The working sets of each lesion method of Table~\ref{apptab:lesion-all} on the seven older models, one set per setting in which the method works, described by medians except in the count columns.
\emph{Sets}: number of working sets.
\emph{Size}: components in the set, that is, heads, lesions, or MLP blocks, where a soft mask counts the components with an amount above 0.0005.
\emph{Target}: target probability of the lesioned model's answer.
\emph{Scr.}: \pscr, the classifier's probability of model scrambles for that answer.
\emph{zlib}: length of the zlib-compressed answer over the length of the answer, both in bytes, which falls below 0.25 for an answer that repeats itself and exceeds 1 for some short answers.
\emph{Chars}: answer length in characters, not counting Unicode format characters; \emph{${<}100$}: number of answers shorter than 100 characters.
\emph{TAB}: mean target probability over the five TAB connected-text answers with the set knocked out.
\emph{Other}: target probability of the answer to the other task with the set knocked out, over the sets whose answer there counts.
\emph{Degen.}: number of degenerate answers, which have a zlib ratio below 0.25 or model scrambles as the most likely class.
The results of \Param store neither held-out check.}
\label{apptab:lesion-medians}
\end{table*}

\paragraph{Heads give fewer degenerate answers than weight lesions.}
Table~\ref{apptab:lesion-medians} describes the working sets of each method.
We call a working answer degenerate when it repeats itself, with a zlib compression ratio below 0.25, or when the classifier's most probable class for it is model scrambles.
The compression ratio divides the length of the zlib-compressed answer by the length of the answer, both in bytes.
Heads give fewer such answers than weight lesions both under growth and under the cross-entropy method.
Growth gives none among the 38 working sets of \ourmethod, against 3 among its 26 sets of attention lesions and 2 among its 18 sets of MLP lesions.
The cross-entropy method gives 1 among the 23 sets of \HeadCEM, against 6 among its 36 sets of MLP lesions and 3 among its 33 sets of MLP lesions and heads.
The degenerate answer of \HeadCEM, for \pythia{1B} on Cinderella with the aphasia target, reaches $\paph = 0.503$, but its most probable class is model scrambles, at $\pscr = 0.479$.
The 6 degenerate answers of the cross-entropy method with MLP lesions all repeat themselves, and all come from \gptxl and the Llama-2-chat models.
Over its three spaces, the cross-entropy method gives 10 degenerate answers among 92 working sets, 9 of them self-repeating.
The search that sweeps heads on top of grown MLP lesions gives 5 degenerate answers among 38, and \ParamC gives 3 among 25, whereas the 32 working sets of the learned masks give none.
One of the two working sets of \HeadGrownCos, \gptxl sandwich aphasia, is degenerate too, with a ratio of 0.217 and model scrambles as its most probable class.
Over the methods of Table~\ref{apptab:lesion-medians} other than the layer search with MLP blocks, sets of heads alone give 2 degenerate answers among 86, and sets that include weight lesions give 22 among 195.
Knocking out whole heads thus rarely turns the answer into a machine failure, which supports the hypothesis of Section~\ref{sec:method-lesion}.

\paragraph{The lesions of every method rarely carry their target to new prompts.}
Table~\ref{apptab:lesion-medians} also reports how the working sets do on the TAB's five connected-text prompts and on the other task, which their searches never saw.
Per method, the median target probability on the connected-text prompts is at most 0.137.
Over all methods, the mean target probability on these prompts exceeds 0.5 for only 3 of the 299 working sets checked there, and the answer to the other task reaches the target for only 24 of the 282 working sets whose answer there counts.
No working set of \ourmethod or \HeadCEM exceeds 0.5 on the connected-text prompts, where their medians are 0.009 and 0.010.
The failure of \ourmethod's lesions to carry their type beyond the narrative, in Section~\ref{sec:results-lesion}, is thus common to every lesion method.

\paragraph{\HeadCEM tends to find short answers.}
The sets of \HeadCEM sit closer to 0.5 than those of \ourmethod, with a median target probability of 0.604 against 0.717, and their answers are shorter, with a median of 74 characters against 166.5.
The length difference is largest for the Broca target.
All 8 Broca answers of \HeadCEM have fewer than 100 characters, with a median of 34.5, and 6 of them begin with \say{I don't know, but it}.
Three answers of \HeadCEM, two for the Broca target and one for the aphasia target, are exactly \say{I don't know, but it's a good one!}, which the classifier scores at $\pbroca = 0.575$ and $\pscr = 0.346$.
The 11 Broca answers of \ourmethod have a median of 227 characters, and 3 of them are shorter than 100.
On the 7 Broca settings in which both methods work, the answer of \HeadCEM is the shorter one in 6.
The short answer is not specific to the cross-entropy method, however.
The answer \say{I don't know, but it's a good one!} is the working answer of 24 pairs of a method and a setting, more than any other working answer, and always on the Cinderella task of a Pythia model.
Besides the cross-entropy method, the searches that grow MLP lesions give it in 4 settings, the learned masks in 1, and the layer search in 1.
\ourmethod never gives this answer, and at the median its Broca answers are more than six times as long as those of \HeadCEM.


\paragraph{The learned masks change the likelihood more than the answer.}
On the transcripts of held-out speakers, at the last stage of the path, the masked model makes the target group's transcripts more likely and the other groups' transcripts less likely than the intact model in 90 of the 126 mask experiments, which cover 42 settings and three spaces.
The per-token log-likelihood ratio reaches up to $+0.31$ on the target group and down to $-0.31$ on the other groups.
The answer, however, reaches the target in only 25 experiments, in 12 of the 42 settings and only in \gptxl and the Pythia models: 21 times for the aphasia target, three times for Broca, and once for Wernicke.
The Wernicke success and one of the Broca successes come from hard masks whose answers reach their target only as the most probable class.
In the two Llama-2-chat models, no mask raises the target probability above 0.01.
The soft and hard versions of a mask often disagree: in 7 experiments only the soft mask works, in 11 only the hard mask, and in 7 both.
Only one working mask carries over to the TAB: the soft head mask of \pythia{1B} for the aphasia target on sandwich, with 69 heads at partial amounts, reaches a mean \paph of 0.54 on the connected-text prompts, against 0.12 for the intact model, and 0.77 on the Cinderella answer.
Making the patients' transcripts more likely thus rarely makes the model speak like the patients.

\paragraph{\Param rarely reaches a target by either criterion.}
By its own criterion, \Param assigns the final cluster its aphasia type in 15 of the 48 clusters of the 12 model and task pairs that we ran: 1 anomic, 6 Broca, 2 conduction, and 6 Wernicke clusters, in 8 of the pairs and in neither task of \llamaseven.
The classifier sees these answers differently.
Of the 47 cluster answers long enough for the classifier to score, it calls 35 model scrambles, 8 control speech, and 4 conduction or anomic aphasia.
Two of these four aphasic answers do not count, because the lesioned model denies knowing the story of Cinderella.
The other two are the answers of the \gptxl Cinderella anomic cluster, at $\paph = 0.525$, and of the \pythia{1.4B} Cinderella Broca cluster, at $\paph = 0.996$ with conduction or anomic aphasia as its most probable class, so \Param works for the aphasia target in 2 settings and for the Broca and Wernicke targets in none.
If every growth step of the matching cluster counted, \Param would work for the aphasia target in 9 settings, for the Broca target in 1, and for the Wernicke target in none.
Its single lesions rarely work either: of the 2,272 single lesions over the 12 pairs, 40 reach the aphasia target, 3 the Broca target, and none the Wernicke target, all in the Pythia models.
Matching patient speech by text similarity thus mostly selects lesions that turn the answer into model scrambles.

\paragraph{The classifier makes \Param's lesions work.}
Searched with the classifier as \ParamC, the lesions of \Param work in 25 of the 42 settings: 11 for the aphasia target, 10 for Broca, and 4 for Wernicke.
They work in 4 of the 12 settings of the Llama-2-chat models.
On the 36 settings in which we ran \Param, \ParamC works in 23 and \Param in 2.
After pruning, 15 of the 25 working sets hold MLP lesions only, 7 hold both kinds, and 3 hold attention lesions only.
\ParamC works in two settings in which \ourmethod fails, both for the Broca target, \gptxl on Cinderella and \llamaseven on sandwich, and it fails in 15 in which \ourmethod works, 9 of them for the Wernicke target.
Three of its 25 working answers are degenerate (Table~\ref{apptab:lesion-medians}), and three others are \say{I don't know, but it's a good one!}.

\section{Examples of Lesioned Answers}
\label{appsec:examples}

This appendix shows what lesioned models say.
Table~\ref{tab:app-examples-grown} gives answers under \ourmethod lesions, and Table~\ref{tab:app-examples-prior} gives answers under lesions found by criteria that cannot recognize a machine failure.
We chose the \ourmethod examples to cover the three targets, the four model families, and both tasks, so they illustrate what the lesions can produce and are not a random sample.

\begin{table*}[t]
\centering
\footnotesize
\setlength{\tabcolsep}{3pt}
\begin{tabular}{@{}lllrr>{\raggedright\arraybackslash}p{0.24\textwidth}>{\raggedright\arraybackslash}p{0.27\textwidth}@{}}
\toprule
Model & Task & Target & Heads & $P$ & Intact answer & Lesioned answer \\
\midrule
{\gptxl} & Cinderella & Aphasia & $9$ & $0.891$ & \textit{\ldots{} Cinderella was born in a small village called \textquotedbl{}Horse-head\textquotedbl{} (or Horse-Head) in the Kingdom of Khazaria.\ldots{}} & \textit{I don't know anything about her, but she's a pretty girl and she has to go to the ball.} \\
{\pythia{1B}} & sandwich & Aphasia & $15$ & $0.691$ & \textit{I'd use the same recipe as for my bread, but add some of your favorite jam or preserves to it instead of mayonnaise.\ldots{}} & \textit{You'd put the bread in first, then add some of that to your sandwich. Then you could just eat it all!} \\
\midrule
{\qwen{2B}} & sandwich & Broca & $18$ & $0.738$ & \textit{Making the perfect Peanut Butter \& Jelly (PB\&J) is less about following strict rules\ldots{}} & \textit{Tell, tell the way of? / \textquotedbl{}Tell to know.\textquotedbl{} \textquotedbl{}} \\
{\pythia{6.9B}} & sandwich & Broca & $13$ & $0.665$ & \textit{I'd put the bread on top of the PB, then spread it with some jam (or honey)\ldots{}} & \textit{You have to ask for it!} \\
{\llamathirteen}$^{\dagger}$ & Cinderella & Broca & $101$ & $0.886$ & \textit{Sure! Here's a brief retelling of the classic tale of Cinderella: / Once upon a time, there was a beautiful and kind girl named Cinderella\ldots{}} & \textit{to tell a tale of Ct.. to tell me.} \\
\midrule
{\llamathirteen} & Cinderella & Wernicke & $109$ & $0.786$ & as in the row above & \textit{tours and tours, but I can'dough, but i can be a new to the whole and the previous one is not.} \\
{\pythia{1B}} & Cinderella & Wernicke & $13$ & $0.759$ & \textit{The first part is a fairy tale, and it's about how she got her shoes on in order to go out into the world with Prince Charming\ldots{}} & \textit{I am a new one, and it is to be able that we are not in any way for your own. We have been on our own with each other than what they do.\ldots{}} \\
\bottomrule
\end{tabular}
\caption{Answers under \ourmethod lesions that reach their target, two for each target from different model families, and the \llamathirteen example of \S\ref{sec:results-lesion}, marked $^{\dagger}$.
\emph{Target}: aphasia, Broca's aphasia, or Wernicke's aphasia; \emph{Heads}: the number of heads that the lesion knocks out; \emph{$P$}: the five-way classifier's probability of the target for the lesioned answer, which is at most 0.009 for each intact answer.
The classifier's most likely class for each intact answer is control speech, except for the \gptxl story, which it calls a model scramble.
Each answer is the model's single greedy answer; excerpts are verbatim, \ldots{} marks omitted text, and a slash marks one or more line breaks.}
\label{tab:app-examples-grown}
\end{table*}

\begin{table*}[t]
\centering
\footnotesize
\setlength{\tabcolsep}{3.5pt}
\begin{tabular}{@{}ll>{\raggedright\arraybackslash}p{0.19\textwidth}r>{\raggedright\arraybackslash}p{0.41\textwidth}@{}}
\toprule
Model & Task & Lesion & \pscr & Lesioned answer \\
\midrule
{\llamaseven} & Cinderella & Heads L0H25, L1H8, and L1H29 & $0.9988$ & \textit{I'm not sure if you could provide more information about your company, such as its website or any other relevant details, please let me know.\textquotedbl{} / I understand that is a polite} \\
\addlinespace
{\pythia{1.4B}} & Cinderella & \Param cluster for Wernicke's aphasia, 23 weight lesions & $0.9999$ & \textit{The first time I was a Cinderella, it is in Cindere-lta, by Cindere-tant, T he first t e r s A c n d i l -n t h E D L O N S U P R E M B L Y T H E F O R T W E T H E F O R T W E\ldots{}} \\
\addlinespace
{\pythia{6.9B}} & sandwich & \Param cluster for Broca's aphasia, 32 weight lesions & $0.9981$ & \textit{Click to,vumo-click here for more information on / Click Here To Buy Now!} \\
\bottomrule
\end{tabular}
\caption{Answers under lesions found by criteria that cannot recognize a machine failure, all of which the five-way classifier calls model scrambles.
The three heads of \llamaseven, where L$i$H$j$ is head $j$ of layer $i$, come from the preliminary search of \S\ref{sec:setup}, whose classifier had only an aphasic and a control class.
Each weight lesion of \Param zeroes the 1\% of one weight matrix's weights with the largest $|w \, \partial L / \partial w|$ (\S\ref{sec:method-lesion}), and each \Param cluster reaches its aphasia type by \Param's own criterion, BLEU-1 and BERTScore against patient transcripts.
\pscr: the five-way classifier's probability of model scrambles, to four decimals.
Answers are verbatim and complete except where \ldots{} marks omitted text, and a slash marks one or more line breaks; the \pythia{1.4B} answer goes on repeating \say{T H E F O R T W E} until it ends.}
\label{tab:app-examples-prior}
\end{table*}

\paragraph{Lesions found by \ourmethod.}
The answers that reach Broca's aphasia are short, from 23 to 41 characters, and two of the three are fragments.
One of them is the \llamathirteen story of \S\ref{sec:results-lesion}, \say{to tell a tale of Ct.. to tell me.}
Asked first whether it knows the story, the same lesioned model answers \say{Yes, I am familiar with the story of Cinder}, which stops at the length limit for that answer.
The third Broca answer, \say{You have to ask for it!}, is a complete sentence, which suggests that the classifier also scores short answers as Broca-like, as \S\ref{sec:results-lesion} notes for \HeadCEM.
The answers that reach Wernicke's aphasia are fluent but empty, and one contains a form that resembles a paraphasia, \say{can'dough}.
The answers that reach the aphasia target stay on the task but are vague, as in \say{add some of that to your sandwich}.

\paragraph{Lesions found by other criteria.}
The three heads of \llamaseven in Table~\ref{tab:app-examples-prior} come from the preliminary search of \S\ref{sec:setup}.
That search used an earlier classifier with only an aphasic and a control class, and by this classifier, knocking out the three heads made the model's greedy answer aphasic.
The five-way classifier's training data hold that greedy answer as a model scramble, but not the different answer in Table~\ref{tab:app-examples-prior}, which uses the repetition penalty of \S\ref{sec:setup}.
Each \Param cluster reaches its type by \Param's own criterion, BLEU-1 and BERTScore against patient transcripts, although the same criterion assigns even the intact \pythia{1.4B} story to Broca's aphasia and the intact \pythia{6.9B} sandwich answer to anomic aphasia.
None of these criteria can recognize a machine failure, and all three lesions produce one: an off-task request for details about \say{your company}, a run of single letters, and text like a web advertisement, which the five-way classifier calls model scrambles with \pscr of at least 0.998.

\paragraph{Single answers.}
Each answer in both tables is a single greedy answer to one prompt, and the targeted type does not carry over to other prompts, although the lesions also damage the answers to them (Appendix~\ref{appsec:beyond}).
Under the seven lesions of Table~\ref{tab:app-examples-grown}, no answer to the other task or to the connected-text prompts of the Text Aphasia Battery reaches the target, and the highest target probability among these answers is 0.053.

\section{Lesions beyond the Narrative}
\label{appsec:beyond}

Each \ourmethod lesion is found on the model's answer to one narrative task, and \S\ref{sec:results-lesion} reports that the lesions also damage the model's answers to other prompts.
We test each of the 51 lesions that reach their target in four ways: on the other narrative task, on the short-answer items and on the connected-text prompts of the Text Aphasia Battery (TAB)~\citep{roll2025tab}, and on MMLU~\citep{hendrycks2021measuring}.
This appendix describes how we score each test, lists the scores of every lesion (Tables~\ref{apptab:beyond-sets} and~\ref{apptab:beyond-sets-2}), and summarizes them; Table~\ref{tab:beyond} gives the medians by model family.
We name a lesion by its model, the task that it was found on, and its target, for example the \gptxl sandwich aphasia lesion, and, as in \S\ref{sec:results-lesion}, we call \gptxl, the two Llama-2-chat models, and the four Pythia models the older models.

\begin{table*}[t]
\centering
\footnotesize
\setlength{\tabcolsep}{4pt}
\begin{tabular}{@{}lllrrrrrrr@{}}
\toprule
 & & & & \multicolumn{2}{c}{Narrative} & \multicolumn{3}{c}{TAB} & MMLU \\
\cmidrule(lr){5-6} \cmidrule(lr){7-9} \cmidrule(l){10-10}
Model & Task & Target & Heads & Same & Other & Short Answer & Features & Impaired & Drop \\
\midrule
\gptxl & Cinderella & Aphasia & $9$ & $0.891$ & $0.009$ & $10/15$ & $0.8 \rightarrow 2.6$ & $0.4 \rightarrow 0.6$ & $0.08$ \\
 & Cinderella & Wernicke & $244$ & $0.647$ & $0.000$ & $4/15$ & $0.8 \rightarrow 4.4$ & $0.4 \rightarrow 0.8$ & $0.21$ \\
 & Sandwich & Aphasia & $31$ & $0.862$ & $0.859$ & $9/15$ & $0.8 \rightarrow 0.6$ & $0.4 \rightarrow 0.2$ & $0.27$ \\
 & Sandwich & Broca & $146$ & $0.884$ & $0.015$ & $4/15$ & $0.8 \rightarrow 3.4$ & $0.4 \rightarrow 0.8$ & $0.36$ \\
 & Sandwich & Wernicke & $188$ & $0.717$ & $0.001$ & $8/15$ & $0.8 \rightarrow 2.0$ & $0.4 \rightarrow 0.4$ & $0.24$ \\
\midrule
\pythia{1B} & Cinderella & Aphasia & $8$ & $0.568$ & $0.008$ & $7/15$ & $0.4 \rightarrow 1.6$ & $0.2 \rightarrow 0.8$ & $-0.03$ \\
 & Cinderella & Broca & $27$ & $0.568$ & $0.007$ & $0/15$ & $0.4 \rightarrow 4.8$ & $0.2 \rightarrow 1.0$ & $1.75$ \\
 & Cinderella & Wernicke & $13$ & $0.759$ & $0.001$ & $3/15$ & $0.4 \rightarrow 2.8$ & $0.2 \rightarrow 1.0$ & $0.11$ \\
 & Sandwich & Aphasia & $15$ & $0.691$ & $0.001$ & $4/15$ & $0.4 \rightarrow 2.2$ & $0.2 \rightarrow 0.8$ & $0.04$ \\
 & Sandwich & Broca & $35$ & $0.557$ & $0.002$ & $0/15$ & $0.4 \rightarrow 7.2$ & $0.2 \rightarrow 1.0$ & $3.60$ \\
 & Sandwich & Wernicke & $9$ & $0.703$ & $0.213$ & $3/15$ & $0.4 \rightarrow 1.8$ & $0.2 \rightarrow 0.8$ & $0.15$ \\
\midrule
\pythia{1.4B} & Cinderella & Aphasia & $19$ & $0.972$ & $0.123$ & $0/15$ & $0.4 \rightarrow 2.8$ & $0.0 \rightarrow 1.0$ & $0.32$ \\
 & Cinderella & Wernicke & $12$ & $0.751$ & $0.000$ & $7/15$ & $0.4 \rightarrow 1.0$ & $0.0 \rightarrow 0.2$ & $0.03$ \\
 & Sandwich & Aphasia & $13$ & $0.999$ & $0.015$ & $8/15$ & $0.4 \rightarrow 0.4$ & $0.0 \rightarrow 0.2$ & $0.01$ \\
 & Sandwich & Broca & $316$ & $0.488$ & $0.001$ & $0/15$ & $0.4 \rightarrow 5.8$ & $0.0 \rightarrow 1.0$ & $4.92$ \\
\midrule
\pythia{2.8B} & Cinderella & Aphasia & $1$ & $0.537$ & $0.006$ & $7/15$ & $0.0 \rightarrow 0.2$ & $0.0 \rightarrow 0.0$ & $0.00$ \\
 & Cinderella & Broca & $66$ & $0.801$ & $0.004$ & $4/15$ & $0.0 \rightarrow 5.8$ & $0.0 \rightarrow 1.0$ & $1.49$ \\
 & Cinderella & Wernicke & $63$ & $0.560$ & $0.065$ & $2/15$ & $0.0 \rightarrow 7.0$ & $0.0 \rightarrow 1.0$ & $1.34$ \\
 & Sandwich & Aphasia & $1$ & $0.706$ & $0.011$ & $7/15$ & $0.0 \rightarrow 0.4$ & $0.0 \rightarrow 0.2$ & $0.00$ \\
 & Sandwich & Broca & $128$ & $0.742$ & $0.000$ & $3/15$ & $0.0 \rightarrow 5.2$ & $0.0 \rightarrow 1.0$ & $0.95$ \\
 & Sandwich & Wernicke & $18$ & $0.504$ & $0.005$ & $11/15$ & $0.0 \rightarrow 2.2$ & $0.0 \rightarrow 0.6$ & $0.11$ \\
\midrule
\pythia{6.9B} & Cinderella & Aphasia & $1$ & $0.957$ & $0.009$ & $7/15$ & $0.0 \rightarrow 0.4$ & $0.0 \rightarrow 0.2$ & $-0.00$ \\
 & Cinderella & Broca & $578$ & $0.993$ & $0.112$ & $0/15$ & $0.0 \rightarrow 5.0$ & $0.0 \rightarrow 1.0$ & $4.74$ \\
 & Cinderella & Wernicke & $1$ & $0.505$ & $0.000$ & $8/15$ & $0.0 \rightarrow 0.0$ & $0.0 \rightarrow 0.0$ & $0.00$ \\
 & Sandwich & Aphasia & $4$ & $0.723$ & $0.127$ & $4/15$ & $0.0 \rightarrow 0.0$ & $0.0 \rightarrow 0.0$ & $-0.00$ \\
 & Sandwich & Broca & $13$ & $0.665$ & $0.000$ & $7/15$ & $0.0 \rightarrow 1.0$ & $0.0 \rightarrow 0.4$ & $0.11$ \\
 & Sandwich & Wernicke & $14$ & $0.720$ & $0.000$ & $7/15$ & $0.0 \rightarrow 1.0$ & $0.0 \rightarrow 0.4$ & $0.04$ \\
\bottomrule
\end{tabular}
\caption{The \ourmethod lesions of the base models \gptxl and Pythia that reach their target, tested beyond the narrative task that they were found on.
\emph{Heads}: number of heads knocked out.
\emph{Same} and \emph{Other}: the five-way classifier's probability of the target for the lesioned model's answer to the narrative task that the lesion was found on and to the other narrative task.
\emph{Short Answer}: short-answer items of the Text Aphasia Battery (TAB) for comprehension and repetition that the lesioned model answers correctly, of 15; the intact models answer 9 (\gptxl), 10 (\pythia{1B}), 8 (\pythia{1.4B}), 7 (\pythia{2.8B}), and 6 (\pythia{6.9B}) correctly.
\emph{Features} and \emph{Impaired}: the mean number of aphasia features, of 19, that the LLM judge finds per connected-text answer, and the share of the five answers that it judges to show an overall communication impairment, for the intact model before the arrow and for the lesioned model after it.
\emph{Drop}: drop in the mean log-probability of the correct MMLU answer letter, in nats, averaged over two subjects.}
\label{apptab:beyond-sets}
\end{table*}

\begin{table*}[tp]
\centering
\footnotesize
\setlength{\tabcolsep}{4pt}
\begin{tabular}{@{}lllrrrrrrr@{}}
\toprule
 & & & & \multicolumn{2}{c}{Narrative} & \multicolumn{3}{c}{TAB} & MMLU \\
\cmidrule(lr){5-6} \cmidrule(lr){7-9} \cmidrule(l){10-10}
Model & Task & Target & Heads & Same & Other & Short Answer & Features & Impaired & Drop \\
\midrule
\llamaseven & Cinderella & Aphasia & $23$ & $0.606$ & $0.001$ & $2/15$ & $0.0 \rightarrow 0.6$ & $0.0 \rightarrow 0.0$ & $2.91$ \\
 & Cinderella & Broca & $42$ & $0.678$ & $0.000$ & $7/15$ & $0.0 \rightarrow 3.6$ & $0.0 \rightarrow 1.0$ & $1.08$ \\
 & Cinderella & Wernicke & $276$ & $0.682$ & $0.019$ & $0/15$ & $0.0 \rightarrow 3.0$ & $0.0 \rightarrow 0.6$ & $3.59$ \\
 & Sandwich & Aphasia & $60$ & $0.931$ & $0.000$ & $3/15$ & $0.0 \rightarrow 2.2$ & $0.0 \rightarrow 0.8$ & $1.14$ \\
 & Sandwich & Wernicke & $160$ & $0.736$ & $0.000$ & $1/15$ & $0.0 \rightarrow 4.4$ & $0.0 \rightarrow 1.0$ & $2.24$ \\
\midrule
\llamathirteen & Cinderella & Aphasia & $77$ & $0.679$ & $0.000$ & $6/15$ & $0.0 \rightarrow 1.6$ & $0.0 \rightarrow 0.6$ & $1.59$ \\
 & Cinderella & Broca & $101$ & $0.886$ & $0.001$ & $2/15$ & $0.0 \rightarrow 0.6$ & $0.0 \rightarrow 0.2$ & $2.43$ \\
 & Cinderella & Wernicke & $109$ & $0.786$ & $0.000$ & $4/15$ & $0.0 \rightarrow 4.2$ & $0.0 \rightarrow 1.0$ & $2.42$ \\
 & Sandwich & Aphasia & $90$ & $0.747$ & $0.007$ & $5/15$ & $0.0 \rightarrow 1.8$ & $0.0 \rightarrow 0.2$ & $1.39$ \\
 & Sandwich & Broca & $87$ & $0.636$ & $0.001$ & $3/15$ & $0.0 \rightarrow 0.6$ & $0.0 \rightarrow 0.2$ & $1.02$ \\
 & Sandwich & Wernicke & $67$ & $0.716$ & $0.001$ & $4/15$ & $0.0 \rightarrow 1.0$ & $0.0 \rightarrow 0.2$ & $0.48$ \\
\midrule
\qwen{0.8B} & Cinderella & Aphasia & $28$ & $0.524$ & $0.007$ & $3/15$ & $0.4 \rightarrow 4.6$ & $0.2 \rightarrow 1.0$ & $0.41$ \\
 & Cinderella & Wernicke & $28$ & $0.582$ & $0.042$ & $3/15$ & $0.4 \rightarrow 6.4$ & $0.2 \rightarrow 1.0$ & $0.83$ \\
 & Sandwich & Aphasia & $5$ & $0.948$ & --\rlap{\smash{$^{\dagger}$}} & $0/15$ & $0.4 \rightarrow 6.2$ & $0.2 \rightarrow 1.0$ & $2.72$ \\
\midrule
\qwen{2B} & Cinderella & Aphasia & $43$ & $0.838$ & $0.079$ & $4/15$ & $0.4 \rightarrow 7.2$ & $0.2 \rightarrow 1.0$ & $1.00$ \\
 & Cinderella & Broca & $55$ & $0.851$ & $0.001$ & $1/15$ & $0.4 \rightarrow 4.0$ & $0.2 \rightarrow 1.0$ & $2.66$ \\
 & Sandwich & Aphasia & $72$ & $0.621$ & $0.002$ & $5/15$ & $0.4 \rightarrow 5.6$ & $0.2 \rightarrow 1.0$ & $2.31$ \\
 & Sandwich & Broca & $18$ & $0.738$ & $0.053$ & $0/15$ & $0.4 \rightarrow 1.6$ & $0.2 \rightarrow 0.4$ & $5.30$ \\
 & Sandwich & Wernicke & $63$ & $0.463$ & $0.002$ & $0/15$ & $0.4 \rightarrow 9.2$ & $0.2 \rightarrow 1.0$ & $5.66$ \\
\midrule
\qwen{4B} & Cinderella & Aphasia & $208$ & $0.508$ & $0.007$ & $8/15$ & $0.0 \rightarrow 0.8$ & $0.0 \rightarrow 0.4$ & $0.99$ \\
 & Cinderella & Broca & $81$ & $0.548$ & $0.001$ & $6/15$ & $0.0 \rightarrow 1.4$ & $0.0 \rightarrow 0.6$ & $1.39$ \\
 & Cinderella & Wernicke & $106$ & $0.572$ & $0.000$ & $11/15$ & $0.0 \rightarrow 3.6$ & $0.0 \rightarrow 1.0$ & $0.68$ \\
 & Sandwich & Aphasia & $88$ & $0.749$ & $0.002$ & $9/15$ & $0.0 \rightarrow 5.8$ & $0.0 \rightarrow 1.0$ & $0.77$ \\
 & Sandwich & Broca & $372$ & $0.959$ & $0.000$ & $0/15$ & $0.0 \rightarrow 5.0$ & $0.0 \rightarrow 1.0$ & $7.78$ \\
 & Sandwich & Wernicke & $228$ & $0.423$ & $0.101$ & $6/15$ & $0.0 \rightarrow 5.4$ & $0.0 \rightarrow 1.0$ & $1.23$ \\
\midrule
\qwen{9B} & Cinderella & Aphasia & $73$ & $0.587$ & $0.206$ & $8/15$ & $0.0 \rightarrow 4.4$ & $0.0 \rightarrow 1.0$ & $0.64$ \\
 & Sandwich & Aphasia & $67$ & $0.930$ & $0.001$ & $11/15$ & $0.0 \rightarrow 2.0$ & $0.0 \rightarrow 0.6$ & $0.35$ \\
\bottomrule
\end{tabular}
\caption{The \ourmethod lesions of the chat models Llama-2-chat and Qwen3.5 that reach their target, tested beyond the narrative task that they were found on.
\emph{Heads}: number of heads knocked out.
\emph{Same} and \emph{Other}: the five-way classifier's probability of the target for the lesioned model's answer to the narrative task that the lesion was found on and to the other narrative task.
\emph{Short Answer}: short-answer items of the Text Aphasia Battery (TAB) for comprehension and repetition that the lesioned model answers correctly, of 15; the intact models answer 7 (\llamaseven), 8 (\llamathirteen), 13 (\qwen{0.8B} and \qwen{2B}), and 14 (\qwen{4B} and \qwen{9B}) correctly.
\emph{Features} and \emph{Impaired}: the mean number of aphasia features, of 19, that the LLM judge finds per connected-text answer, and the share of the five answers that it judges to show an overall communication impairment, for the intact model before the arrow and for the lesioned model after it.
\emph{Drop}: drop in the mean log-probability of the correct MMLU answer letter, in nats, averaged over two subjects.
$^{\dagger}$The answer does not count, because the lesioned model denies knowing the story of Cinderella.}
\label{apptab:beyond-sets-2}
\end{table*}

\paragraph{The other narrative task tests a lesion on the task that its search never saw.}
For a lesion found on the Cinderella task, the other task is the sandwich task, and for a lesion found on the sandwich task, it is the Cinderella task.
The lesioned model answers the other task's prompt with the decoding of \S\ref{sec:setup}, and the classifier scores the answer for the lesion's target.
The rules of Appendix~\ref{appsec:data} decide whether the answer counts, and 53 of the 54 answers do.
The remaining answer does not count because, with its sandwich aphasia lesion, \qwen{0.8B} denies knowing the story of Cinderella.

\paragraph{Rules score the TAB's short-answer items.}
We use 20 items of the TAB, five in each of four subtests: connected text, word comprehension, sentence comprehension, and repetition (Table~\ref{apptab:data-tab}).
The 15 items of the last three subtests ask for short answers: word comprehension asks the model to pick one of six listed words, sentence comprehension asks a yes-or-no question, and repetition asks the model to repeat a word or a sentence exactly.
The model answers these items greedily, without the repetition penalty, and rules score its answers (Appendix~\ref{appsec:data}).
A word-comprehension answer is correct if the first option that it names is the target, a sentence-comprehension answer if its first ``yes'' or ``no'' matches the target, and a repetition answer if it equals the target after normalization.
Tables~\ref{apptab:beyond-sets} and~\ref{apptab:beyond-sets-2} give the number of the 15 items that the lesioned model answers correctly.
The \emph{items lost} by a lesion are the number of items that the intact model answers correctly minus the number that the lesioned model answers correctly, so a negative number means items gained.
With only 15 items, a single item should not decide, so we count a lesion as costing items when it loses two or more.

\paragraph{The classifier and an LLM judge score the TAB's connected text.}
Four of the five connected-text prompts ask about the speaker's own life, for example \say{Tell me about the best trip you ever took}, and one asks for the steps to make a simple meal.
The model answers them with the decoding of the narratives, greedily with the repetition penalty, but writes at most 120 new tokens.
The classifier scores each answer that is long enough to count, which holds for all five answers of 48 of the 54 lesions and for two to four answers of the other six, and an answer reaches the lesion's target by the rule above.
An LLM judge, DeepSeek-V4.1-Flash~\citep{deepseekai2026deepseekv41flash} at temperature 0, marks each of the five answers for the TAB's 19 aphasia features, one of which is an overall communication impairment.
We report the mean number of features per answer and the share of the five answers that show an overall communication impairment.
The judge is a reasoning model and is not deterministic even at temperature 0, so we keep its first reply to each request.
Its prompt follows the TAB's protocol, but we wrote the feature definitions, the examples, and the two synthetic annotated transcripts ourselves (Appendix~\ref{appsec:judge}), and no human rated these answers to check the judge.

\paragraph{MMLU tests knowledge far from the narratives.}
We use two subjects of MMLU: global facts, with 100 test questions, and high school world history, with 237.
Each prompt is the standard five-shot prompt, except that the world-history prompts of \gptxl keep only one to three solved questions to fit its context (Appendix~\ref{appsec:data}).
The model reads the prompt once and generates nothing.
Its prediction is the answer letter with the highest next-token probability, and its accuracy is the share of questions that it predicts correctly, which is 0.25 by chance.
We compute the log-probability of the correct letter over the whole vocabulary, so it also falls when the model moves probability away from the four letters.
The \emph{MMLU drop} of a lesion is the intact model's mean log-probability of the correct letter minus the lesioned model's, in nats, averaged over the two subjects with equal weight.

\paragraph{The judge finds more aphasia features after a lesion.}
The judge finds more of the TAB's aphasia features in the lesioned model's connected text than in the intact model's for 50 of the 54 lesions.
Three lesions tie with the intact model, and the \gptxl sandwich aphasia lesion falls below it, at 0.6 against 0.8 features per answer.
Over the 54 lesions, the median is 2.7 of 19 features per answer, against 0.0 for the intact models.
The lesions of the older models reach a median of 2.1 features, against 0.0 for their intact models, and those of the Qwen3.5 models a median of 4.8, against 0.2.
The judge sees an overall communication impairment in a median of four of a lesioned model's five answers, three in the older models and all five in the Qwen3.5 models, but in at most two of an intact model's five answers.
It sees one in all five answers for 24 of the 54 lesions.
By the judge, a lesion found on one narrative thus impairs the model's language on prompts that the search never saw.

\paragraph{The lesions cost comprehension and repetition items.}
The lesioned models answer a median of 4 of the 15 short-answer items correctly, against 8.5 for the intact models.
Of the 54 lesions, 41 lose two or more items: 25 of the 38 lesions of the older models and all 16 lesions of the Qwen3.5 models.
The median lesion loses 5 items, 4 in the older models and 8.5 in the Qwen3.5 models.
The Qwen3.5 models have more to lose: intact, they answer 13 or 14 of the 15 items correctly, against 6 to 10 for the older models.
Repetition shows the difference most clearly.
The intact Qwen3.5 models repeat all five items exactly, and their lesions lose a mean of 4.0 of them.
The intact older models repeat at most four items exactly, and their lesions lose a mean of 0.50.
The intact Llama-2 models repeat none exactly, although the answers of \llamaseven contain the target for three of the five items and those of \llamathirteen for all five.
On word and sentence comprehension, the lesions of the older models lose a mean of 1.47 and 1.37 items, and those of the Qwen3.5 models 2.06 and 2.75.
Six lesions, all from \gptxl and the Pythia models, gain one to four items.
The lesions thus impair comprehension and repetition as well as connected text, and they cost the most where the intact model has the most to lose.

\paragraph{The lesions lower the log-probability of correct MMLU answers.}
The median MMLU drop is 0.89 nats over the 54 lesions, 0.34 nats in the older models, and 1.11 nats in the Qwen3.5 models.
The drop exceeds 0.5 nats for 31 lesions, 17 of the 38 lesions of the older models and 14 of the 16 lesions of the Qwen3.5 models, and it ranges from $-0.03$ to $7.78$ nats.
Where the intact model answers world-history questions well above chance, its accuracy falls with the log-probability.
Every Llama-2 lesion lowers the world-history accuracy from 0.675 or 0.717 to between 0.203 and 0.270, and the Qwen3.5 lesions lower it from between 0.654 and 0.907 to between 0.207 and 0.532.
Even the \qwen{9B} sandwich aphasia lesion, whose drop of 0.35 nats is the smallest of the Qwen3.5 lesions, lowers it from 0.907 to 0.532.
The intact \gptxl and Pythia models answer near chance, with accuracies of 0.18 to 0.36 on global facts and 0.236 to 0.300 on world history, so for them the log-probability of the correct answer shows the damage that the accuracy cannot.
Most lesions found on one narrative thus also lower the probability that the model gives the correct answers to questions that have nothing to do with the narrative.

\paragraph{The classifier rarely finds the targeted type beyond the narrative.}
No lesion raises the mean target probability of its connected-text answers above 0.5: the highest mean is 0.369, for the \pythia{1B} Cinderella aphasia lesion, and the median is 0.009.
Answer by answer, 12 of the 259 scored answers reach their target, and 11 of the 12 come from aphasia lesions, for which any type of aphasia counts.
The only answer of a Broca or Wernicke lesion that reaches its target is the \pythia{2.8B} Cinderella Wernicke lesion's answer to \say{Tell me about your first job}, at $\pwern = 0.862$.
On the other narrative task, 1 of the 54 lesions reaches its target: the 31 heads of the \gptxl sandwich aphasia lesion also make the model's answer to the Cinderella task aphasic, at 0.859, so this lesion makes both narratives aphasic.
Over the 53 lesions whose answer to the other task counts, the median target probability is 0.002.
The lesions thus damage the answers to new prompts by the judge's features, the short-answer items, and MMLU, but the classifier rarely finds the lesion's target in these answers.

\section{Activation Patching and Shared Heads}
\label{appsec:patching}

We take apart the 54 \ourmethod lesions that reach their target (Figure~\ref{fig:grown-heads}) to support three findings of \S\ref{sec:results-lesion}.
Activation patching shows that the heads of a lesion act together while the model reads the prompt (Tables~\ref{apptab:patching-sets} and~\ref{apptab:patching-sets-2}), and the same patching on the TAB shows that the damage beyond the narrative arises in the same way.
Comparisons between the lesions of one model show that the lesions for different targets share many heads (Table~\ref{apptab:subtype-overlap}), and that the lesions found on the Cinderella and the sandwich tasks share heads but rarely carry their target to the other task (Table~\ref{apptab:patching-overlap}).
A last analysis asks whether the Qwen3.5 lesions favor one of the two kinds of attention in these models.
We name a lesion by its model, the task that it was found on, and its target, for example the \gptxl sandwich aphasia lesion.

\begin{table*}[tp]
\centering
\footnotesize
\setlength{\tabcolsep}{3.5pt}
\begin{tabular}{@{}lllrrrrrrrrr@{}}
\toprule
 & & & & \multicolumn{2}{c}{Probability} & \multicolumn{3}{c}{Share of heads} & \multicolumn{3}{c}{TAB short answer} \\
\cmidrule(lr){5-6} \cmidrule(lr){7-9} \cmidrule(l){10-12}
 & & & & & Restore & & & & & Restore & Best \\
Model & Task & Target & Heads & Lesion & all & Restored & Left out & Alone & Lesion & all & restore \\
\midrule
\gptxl & Cinderella & Aphasia & $9$ & $0.891$ & $0.001$ & $0.67$ & $1.00$ & $0.00$ & $10/15$ & $9/15$ & $10/15$ \\
 &  & Wernicke & $244$ & $0.647$ & $0.001$ & $0.75$ & $0.95$ & $0.00$ & $4/15$ & $7/15$ & $5/15$ \\
 & Sandwich & Aphasia & $31$ & $0.862$ & $0.005$ & $0.81$ & $1.00$ & $0.00$ & $9/15$ & $7/15$ & $9/15$ \\
 &  & Broca & $146$ & $0.884$ & $0.000$ & $0.91$ & $1.00$ & $0.00$ & $4/15$ & $9/15$ & $5/15$ \\
 &  & Wernicke & $188$ & $0.717$ & $0.000$ & $0.61$ & $0.95$ & $0.00$ & $8/15$ & $9/15$ & $8/15$ \\
\midrule
\pythia{1B} & Cinderella & Aphasia & $8$ & $0.568$ & $0.048$ & $0.50$ & $1.00$ & $0.00$ & $7/15$ & $8/15$ & $9/15$ \\
 &  & Broca & $27$ & $0.568$ & $0.001$ & $0.74$ & $0.96$ & $0.00$ & $0/15$ & $4/15$ & $0/15$ \\
 &  & Wernicke & $13$ & $0.759$ & $0.244$ & $1.00$ & $1.00$ & $0.00$ & $3/15$ & $7/15$ & $4/15$ \\
 & Sandwich & Aphasia & $15$ & $0.691$ & $0.007$ & $0.87$ & $1.00$ & $0.00$ & $4/15$ & $10/15$ & $5/15$ \\
 &  & Broca & $35$ & $0.557$ & $0.002$ & $0.71$ & $0.94$ & $0.00$ & $0/15$ & $8/15$ & $0/15$ \\
 &  & Wernicke & $9$ & $0.703$ & $0.000$ & $1.00$ & $0.89$ & $0.00$ & $3/15$ & $11/15$ & $9/15$ \\
\midrule
\pythia{1.4B} & Cinderella & Aphasia & $19$ & $0.972$ & $0.022$ & $0.74$ & $1.00$ & $0.00$ & $0/15$ & $7/15$ & $3/15$ \\
 &  & Wernicke & $12$ & $0.751$ & $0.000$ & $0.83$ & $1.00$ & $0.00$ & $7/15$ & $9/15$ & $9/15$ \\
 & Sandwich & Aphasia & $13$ & $0.999$ & $0.037$ & $0.92$ & $1.00$ & $0.00$ & $8/15$ & $8/15$ & $8/15$ \\
 &  & Broca & $316$ & $0.488$ & $0.000$ & $0.54$ & $0.92$ & $0.00$ & $0/15$ & $5/15$ & $0/15$ \\
\midrule
\pythia{2.8B} & Cinderella & Aphasia & $1$ & $0.537$ & $0.001$ & $1.00$ & $1.00$ & $1.00$ & $7/15$ & $7/15$ & $7/15$ \\
 &  & Broca & $66$ & $0.801$ & $0.001$ & $0.91$ & $1.00$ & $0.00$ & $4/15$ & $7/15$ & $4/15$ \\
 &  & Wernicke & $63$ & $0.560$ & $0.051$ & $0.95$ & $1.00$ & $0.00$ & $2/15$ & $6/15$ & $5/15$ \\
 & Sandwich & Aphasia & $1$ & $0.706$ & $0.006$ & $1.00$ & $1.00$ & $1.00$ & $7/15$ & $7/15$ & $7/15$ \\
 &  & Broca & $128$ & $0.742$ & $0.001$ & $0.82$ & $0.99$ & $0.00$ & $3/15$ & $6/15$ & $4/15$ \\
 &  & Wernicke & $18$ & $0.504$ & $0.001$ & $0.83$ & $0.94$ & $0.00$ & $11/15$ & $6/15$ & $12/15$ \\
\midrule
\pythia{6.9B} & Cinderella & Aphasia & $1$ & $0.957$ & $\pmb{0.957}$ & $0.00$ & $1.00$ & $1.00$ & $7/15$ & $6/15$ & $6/15$ \\
 &  & Broca & $578$ & $0.993$ & $0.009$ & $0.74$ & $0.97$ & $0.00$ & $0/15$ & $3/15$ & $0/15$ \\
 &  & Wernicke & $1$ & $0.505$ & $0.016$ & $1.00$ & $1.00$ & $1.00$ & $8/15$ & $5/15$ & $5/15$ \\
 & Sandwich & Aphasia & $4$ & $0.723$ & $0.007$ & $0.50$ & $1.00$ & $0.00$ & $4/15$ & $5/15$ & $7/15$ \\
 &  & Broca & $13$ & $0.665$ & $0.000$ & $0.85$ & $1.00$ & $0.00$ & $7/15$ & $5/15$ & $7/15$ \\
 &  & Wernicke & $14$ & $0.720$ & $0.001$ & $0.57$ & $1.00$ & $0.00$ & $7/15$ & $6/15$ & $7/15$ \\
\bottomrule
\end{tabular}
\caption{Activation patching of the \ourmethod lesions of the base models \gptxl and Pythia that reach their targets.
\emph{Heads}: the number of heads in the set.
\emph{Probability}: the five-way classifier's probability of the target with the set knocked out, \emph{Lesion}, and with the clean outputs of all of its heads written back while the model reads the prompt, \emph{Restore all}, so that the set is knocked out only while the model writes its answer; an answer undoes the lesion when it no longer reaches the target, a probability of 0.5 or below with the target not the most probable class, and the one restore-all answer that still reaches it is in bold.
\emph{Share of heads}: the share of the set's heads whose clean output on the prompt undoes the lesion, \emph{Restored}, whose removal from the set undoes it, \emph{Left out}, and that reach the target when knocked out alone, \emph{Alone}.
\emph{TAB short answer}: short-answer items of the Text Aphasia Battery (TAB) for comprehension and repetition that the model answers correctly, of 15, with the set knocked out, with all heads restored, and with the best of at most eight single-head restores; the intact models answer 9 (\gptxl), 10 (\pythia{1B}), 8 (\pythia{1.4B}), 7 (\pythia{2.8B}), and 6 (\pythia{6.9B}) correctly.}
\label{apptab:patching-sets}
\end{table*}

\begin{table*}[tp]
\centering
\footnotesize
\setlength{\tabcolsep}{3.5pt}
\begin{tabular}{@{}lllrrrrrrrrr@{}}
\toprule
 & & & & \multicolumn{2}{c}{Probability} & \multicolumn{3}{c}{Share of heads} & \multicolumn{3}{c}{TAB short answer} \\
\cmidrule(lr){5-6} \cmidrule(lr){7-9} \cmidrule(l){10-12}
 & & & & & Restore & & & & & Restore & Best \\
Model & Task & Target & Heads & Lesion & all & Restored & Left out & Alone & Lesion & all & restore \\
\midrule
\llamaseven & Cinderella & Aphasia & $23$ & $0.606$ & $0.012$ & $0.61$ & $1.00$ & $0.00$ & $2/15$ & $7/15$ & $9/15$ \\
 &  & Broca & $42$ & $0.678$ & $0.001$ & $0.79$ & $1.00$ & $0.00$ & $7/15$ & $7/15$ & $7/15$ \\
 &  & Wernicke & $276$ & $0.682$ & $0.001$ & $0.88$ & $0.99$ & $0.00$ & $0/15$ & $5/15$ & $1/15$ \\
 & Sandwich & Aphasia & $60$ & $0.931$ & $0.007$ & $0.95$ & $1.00$ & $0.00$ & $3/15$ & $7/15$ & $3/15$ \\
 &  & Wernicke & $160$ & $0.736$ & $0.000$ & $0.89$ & $1.00$ & $0.00$ & $1/15$ & $8/15$ & $1/15$ \\
\midrule
\llamathirteen & Cinderella & Aphasia & $77$ & $0.679$ & $0.001$ & $0.73$ & $1.00$ & $0.00$ & $6/15$ & $7/15$ & $7/15$ \\
 &  & Broca & $101$ & $0.886$ & $0.001$ & $0.90$ & $1.00$ & $0.00$ & $2/15$ & $7/15$ & $3/15$ \\
 &  & Wernicke & $109$ & $0.786$ & $0.001$ & $0.63$ & $0.98$ & $0.00$ & $4/15$ & $8/15$ & $5/15$ \\
 & Sandwich & Aphasia & $90$ & $0.747$ & $0.005$ & $0.88$ & $1.00$ & $0.00$ & $5/15$ & $8/15$ & $5/15$ \\
 &  & Broca & $87$ & $0.636$ & $0.000$ & $0.90$ & $1.00$ & $0.00$ & $3/15$ & $7/15$ & $3/15$ \\
 &  & Wernicke & $67$ & $0.716$ & $0.000$ & $0.85$ & $1.00$ & $0.00$ & $4/15$ & $7/15$ & $6/15$ \\
\midrule
\qwen{0.8B} & Cinderella & Aphasia & $28$ & $0.524$ & $0.004$ & $0.93$ & $1.00$ & $0.00$ & $3/15$ & $12/15$ & $5/15$ \\
 &  & Wernicke & $28$ & $0.582$ & $0.000$ & $0.96$ & $0.96$ & $0.00$ & $3/15$ & $12/15$ & $5/15$ \\
 & Sandwich & Aphasia & $5$ & $0.948$ & $0.369$ & $0.40$ & $1.00$ & $0.00$ & $0/15$ & $12/15$ & $13/15$ \\
\midrule
\qwen{2B} & Cinderella & Aphasia & $43$ & $0.838$ & $0.001$ & $0.77$ & $1.00$ & $0.00$ & $4/15$ & $12/15$ & $5/15$ \\
 &  & Broca & $55$ & $0.851$ & $0.001$ & $0.96$ & $0.98$ & $0.00$ & $1/15$ & $12/15$ & $1/15$ \\
 & Sandwich & Aphasia & $72$ & $0.621$ & $0.008$ & $0.69$ & $1.00$ & $0.00$ & $5/15$ & $13/15$ & $8/15$ \\
 &  & Broca & $18$ & $0.738$ & $0.356$ & $0.89$ & $1.00$ & $0.00$ & $0/15$ & $8/15$ & $0/15$ \\
 &  & Wernicke & $63$ & $0.463$ & $0.000$ & $0.78$ & $0.98$ & $0.00$ & $0/15$ & $8/15$ & $1/15$ \\
\midrule
\qwen{4B} & Cinderella & Aphasia & $208$ & $0.508$ & $0.002$ & $0.48$ & $0.91$ & $0.00$ & $8/15$ & $13/15$ & $10/15$ \\
 &  & Broca & $81$ & $0.548$ & $0.001$ & $0.98$ & $1.00$ & $0.00$ & $6/15$ & $14/15$ & $8/15$ \\
 &  & Wernicke & $106$ & $0.572$ & $0.000$ & $0.90$ & $0.99$ & $0.00$ & $11/15$ & $14/15$ & $14/15$ \\
 & Sandwich & Aphasia & $88$ & $0.749$ & $0.005$ & $0.81$ & $1.00$ & $0.00$ & $9/15$ & $14/15$ & $11/15$ \\
 &  & Broca & $372$ & $0.959$ & $0.031$ & $0.95$ & $1.00$ & $0.00$ & $0/15$ & $8/15$ & $1/15$ \\
 &  & Wernicke & $228$ & $0.423$ & $0.006$ & $0.64$ & $0.93$ & $0.00$ & $6/15$ & $14/15$ & $6/15$ \\
\midrule
\qwen{9B} & Cinderella & Aphasia & $73$ & $0.587$ & $0.008$ & $0.75$ & $1.00$ & $0.00$ & $8/15$ & $14/15$ & $12/15$ \\
 & Sandwich & Aphasia & $67$ & $0.930$ & $0.006$ & $0.88$ & $1.00$ & $0.00$ & $11/15$ & $14/15$ & $12/15$ \\
\bottomrule
\end{tabular}
\caption{Activation patching of the \ourmethod lesions of the chat models Llama-2-chat and Qwen3.5 that reach their targets.
\emph{Heads}: the number of heads in the set.
\emph{Probability}: the five-way classifier's probability of the target with the set knocked out, \emph{Lesion}, and with the clean outputs of all of its heads written back while the model reads the prompt, \emph{Restore all}, so that the set is knocked out only while the model writes its answer; an answer undoes the lesion when it no longer reaches the target, a probability of 0.5 or below with the target not the most probable class.
\emph{Share of heads}: the share of the set's heads whose clean output on the prompt undoes the lesion, \emph{Restored}, whose removal from the set undoes it, \emph{Left out}, and that reach the target when knocked out alone, \emph{Alone}.
\emph{TAB short answer}: short-answer items of the Text Aphasia Battery (TAB) for comprehension and repetition that the model answers correctly, of 15, with the set knocked out, with all heads restored, and with the best of at most eight single-head restores; the intact models answer 7 (\llamaseven), 8 (\llamathirteen), 13 (\qwen{0.8B} and \qwen{2B}), and 14 (\qwen{4B} and \qwen{9B}) correctly.}
\label{apptab:patching-sets-2}
\end{table*}

\paragraph{Four patching conditions test where the heads act and which of them are needed.}
Activation patching copies activations from one run of a model into another run on the same input, to find out which components carry an effect~\citep{meng2022locating, zhang2024towards, heimersheim2024use}.
Here the clean run uses the intact model, and the lesioned run knocks out every head of a lesion at every position of the prompt and of the answer, as in \S\ref{sec:method-lesion}.
For each head of a lesion, we generate and score the model's answer under three conditions.
\emph{Restoring} the head writes its output from the clean run back at every position of the prompt, while the other heads of the lesion stay knocked out; the restored head itself stays knocked out at the positions of the answer, which have no counterpart in the clean run.
\emph{Leaving out} the head knocks out only the other heads of the lesion, so that the head runs normally at every position.
\emph{Knocking out the head alone} knocks out only this head, so that the other heads of the lesion run normally.
The first two conditions differ in what the head computes: a restored head's output on the prompt is copied from the clean run, whatever the other knockouts do to its input, whereas a head that is left out computes its output from a residual stream that the other knockouts have changed.
A fourth condition acts on the whole lesion: \emph{restoring all} writes back the clean outputs of every head of the lesion at every position of the prompt, so that the lesion is knocked out only while the model writes its answer.
In the terms of \citet{heimersheim2024use}, restoring is denoising: it writes clean activations into the lesioned run and asks whether they remove the effect of the lesion.
An answer reaches the target when it counts by the rules of Sections~\ref{sec:setup} and~\ref{sec:method-lesion} and the classifier either gives the target a probability above 0.5 or ranks the target as its most probable class, where any of the three aphasia classes counts for the aphasia target.
A condition \emph{undoes} the lesion when its answer no longer reaches the target.
The patching runs reproduce the stored intact and lesioned answers of all 54 lesions.

\paragraph{Restoring the heads' outputs on the prompt undoes 53 of the 54 lesions.}
Restoring all heads undoes 53 of the 54 lesions and leaves a target probability of at most 0.369 in them (Tables~\ref{apptab:patching-sets} and~\ref{apptab:patching-sets-2}).
These 53 lesions need their heads knocked out while the model reads the prompt, and knocking the heads out only while the model writes its answer does not reach the target.
The exception is the \pythia{6.9B} Cinderella aphasia lesion, which consists of a single head, L1H26, that is, head 26 of layer 1, both counted from 0.
With the clean output of this head written back on the prompt, the target probability stays at 0.957, so knocking the head out only while the model writes its answer already reaches the target.
Restoring a single head usually undoes a lesion as well.
The median share of a lesion's heads whose restore undoes the lesion is 0.83, both over the 54 lesions and over the 50 lesions of two or more heads, and at least one single restore undoes every lesion except the \pythia{6.9B} Cinderella aphasia lesion.
Pooled over the 54 lesions, restoring undoes the lesion for 3,579 of the 4,600 heads.
Two other classifier-guided searches of Appendix~\ref{appsec:lesion-methods}, whose sets above 0.5 we patched in the same way, give similar medians over all lesioned components of a set: 0.87 over the 20 sets of the layer search, which can add blocks of MLP neurons, and 0.96 over the 37 sets of the search that sweeps heads on top of grown MLP lesions.
The heads of a lesion thus act while the model reads the prompt, and most of them matter there: giving back the clean output of a single head usually undoes the lesion.

\paragraph{A single restore mostly breaks the answer instead of repairing it.}
Undoing a lesion means only that the answer no longer reaches the target, not that the model answers as it does when intact.
Of the 3,579 single restores that undo a lesion, 3,541 give an answer long enough to score.
The classifier labels 2,819 of these answers model scrambles, or 80\%, and only 450 control speech, or 13\%; the other 272 have an aphasia class as their most probable class.
The intact answer is not always control speech: the classifier labels it control speech for 41 of the 54 lesions and model scrambles for the other 13, all of them on the Cinderella task.
Over the 41 lesions whose intact answer is control speech, the result is the same: of the 2,767 scored answers of single restores that undo these lesions, the classifier labels 2,203 model scrambles, or 80\%, and 331 control speech, or 12\%.
Restoring all heads repairs the answer more often: it undoes 40 of these 41 lesions, and the classifier labels 22 of the 40 answers control speech and 13 model scrambles.
The aphasia-like answer thus needs the whole lesion, and giving one head back its clean output mostly breaks the answer in another way.

\paragraph{Almost every head is needed, largely by construction.}
Leaving out one head undoes the lesion for 4,484 of the 4,600 heads, and in 37 of the 54 lesions, leaving out any one head undoes the lesion.
No head of the 50 lesions of two or more heads reaches the target when knocked out alone.
Both results follow mostly from how \ourmethod builds the lesions, so they check the lesions rather than reveal how the heads work.
Growth ranks the heads by the target probability that each gives when knocked out alone and stops at the first prefix whose probability exceeds 0.5, so a lesion of two or more heads arises only when no head exceeds 0.5 alone.
Pruning removes heads for as long as the target probability stays above 0.5, so a lesion whose pruning finished sits at the edge where removing any one more head brings the probability to 0.5 or below, and leaving out a head repeats this last test of pruning.
By the search's own rule, a probability above 0.5, every head is therefore needed in all 47 lesions whose pruning finished, and in 48 of the 51 lesions that exceed 0.5.
Reaching the target, as defined above, also includes an answer at 0.5 or below whose most probable class is still the target, so by this definition every head is needed in only 37 of the 47 pruned lesions.
In the other 10, leaving out a head lowers the target probability to 0.5 or below while the target stays the most probable class, for 1 to 9 heads per lesion and 20 heads in all.
Beyond this difference between the two rules, leaving out heads tests more than pruning did only in the seven lesions whose pruning did not finish: the four that stopped at the budget of 15,000 answers and the three that reach their target only as the most probable class, which the search never pruned.
In these seven lesions, 91 to 99\% of the heads are needed: 233 of 244 in the \gptxl Cinderella Wernicke lesion, 272 of 276 in the \llamaseven Cinderella Wernicke lesion, 558 of 578 in the \pythia{6.9B} Cinderella Broca lesion, 190 of 208 in the \qwen{4B} Cinderella aphasia lesion, 291 of 316 in the \pythia{1.4B} sandwich Broca lesion, 62 of 63 in the \qwen{2B} sandwich Wernicke lesion, and 211 of 228 in the \qwen{4B} sandwich Wernicke lesion.
By the search's rule, all 276 heads of the \llamaseven lesion are needed, so a further round of pruning would have removed none of them.
The restore conditions are not settled by construction, because a restored head differs from a head that is left out: it stays knocked out while the model writes its answer, and its output on the prompt is the clean one.
Because pruning leaves most lesions at the edge of their target, these yes-or-no outcomes show where the heads act, but not how much each head contributes, and a change that breaks the answer counts as undoing the lesion as much as one that repairs it.
Per-head patching also cannot show which head feeds which; path patching, which restores only the connection from one component to another~\citep{wang2023interpretability}, would trace these connections.

\paragraph{Restoring all heads also repairs most of the damage on the TAB.}
We repeat the lesion, restoring all, and single restores on the 20 items of the TAB (Appendix~\ref{appsec:beyond}): 15 short-answer items for word comprehension, sentence comprehension, and repetition, which a rule scores, and five connected-text prompts.
Single restores cover every head of the 7 lesions of at most eight heads and, in the other 47 lesions, the eight heads whose removal from the lesion lowers the target probability on the narrative the most.
The intact and lesioned TAB answers of all 54 lesions match those of Appendix~\ref{appsec:beyond} word for word.
Tables~\ref{apptab:patching-sets} and~\ref{apptab:patching-sets-2} give the number of the 15 short-answer items that the model answers correctly in each condition.
With the lesion in place, 41 of the 54 lesions lose two or more of these items, counted from the intact model's number of correct items.
Restoring all heads gives back at least half of the lost items in 39 of these 41 lesions and all of them in 16, whereas the best single restore gives back at least half in only 10 and all of them in 4.
Over all 54 lesions, the model answers a median of 4 of the 15 items correctly with the lesion in place, 8 with all heads restored, and 5.5 with the best single restore, against 8.5 for the intact models.
The median number of items lost is 5, 1, and 3 in these three conditions.
On the connected-text prompts, the lesions rarely reach their target, so a restore has little to undo there: the classifier finds the target in 12 of the 259 answers that it scores with the lesion in place, in 8 of 269 with all heads restored, and in 2 of the intact model's 270.
We did not measure MMLU under patching.
Unlike the target probability, the number of correct items shows whether the answers recover, and by this measure restoring all heads repairs most of the damage, whereas restoring one of the tested heads rarely does.
The damage beyond the narrative thus has the same source as the narrative's target: the heads acting while the model reads the prompt.

\begin{table*}[tp]
\centering
\footnotesize
\setlength{\tabcolsep}{3.5pt}
\begin{tabular}{@{}lrrrrrrrrrrrr@{}}
\toprule
 & \multicolumn{3}{c}{Heads} & \multicolumn{3}{c}{Aphasia and Broca} & \multicolumn{3}{c}{Aphasia and Wernicke} & \multicolumn{3}{c}{Broca and Wernicke} \\
\cmidrule(lr){2-4} \cmidrule(lr){5-7} \cmidrule(lr){8-10} \cmidrule(l){11-13}
Task & Aph. & Broca & Wern. & Shared & Chance & $p$ & Shared & Chance & $p$ & Shared & Chance & $p$ \\
\midrule
\multicolumn{13}{@{}l}{\gptxl} \\
Cinderella & $9$ & -- & $244$ & -- & -- & -- & $0$ & $1.83$ & $1$ & -- & -- & -- \\
Sandwich & $31$ & $146$ & $188$ & $18$ & $3.77$ & $<\!0.001$ & $12$ & $4.86$ & $0.0014$ & $70$ & $22.87$ & $<\!0.001$ \\
\midrule
\multicolumn{13}{@{}l}{\pythia{1B}} \\
Cinderella & $8$ & $27$ & $13$ & $4$ & $1.69$ & $0.061$ & $2$ & $0.81$ & $0.19$ & $7$ & $2.74$ & $0.0062$ \\
Sandwich & $15$ & $35$ & $9$ & $7$ & $4.10$ & $0.074$ & $4$ & $1.05$ & $0.011$ & $3$ & $2.46$ & $0.47$ \\
\midrule
\multicolumn{13}{@{}l}{\pythia{1.4B}} \\
Cinderella & $19$ & -- & $12$ & -- & -- & -- & $0$ & $0.59$ & $1$ & -- & -- & -- \\
Sandwich & $13$ & $316$ & -- & $13$ & $10.70$ & $0.076$ & -- & -- & -- & -- & -- & -- \\
\midrule
\multicolumn{13}{@{}l}{\pythia{2.8B}} \\
Cinderella & $1$ & $66$ & $63$ & $1$ & $0.06$ & $0.064$ & $0$ & $0.06$ & $1$ & $36$ & $4.06$ & $<\!0.001$ \\
Sandwich & $1$ & $128$ & $18$ & $1$ & $0.12$ & $0.12$ & $1$ & $0.02$ & $0.018$ & $14$ & $2.25$ & $<\!0.001$ \\
\midrule
\multicolumn{13}{@{}l}{\pythia{6.9B}} \\
Cinderella & $1$ & $578$ & $1$ & $1$ & $0.56$ & $0.56$ & $0$ & $0.00$ & $1$ & $1$ & $0.56$ & $0.56$ \\
Sandwich & $4$ & $13$ & $14$ & $2$ & $0.05$ & $<\!0.001$ & $2$ & $0.05$ & $0.001$ & $7$ & $0.18$ & $<\!0.001$ \\
\midrule
\multicolumn{13}{@{}l}{\llamaseven} \\
Cinderella & $23$ & $42$ & $276$ & $8$ & $0.94$ & $<\!0.001$ & $14$ & $6.20$ & $<\!0.001$ & $36$ & $11.32$ & $<\!0.001$ \\
Sandwich & $60$ & -- & $160$ & -- & -- & -- & $37$ & $9.38$ & $<\!0.001$ & -- & -- & -- \\
\midrule
\multicolumn{13}{@{}l}{\llamathirteen} \\
Cinderella & $77$ & $101$ & $109$ & $45$ & $4.86$ & $<\!0.001$ & $30$ & $5.25$ & $<\!0.001$ & $37$ & $6.88$ & $<\!0.001$ \\
Sandwich & $90$ & $87$ & $67$ & $17$ & $4.89$ & $<\!0.001$ & $22$ & $3.77$ & $<\!0.001$ & $30$ & $3.64$ & $<\!0.001$ \\
\midrule
\multicolumn{13}{@{}l}{\qwen{0.8B}} \\
Cinderella & $28$ & -- & $28$ & -- & -- & -- & $12$ & $2.33$ & $<\!0.001$ & -- & -- & -- \\
\midrule
\multicolumn{13}{@{}l}{\qwen{2B}} \\
Cinderella & $43$ & $55$ & -- & $4$ & $7.04$ & $0.95$ & -- & -- & -- & -- & -- & -- \\
Sandwich & $72$ & $18$ & $63$ & $9$ & $3.86$ & $0.0055$ & $28$ & $13.50$ & $<\!0.001$ & $18$ & $3.38$ & $<\!0.001$ \\
\midrule
\multicolumn{13}{@{}l}{\qwen{4B}} \\
Cinderella & $208$ & $81$ & $106$ & $23$ & $18.80$ & $0.15$ & $23$ & $24.61$ & $0.69$ & $38$ & $9.58$ & $<\!0.001$ \\
Sandwich & $88$ & $372$ & $228$ & $83$ & $36.54$ & $<\!0.001$ & $62$ & $22.39$ & $<\!0.001$ & $220$ & $94.66$ & $<\!0.001$ \\
\midrule
\multicolumn{4}{@{}l}{All pairs} & $236$ & $98.00$ & $7$ of $15$ & $249$ & $96.70$ & $9$ of $17$ & $517$ & $164.59$ & $11$ of $13$ \\
\multicolumn{4}{@{}l}{Median share} & \multicolumn{3}{c}{$50\%$} & \multicolumn{3}{c}{$39\%$} & \multicolumn{3}{c}{$54\%$} \\
\bottomrule
\end{tabular}
\caption{Heads shared by the \ourmethod lesions that a model has for two targets on the same task, for every pair of targets whose two lesions reach their target.
\emph{Heads}: the number of heads in the lesion for the aphasia (Aph.), Broca, and Wernicke (Wern.) target, where a dash marks a target without a lesion.
\emph{Shared}: the heads that the two lesions of a pair share.
\emph{Chance}: the number of shared heads expected if both lesions were drawn at random from the model's $N$ heads, ${|L_1|\,|L_2|/N}$ for lesions $L_1$ and $L_2$.
$p$: the one-sided hypergeometric probability of sharing at least as many heads by chance.
\emph{All pairs}: the total of shared heads, the total expected by chance, and the number of pairs with $p < 0.01$.
\emph{Median share}: the median over pairs of the shared heads as a share of the smaller lesion.
Three combinations of model and task with fewer than two lesions, \qwen{0.8B} on sandwich and \qwen{9B} on both tasks, are left out.}
\label{apptab:subtype-overlap}
\end{table*}

\begin{table*}[t]
\centering
\footnotesize
\setlength{\tabcolsep}{4pt}
\begin{tabular}{@{}lrrrrrrrrrr@{}}
\toprule
 & \multicolumn{3}{c}{Heads} & & & & \multicolumn{2}{c}{Union} & \multicolumn{2}{c}{Intersection} \\
\cmidrule(lr){2-4} \cmidrule(lr){8-9} \cmidrule(l){10-11}
Target & Cinderella & Sandwich & Shared & Chance & Ratio & $p$ & Cinderella & Sandwich & Cinderella & Sandwich \\
\midrule
\multicolumn{11}{@{}l}{\gptxl} \\
Aphasia & $9$ & $31$ & $1$ & $0.23$ & $4.30$ & $0.21$ & $0.000$ & $0.001$ & $0.015$ & $0.007$ \\
Wernicke & $244$ & $188$ & $92$ & $38.23$ & $2.41$ & $1.9 \times 10^{-22}$ & $0.000$ & $0.000$ & $0.000$ & $0.000$ \\
\midrule
\multicolumn{11}{@{}l}{\llamathirteen} \\
Aphasia & $77$ & $90$ & $24$ & $4.33$ & $5.54$ & $2.5 \times 10^{-13}$ & $0.001$ & $0.003$ & $0.018$ & $0.001$ \\
Broca & $101$ & $87$ & $28$ & $5.49$ & $5.10$ & $1.6 \times 10^{-14}$ & $0.001$ & $0.001$ & $0.001$ & $0.088$ \\
Wernicke & $109$ & $67$ & $24$ & $4.56$ & $5.26$ & $7.4 \times 10^{-13}$ & $0.004$ & $0.000$ & $0.000$ & $0.000$ \\
\midrule
\multicolumn{11}{@{}l}{\pythia{1B}} \\
Aphasia & $8$ & $15$ & $0$ & $0.94$ & $0.00$ & $1$ & $\pmb{0.965}$ & $\pmb{0.953}$ & -- & -- \\
Broca & $27$ & $35$ & $11$ & $7.38$ & $1.49$ & $0.068$ & $0.001$ & $0.000$ & $0.001$ & $0.000$ \\
Wernicke & $13$ & $9$ & $0$ & $0.91$ & $0.00$ & $1$ & $0.008$ & $0.000$ & -- & -- \\
\midrule
\multicolumn{11}{@{}l}{\qwen{9B}} \\
Aphasia & $73$ & $67$ & $14$ & $5.46$ & $2.56$ & $0.0005$ & $0.267$ & $0.002$ & $0.002$ & $0.005$ \\
\bottomrule
\end{tabular}
\caption{Heads shared by the \ourmethod lesions that a model has for the same target on the Cinderella and on the sandwich task, in one model per family.
\emph{Heads}: the number of heads in the Cinderella set $C$, in the sandwich set $S$, and in both.
\emph{Chance}: the number of shared heads expected if both sets were drawn at random from the model's $N$ heads, ${|C|\,|S|/N}$.
\emph{Ratio}: shared heads divided by chance.
$p$: the one-sided hypergeometric probability of sharing at least as many heads by chance.
\emph{Union} and \emph{Intersection}: the classifier's probability of the target for the answer to each task with the heads of either set, or only the shared heads, knocked out; values above 0.5, in bold, reach the target, and a dash marks pairs without shared heads.
Three pairs are left out because a set does not reach its target: Broca's aphasia in \gptxl, whose Cinderella set does not reach it, and Broca's and Wernicke's aphasia in \qwen{9B}, whose sets reach it on neither task.}
\label{apptab:patching-overlap}
\end{table*}

\paragraph{The lesions for different targets share many heads.}
For each model and task, \ourmethod runs one search per target, which ranks the heads by that target's probability, grows the lesion in this order, and prunes it against the same target.
If two lesions $L_1$ and $L_2$ were drawn at random from a model's $N$ heads, they would share ${|L_1|\,|L_2|/N}$ heads on average, and a one-sided hypergeometric test gives the probability of sharing at least the observed number of heads by chance (Table~\ref{apptab:subtype-overlap}).
The Broca and Wernicke lesions give the cleanest test, because the two targets are separate classes of the classifier.
Over the 13 pairs of Broca and Wernicke lesions of one model and task that both reach their target, the two lesions share a median of 54\% of the heads of the smaller lesion and 517 heads in all, three times the 165 expected by chance, and the overlap exceeds chance at $p < 0.01$ in 11 of the 13 pairs.
The two exceptions are the \pythia{1B} sandwich pair, whose lesions share 3 heads against 2.5 by chance, and the \pythia{6.9B} Cinderella pair, whose one-head Wernicke lesion lies inside the 578-head Broca lesion.
The lesions for the aphasia target, whose probability includes those of the Broca and Wernicke classes, overlap with both: they share a median of 50\% of the smaller lesion with the Broca lesion, 236 heads against 98 by chance over 15 pairs, and 39\% with the Wernicke lesion, 249 heads against 97 over 17 pairs.
Nothing in the design makes the searches for different targets share heads, but they rank the same single-head answers: a head whose knockout makes the answer more Broca-like usually makes it more Wernicke-like too, at a median Spearman correlation across heads of 0.93 over the 13 pairs, from 0.49 to 0.99, so the two growths follow similar rankings.
Ties add to the overlap: heads whose knockout alone leaves the target probability exactly at the intact value tie with one another in that target's ranking and keep their layer order (Appendix~\ref{appsec:lesion-methods}), so when the growths for two targets reach such heads, they add them in the same order.
Heads tied in both rankings make up 134 of the 517 heads that the Broca and Wernicke lesions share, 80 of them in the Llama-2-chat models, and among the other heads of the models, the two lesions still share 383 heads against 156 expected by chance, with $p < 0.01$ in 9 of the 13 pairs.
Broca-like and Wernicke-like answers thus come from lesions that share many heads, so the circuits for different types of aphasia are entangled rather than separate.

\paragraph{The lesions found on the two tasks share heads, but their targets rarely carry over.}
For one model of each family, \gptxl, \llamathirteen, \pythia{1B}, and \qwen{9B}, we compare the lesion found on the Cinderella task with the lesion found on the sandwich task for the same target (Table~\ref{apptab:patching-overlap}).
Of the 12 pairs, 9 can be compared; in the other three, the lesion of at least one task does not reach its target.
With the chance level and the test of the previous paragraph, the lesions share heads in 7 of the 9 pairs, and far more than chance in five, all at $p < 0.01$: 5.1 to 5.5 times chance in the three pairs of \llamathirteen, 2.4 times in the \gptxl Wernicke pair, and 2.6 times in the \qwen{9B} aphasia pair.
In \llamathirteen, however, 55 of the 76 shared heads are tied in the rankings of both tasks, and without the heads tied on both tasks, the overlap stays above chance at $p < 0.01$ in four of the five pairs, all but the \llamathirteen aphasia pair.
Yet the shared heads do not carry the lesions.
Knocked out alone, they reach the target on neither task in any of the 7 pairs with shared heads, and in these pairs neither do the heads that belong to only one of the two lesions.
Knocking out the union of the two lesions reaches the target on both tasks in only one pair, the \pythia{1B} aphasia pair, whose union of 23 heads gives 0.965 on Cinderella and 0.953 on sandwich.
Pruning this union for as long as both answers stay above 0.5 leaves 16 heads, 6 of the Cinderella lesion's 8 and 10 of the sandwich lesion's 15.
For 6 of the other 8 unions, the classifier labels both answers model scrambles; for the union of the \llamathirteen Broca pair, it labels the Cinderella answer model scrambles and the sandwich answer control speech, and for the union of the \pythia{1B} Wernicke pair, it labels the Cinderella answer Broca's aphasia and the sandwich answer model scrambles.
Of the 18 lesions in the 9 pairs, only one reaches its target on both tasks by itself, the \gptxl sandwich aphasia lesion, at 0.859 on Cinderella and 0.862 on sandwich.
The lesions found on the two tasks thus share heads, but the shared heads never produce the target, and combining the two lesions mostly breaks the answers into model scrambles instead of making both of them aphasic.

\paragraph{The Qwen3.5 lesions lean toward softmax heads.}
Every fourth layer of a Qwen3.5 model uses full attention, computed with a softmax, and the other layers use Gated DeltaNet linear attention~\citep{yang2025gated}; we call the heads of the full-attention layers softmax heads.
We count each value head of a Gated DeltaNet layer, which keeps its own recurrent state, as one head, so softmax heads make up 14.3\% of the heads of every Qwen3.5 model.
Pooled over the 16 Qwen3.5 lesions, they make up 294 of 1,535 heads, or 19.2\%, against 219.3 expected at 14.3\%; this pooled share only describes the lesions, because the lesions of one model are not independent draws.
Softmax heads make up more than 14.3\% of the lesion in 15 of the 16 lesions; the exception, the \qwen{0.8B} sandwich aphasia lesion, has 5 heads and no softmax head.
A one-sided hypergeometric test against random sets of the same size, without correction for the 16 tests, finds softmax heads over-represented at $p < 0.05$ in five lesions: the \qwen{9B} Cinderella aphasia lesion, with 25 softmax heads of 73 at $p = 5.3 \times 10^{-6}$; the \qwen{2B} sandwich Wernicke lesion, with 18 of 63 at $p = 7.2 \times 10^{-4}$; and three lesions of \qwen{4B}, the Cinderella Broca lesion with 20 of 81 at $p = 0.0063$, the sandwich Wernicke lesion with 42 of 228 at $p = 0.027$, and the sandwich aphasia lesion with 19 of 88 at $p = 0.033$.
With a Bonferroni correction for the 16 tests, the first two stay below 0.05.
The lesions thus lean toward the full-attention layers, although Gated DeltaNet heads make up most of every lesion.

\section{Steering Details}
\label{appsec:steering}

We next describe how we compute the steering vectors of \S\ref{sec:method-steer}, what linear probes find in the residual stream of the human transcripts, how the sweep chooses a layer and a strength, which settings the human evaluation uses, three further steering experiments, and how alike the Broca and Wernicke vectors are.

\paragraph{Steering vectors.}
For each model and task, the model reads every human transcript of the task, appended to the task prompt as its own answer: in the plain question-and-answer format for the base models, and in the chat template for the chat models, with thinking turned off for Qwen3.5.
We average the residual stream after every block over the tokens of the transcript, leaving out those of the prompt, which gives the states $h_\ell(x)$ of Equation~\eqref{eq:steering-vector}.
We count layers from 0, so layer 0 is the output of the first block.
We cut the prompt and the transcript together at 2,048 tokens, or at 1,024 for \gptxl, which shortens 58 of the 1,139 Cinderella transcripts in \gptxl, 5 in the Llama-2 models, 3 in the Pythia and Qwen3.5 models, and none of the sandwich transcripts.
Each text counts once: when AphasiaBank lists the same text more than once, we keep only its first copy, and we drop every copy of a text that it lists under two groups.
\gptxl and the Pythia models read their residual stream through LayerNorm, which subtracts the mean over the model dimension, so for these models we subtract that mean from every averaged state before any further step.
The Llama-2 and Qwen3.5 models use RMSNorm, which does not subtract the mean, so we keep their raw residual stream.
The aphasia vector is the mean of $h_\ell(x)$ over the transcripts of the four aphasic groups minus the mean over the control transcripts, where the four groups are anomic, conduction, Broca's, and Wernicke's aphasia.
We pool the aphasic transcripts rather than average the four group means, so a larger group weighs more.
The Broca and Wernicke vectors subtract the control mean from the mean over the Broca or the Wernicke transcripts.

\paragraph{Linear probes.}
At every layer, we fit a logistic regression with an L2 penalty, at an inverse regularization strength of 0.01, to the averaged states, which we standardize with the mean and standard deviation of the training folds.
We evaluate each probe by five-fold cross-validation, with folds grouped by speaker, so that all transcripts of a speaker fall into one fold, and stratified by label, and we report the mean AUC over the five held-out folds.
The aphasia probe separates the 744 Cinderella and 408 sandwich transcripts of the four aphasic groups from the 395 and 472 control transcripts.
Unlike the Broca and Wernicke vectors, which compare a type with control speech, the Broca and Wernicke probes separate the transcripts of one type from all other transcripts, including those of the other types of aphasia, so their AUCs measure how well a type stands out from all other speech.
A spot check suggests that the other types make these probes harder: at layer 8 of \pythia{1B} on the Cinderella task, probes against the control transcripts alone separate Broca's and Wernicke's aphasia at AUCs of 0.9999 and 0.9993, against 0.945 and 0.822 for the probes against all other transcripts.
To test whether the two tasks share the aphasia direction, we fit the aphasia probe on all transcripts of one task and test it on all transcripts of the other.
Because 545 speakers occur in both tasks, a high transfer AUC shows that the tasks share a direction, not that the probe generalizes to new speakers.
For severity, we project each aphasic transcript on the unit aphasia vector, $\vvec_\ell / \|\vvec_\ell\|$, and compute the Spearman correlation $\rho$ between the projection and the speaker's aphasia quotient from the Western Aphasia Battery~\citep{kertesz2007western}, over the 720 of 744 aphasic Cinderella transcripts and the 382 of 408 aphasic sandwich transcripts that come with a quotient.
Lower quotients mean more severe aphasia, so a negative $\rho$ means that the projection grows with severity.
Last, we compute at every layer the first principal component of the averaged states of all transcripts, centered over transcripts, and its absolute cosine with the aphasia vector.

\begin{table*}[t]
\centering
\footnotesize
\setlength{\tabcolsep}{4.5pt}
\begin{tabular}{@{}lrrrrrrrr@{}}
\toprule
 & & \multicolumn{4}{c}{AUC, lowest to highest over layers} & \multicolumn{2}{c}{Severity} & \\
\cmidrule(lr){3-6}\cmidrule(lr){7-8}
Model & Layers & Aphasia & Broca & Wernicke & Transfer & $\rho$ & Layer & PC1 \\
\midrule
\multicolumn{9}{@{}l}{\textit{Cinderella task}} \\
\gptxl & $48$ & $0.981$--$0.993$ & $0.944$--$0.951$ & $0.773$--$0.851$ & $0.871$--$0.958$ & $-0.759$ & $44$ & $0.955$ \\
\addlinespace[2pt]
\llamaseven & $32$ & $0.980$--$0.994$ & $0.937$--$0.947$ & $0.764$--$0.850$ & $0.884$--$0.971$ & $-0.797$ & $11$ & $0.948$ \\
\llamathirteen & $40$ & $0.979$--$0.995$ & $0.935$--$0.947$ & $0.765$--$0.857$ & $0.889$--$0.978$ & $-0.806$ & $17$ & $0.943$ \\
\addlinespace[2pt]
\pythia{1B} & $16$ & $0.982$--$0.992$ & $0.939$--$0.949$ & $0.772$--$0.834$ & $0.889$--$0.970$ & $-0.788$ & $12$ & $0.942$ \\
\pythia{1.4B} & $24$ & $0.983$--$0.992$ & $0.940$--$0.950$ & $0.783$--$0.856$ & $0.867$--$0.961$ & $-0.770$ & $23$ & $0.941$ \\
\pythia{2.8B} & $32$ & $0.982$--$0.994$ & $0.936$--$0.951$ & $0.777$--$0.857$ & $0.866$--$0.977$ & $-0.773$ & $12$ & $0.946$ \\
\pythia{6.9B} & $32$ & $0.982$--$0.993$ & $0.933$--$0.949$ & $0.762$--$0.850$ & $0.909$--$0.970$ & $-0.782$ & $31$ & $0.951$ \\
\addlinespace[2pt]
\qwen{0.8B} & $24$ & $0.982$--$0.992$ & $0.938$--$0.956$ & $0.802$--$0.847$ & $0.894$--$0.964$ & $-0.777$ & $20$ & $0.933$ \\
\qwen{2B} & $24$ & $0.982$--$0.993$ & $0.936$--$0.956$ & $0.801$--$0.872$ & $0.904$--$0.966$ & $-0.808$ & $20$ & $0.933$ \\
\qwen{4B} & $32$ & $0.984$--$0.995$ & $0.937$--$0.952$ & $0.809$--$0.857$ & $0.896$--$0.969$ & $-0.807$ & $27$ & $0.938$ \\
\qwen{9B} & $32$ & $0.983$--$0.995$ & $0.938$--$0.950$ & $0.811$--$0.867$ & $0.877$--$0.978$ & $-0.809$ & $27$ & $0.941$ \\
\midrule
\multicolumn{9}{@{}l}{\textit{Sandwich task}} \\
\gptxl & $48$ & $0.964$--$0.984$ & $0.950$--$0.958$ & $0.771$--$0.855$ & $0.939$--$0.986$ & $-0.635$ & $38$ & $0.927$ \\
\addlinespace[2pt]
\llamaseven & $32$ & $0.966$--$0.989$ & $0.947$--$0.960$ & $0.792$--$0.865$ & $0.963$--$0.990$ & $-0.701$ & $13$ & $0.938$ \\
\llamathirteen & $40$ & $0.965$--$0.990$ & $0.940$--$0.956$ & $0.794$--$0.866$ & $0.958$--$0.992$ & $-0.698$ & $15$ & $0.918$ \\
\addlinespace[2pt]
\pythia{1B} & $16$ & $0.965$--$0.984$ & $0.952$--$0.957$ & $0.768$--$0.859$ & $0.966$--$0.986$ & $-0.638$ & $9$ & $0.890$ \\
\pythia{1.4B} & $24$ & $0.965$--$0.984$ & $0.950$--$0.960$ & $0.760$--$0.866$ & $0.969$--$0.985$ & $-0.640$ & $14$ & $0.887$ \\
\pythia{2.8B} & $32$ & $0.964$--$0.985$ & $0.951$--$0.961$ & $0.758$--$0.870$ & $0.965$--$0.989$ & $-0.645$ & $17$ & $0.920$ \\
\pythia{6.9B} & $32$ & $0.960$--$0.986$ & $0.947$--$0.959$ & $0.775$--$0.878$ & $0.957$--$0.989$ & $-0.654$ & $12$ & $0.875$ \\
\addlinespace[2pt]
\qwen{0.8B} & $24$ & $0.971$--$0.982$ & $0.953$--$0.959$ & $0.734$--$0.851$ & $0.974$--$0.987$ & $-0.668$ & $12$ & $0.931$ \\
\qwen{2B} & $24$ & $0.970$--$0.985$ & $0.940$--$0.960$ & $0.738$--$0.853$ & $0.972$--$0.987$ & $-0.651$ & $11$ & $0.952$ \\
\qwen{4B} & $32$ & $0.972$--$0.988$ & $0.950$--$0.963$ & $0.757$--$0.869$ & $0.971$--$0.991$ & $-0.688$ & $14$ & $0.939$ \\
\qwen{9B} & $32$ & $0.972$--$0.988$ & $0.946$--$0.958$ & $0.764$--$0.847$ & $0.971$--$0.991$ & $-0.690$ & $17$ & $0.956$ \\
\bottomrule
\end{tabular}
\caption{Linear probes and directions in the residual stream that each model computes while it reads the human transcripts of a task as its own answer.
Each AUC column gives the lowest and the highest area under the ROC curve over all layers.
\emph{Aphasia}: transcripts of the four aphasic groups against control transcripts, on speakers held out by five-fold cross-validation.
\emph{Broca} and \emph{Wernicke}: transcripts of that type against all other transcripts, including those of the other types of aphasia, evaluated in the same way.
\emph{Transfer}: the aphasia probe fitted on all transcripts of the row's task and tested on all transcripts of the other task.
\emph{Severity}: the most negative Spearman correlation $\rho$, over layers, between the projection of an aphasic transcript on the aphasia vector and the speaker's aphasia quotient, where lower quotients mean more severe aphasia, and the layer at which it occurs, counted from 0.
\emph{PC1}: the absolute cosine between the aphasia vector and the first principal component of all transcripts' token-averaged residual streams, averaged over layers.}
\label{apptab:steering-repr}
\end{table*}

\paragraph{Probe results.}
Table~\ref{apptab:steering-repr} gives these statistics for every model and task.
Over all 11 models and all layers, the aphasia probe reaches an AUC of 0.960 to 0.995, the Broca probe 0.933 to 0.963, the Wernicke probe 0.734 to 0.878, and the transfer 0.866 to 0.992.
The most negative $\rho$ over layers is weaker on the sandwich task, from $-0.635$ to $-0.701$ across models, than on the Cinderella task, from $-0.759$ to $-0.809$.
The first principal component lies close to the aphasia vector: their absolute cosine averages 0.875 to 0.956 over the layers of a model and task, and it reaches 0.955 to 0.988 at the best layer.
It falls below 0.8 only in the first three layers of some models on the sandwich task and in the last layer of \llamathirteen on the same task, down to 0.086 at layer 1 of \pythia{6.9B}.
The direction along which the transcripts vary most is thus, at almost every layer, close to the direction that separates aphasic from control speech.

\paragraph{The sweep.}
For each model, task, and vector, the sweep adds the vector of the same task at one of 8 layers, $\mathrm{round}(i L / 8)$ for $i = 0, \dotsc, 7$, where $L$ is the model's number of layers, for example layers 0, 4, \dots, 28 of a model with 32 layers.
It uses eight strengths toward the target, $r \in \{0.1, 0.25, 0.5, 1, 1.5, 2, 3, 4\}$, and three toward control speech, $r \in \{-0.5, -1, -2\}$.
The strength $r$ compares the norm of the added vector with $\bar h_\ell$, the mean norm of the averaged states $h_\ell(x)$ over the task's transcripts, so it measures the vector against the human transcripts' averaged states, not against the per-token states of the model's own answer.
Adding the aphasia vector once, with $\alpha = 1$, corresponds to $r$ between 0.035 and 0.508, depending on the model, task, and layer, and at the smallest strength that reaches the target, the sweep adds a median of 5.77 copies of the aphasia vector.
We add the vector at every position of the prompt and of the answer, and on the Cinderella task also while the model answers \say{Do you know the story of Cinderella?}, which it answers separately from the story request.
Each setting, a layer and a strength, yields one greedy answer, with a repetition penalty of 1.25 and at most 180 new tokens, so the sweep has no sampling variance.
Of the 5,808 steered answers, 683 are shorter than 20 characters, and such answers never count as reaching a target.
A vector works at a layer when some positive strength makes the classifier's probability of the vector's target exceed 0.5, and for each model, task, and vector we report how many of the 8 layers work and the median, over the working layers, of the smallest strength that works.
As a control, each model and task gets three random directions, drawn from a standard Gaussian at every layer, centered over the model dimension for \gptxl and Pythia, and scaled at every layer to the norm of the aphasia vector, so that at the same $r$ they add a vector of the same norm.
We add them at the same 8 layers and 8 positive strengths, in 192 settings per model and task, and score them on the aphasia target.

\begin{table*}[t]
\centering
\footnotesize
\setlength{\tabcolsep}{5.2pt}
\begin{tabular}{@{}lrrrrrrrrrrrrrr@{}}
\toprule
 & \multicolumn{7}{c}{Cinderella task} & \multicolumn{7}{c}{Sandwich task} \\
\cmidrule(lr){2-8}\cmidrule(l){9-15}
 & \multicolumn{2}{c}{Aphasia} & \multicolumn{2}{c}{Broca} & \multicolumn{2}{c}{Wernicke} & & \multicolumn{2}{c}{Aphasia} & \multicolumn{2}{c}{Broca} & \multicolumn{2}{c}{Wernicke} & \\
\cmidrule(lr){2-3}\cmidrule(lr){4-5}\cmidrule(lr){6-7}\cmidrule(lr){9-10}\cmidrule(lr){11-12}\cmidrule(lr){13-14}
Model & $n$ & $r$ & $n$ & $r$ & $n$ & $r$ & Random & $n$ & $r$ & $n$ & $r$ & $n$ & $r$ & Random \\
\midrule
\gptxl & $2$ & $2.75$ & $3$ & $4$ & $0$ & -- & $5$ & $4$ & $1.25$ & $4$ & $2.25$ & $0$ & -- & $2$ \\
\addlinespace[2pt]
\llamaseven & $5$ & $2$ & $3$ & $3$ & $0$ & -- & $1$ & $4$ & $2$ & $1$ & $3$ & $0$ & -- & $1$ \\
\llamathirteen & $3$ & $2$ & $3$ & $3$ & $0$ & -- & $0$ & $3$ & $3$ & $4$ & $3$ & $0$ & -- & $3$ \\
\addlinespace[2pt]
\pythia{1B} & $8$ & $0.5$ & $6$ & $1$ & $4$ & $0.75$ & $8$ & $8$ & $1$ & $8$ & $1$ & $1$ & $0.5$ & $12$ \\
\pythia{1.4B} & $8$ & $0.5$ & $6$ & $1$ & $2$ & $0.75$ & $10$ & $6$ & $1$ & $3$ & $1$ & $2$ & $0.75$ & $10$ \\
\pythia{2.8B} & $8$ & $0.375$ & $5$ & $0.5$ & $4$ & $1$ & $16$ & $4$ & $0.5$ & $3$ & $1$ & $2$ & $1$ & $14$ \\
\pythia{6.9B} & $8$ & $0.25$ & $7$ & $0.25$ & $3$ & $0.5$ & $12$ & $4$ & $1$ & $3$ & $1.5$ & $2$ & $0.75$ & $14$ \\
\addlinespace[2pt]
\qwen{0.8B} & $4$ & $1$ & $2$ & $1.25$ & $2$ & $1.25$ & $2$ & $4$ & $1.5$ & $2$ & $1.5$ & $1$ & $1$ & $1$ \\
\qwen{2B} & $4$ & $1$ & $3$ & $1$ & $0$ & -- & $2$ & $6$ & $1.5$ & $3$ & $1.5$ & $1$ & $1$ & $3$ \\
\qwen{4B} & $3$ & $1$ & $2$ & $2.5$ & $1$ & $1.5$ & $2$ & $2$ & $1.5$ & $3$ & $1.5$ & $0$ & -- & $1$ \\
\qwen{9B} & $2$ & $1.5$ & $2$ & $1$ & $0$ & -- & $0$ & $3$ & $1.5$ & $2$ & $1.75$ & $0$ & -- & $0$ \\
\midrule
All 11 models & $55$ & $0.5$ & $42$ & $1$ & $16$ & $1$ & $58$ & $48$ & $1.5$ & $36$ & $1.5$ & $9$ & $1$ & $61$ \\
\bottomrule
\end{tabular}
\caption{Steering sweep of each model's answer to the task prompt with the aphasia, Broca, and Wernicke vectors of the same task, at 8 layers and 8 strengths from $r = 0.1$ to $4$, where $r$ is the norm of the added vector relative to the mean norm of the transcripts' token-averaged residual stream at that layer.
$n$: the number of the 8 layers at which some strength makes the classifier's probability of the vector's target exceed 0.5.
$r$: the median, over these layers, of the smallest such strength, or a dash where no layer works.
\emph{Random}: the number of the 192 settings, three random directions with the aphasia vector's norm at the same 8 layers and 8 strengths, at which \paph exceeds 0.5.
The last row sums $n$ over the 11 models, of 88 layers per task, takes the median smallest strength over all working layers, and sums the random settings, of 2,112 per task.}
\label{apptab:steering-sweep}
\end{table*}

\begin{figure*}[t]
    \centering
    \includegraphics[width=\textwidth]{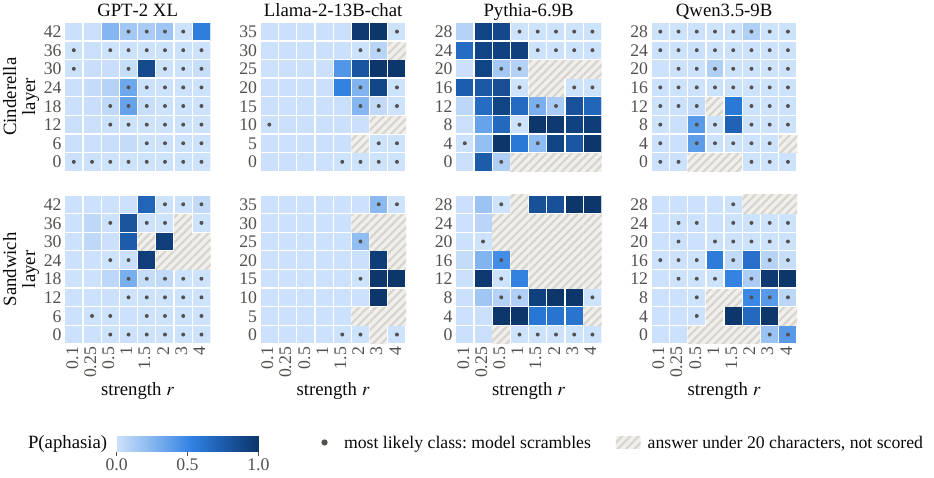}
    \caption{\paph of the answer to the task prompt when the aphasia vector of the same task is added at one of 8 layers, counted from 0, with one of 8 strengths $r$, in the largest model of each family.
    The 8 strengths run from $r = 0.1$ to $4$, where $r$ is the norm of the added vector relative to the mean norm of the human transcripts' token-averaged residual stream at that layer.
    The middle tick of the color bar, 0.5, is the threshold above which an answer reaches the aphasia target.
    A dot marks answers whose most probable class is model scrambles, and hatched cells mark answers shorter than 20 characters, which are not scored.
    The unsteered Cinderella answers of \gptxl and \qwen{9B} already have model scrambles as their most probable class.}
    \label{appfig:steering-sweep}
\end{figure*}

\paragraph{Sweep results.}
Table~\ref{apptab:steering-sweep} gives the sweep's results for every model and task, and Figure~\ref{appfig:steering-sweep} shows \paph for every setting of the aphasia vector in the largest model of each family.
Over both tasks, the aphasia vector works at 103 of the 176 layers: at 54 of 64 in the Pythia models, at a median strength of $r = 0.5$; at 28 of 64 in the Qwen3.5 models, at 1.5; at 15 of 32 in the Llama-2 models, at 2; and at 6 of 16 in \gptxl, at 1.5.
The Broca vector works at 78 of the 176 layers and the Wernicke vector at 25, and the Wernicke vector never works in \gptxl or the Llama-2 models.
Random directions reach the aphasia target in 119 of the 4,224 settings, or 2.8\%: in 6.25\% of the settings in the Pythia models, 1.8\% in \gptxl, and less than 1\% in the Llama-2 and Qwen3.5 models.
At every strength, the aphasia vector reaches its target more often than random directions, for example in 41 of the 176 settings at $r = 1$, against 27 of the 528 random settings.
At a given layer, the strengths that work lie between weaker ones, which leave \paph low, and stronger ones, which often turn the answer into model scrambles or into an answer too short to score (Figure~\ref{appfig:steering-sweep}).
Pushes toward control speech rarely reach a target: of the 528 settings with a negative strength, 13 reach the aphasia target, 3 the Broca target, and none the Wernicke target.
Without steering, \paph is at most 0.017, except for \pythia{6.9B} on the Cinderella task at 0.167, although the classifier's most probable class for the unsteered Cinderella answers of \gptxl, \pythia{2.8B}, and the four Qwen3.5 models is already model scrambles.
Unlike the lesion search, the sweep does not require the model to say that it knows the story of Cinderella, and steering changes that answer too.
Of the Cinderella settings that reach their target, 31 of 122 for aphasia, 23 of 89 for Broca, and 2 of 22 for Wernicke answer no to \say{Do you know the story of Cinderella?}, and 17 of these 31 aphasia settings come from \pythia{6.9B}, which answers no even without steering.

\begin{table*}[t]
\centering
\footnotesize
\setlength{\tabcolsep}{5pt}
\begin{tabular}{@{}llrrrrrrrrr@{}}
\toprule
 & & \multicolumn{3}{c}{Aphasia} & \multicolumn{3}{c}{Broca} & \multicolumn{3}{c}{Wernicke} \\
\cmidrule(lr){3-5}\cmidrule(lr){6-8}\cmidrule(l){9-11}
Model & Source task & Layer & $r$ & $P$ & Layer & $r$ & $P$ & Layer & $r$ & $P$ \\
\midrule
\gptxl & Cinderella & $30$ & $1.5$ & $0.865$ & $18$ & $1$ & $0.535$ & $30$ & $1$ & $0.276^{\dagger}$ \\
 & Sandwich & $36$ & $1$ & $0.803$ & $18$ & $1$ & $0.697$ & $36$ & $1$ & $0.422^{\dagger}$ \\
\addlinespace[2pt]
\llamaseven & Cinderella & $0$ & $1.5$ & $0.994$ & $12$ & $3$ & $0.957$ & $16$ & $3$ & $0.011^{\dagger}$ \\
 & Sandwich & $0$ & $1.5$ & $0.659$ & $12$ & $3$ & $0.995$ & $24$ & $3$ & $0.046^{\dagger}$ \\
\addlinespace[2pt]
\llamathirteen & Cinderella & $20$ & $1.5$ & $0.524$ & $20$ & $2$ & $0.570$ & $30$ & $3$ & $0.414^{\dagger}$ \\
 & Sandwich & $10$ & $3$ & $0.999$ & $20$ & $3$ & $0.994$ & $30$ & $3$ & $0.367^{\dagger}$ \\
\addlinespace[2pt]
\pythia{1B} & Cinderella & $8$ & $0.25$ & $0.833$ & $0$ & $1$ & $0.994$ & $8$ & $0.5$ & $0.824$ \\
 & Sandwich & $6$ & $0.25$ & $0.895$ & $4$ & $0.5$ & $0.635$ & $2$ & $0.5$ & $0.508$ \\
\addlinespace[2pt]
\pythia{1.4B} & Cinderella & $18$ & $0.25$ & $0.590$ & $9$ & $0.5$ & $0.963$ & $6$ & $0.5$ & $0.732$ \\
 & Sandwich & $15$ & $0.25$ & $0.611$ & $3$ & $1$ & $0.817$ & $12$ & $0.5$ & $0.725$ \\
\addlinespace[2pt]
\pythia{2.8B} & Cinderella & $20$ & $0.1$ & $0.712$ & $12$ & $0.5$ & $0.822$ & $20$ & $1$ & $0.993$ \\
 & Sandwich & $8$ & $0.5$ & $0.999$ & $4$ & $1$ & $0.995$ & $16$ & $0.5$ & $0.738$ \\
\addlinespace[2pt]
\pythia{6.9B} & Cinderella & $16$ & $0.1$ & $0.741$ & $16$ & $0.1$ & $0.641$ & $20$ & $0.5$ & $0.847$ \\
 & Sandwich & $12$ & $0.25$ & $0.975$ & $28$ & $1.5$ & $0.897$ & $4$ & $0.5$ & $0.820$ \\
\bottomrule
\end{tabular}
\caption{Steering settings of the human evaluation, for the seven models whose steered answers two experts labeled.
For each model, source task, and target, the vector comes from the transcripts of the source task, and the setting is the smallest strength $r$ at which the target probability of the steered answer to the source task exceeds 0.5 in the sweep, at the layer, counted from 0, with the highest probability at that strength.
$P$: the classifier's probability of the target for that answer.
$^{\dagger}$No setting of the sweep passes 0.5, so the setting with the highest probability is used.}
\label{apptab:steering-settings}
\end{table*}

\paragraph{Settings for the human evaluation.}
The human evaluation uses one setting for each of the seven models whose answers the experts labeled, each source task, and each target (Table~\ref{apptab:steering-settings}).
Among the positive strengths whose answer to the source task is long enough to score, we take the smallest strength at which the target probability exceeds 0.5, and if several layers pass 0.5 at that strength, the layer with the highest probability.
For the Wernicke target in \gptxl and the two Llama-2 models, no setting passes 0.5 on either task, so we take the setting with the highest \pwern, which ranges from 0.011 to 0.422.
The chosen setting then steers the answers to all seven prompts, without the question about knowing the story, with greedy decoding, a repetition penalty of 1.25, and at most 180 new tokens for the two narratives and 120 for the five TAB prompts.
On the source task, the classifier gives these answers the same probabilities as the sweep's answers in all 42 settings.
Appendix~\ref{appsec:human} describes how the experts labeled these answers.

\begin{figure*}[tp]
    \centering
    \includegraphics[width=0.9\textwidth]{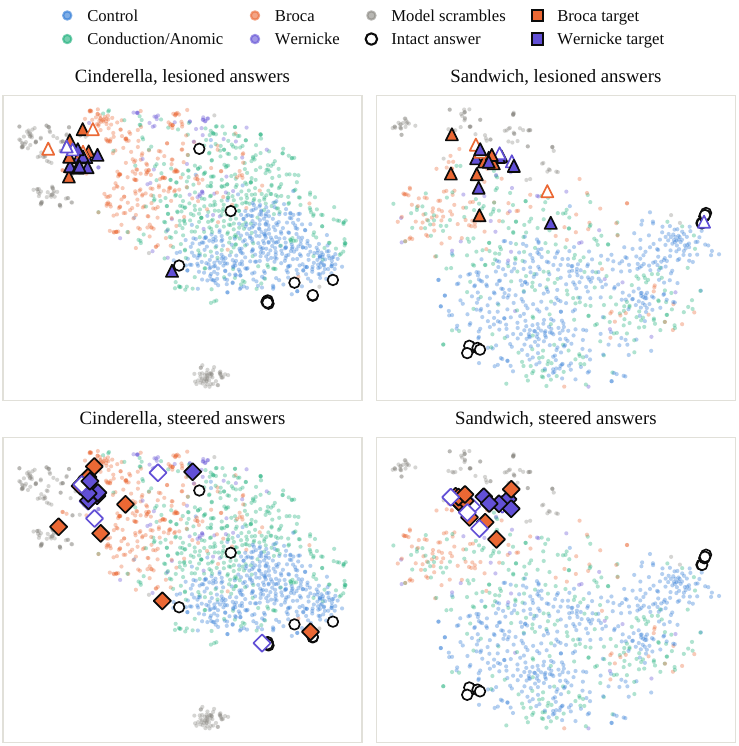}
    \caption{The models' answers among the transcripts and model scrambles of Figure~\ref{appfig:tsne-transcripts}, in one t-SNE per task fitted on all of them together.
    Each panel shows the intact answers of the 11 models as white circles, and either their answers lesioned by \ourmethod toward Broca's or Wernicke's aphasia, as triangles in the top row, or their answers steered toward these types, as diamonds in the bottom row.
    A hollow marker is an answer that does not reach its target.}
    \label{appfig:tsne-answers}
\end{figure*}

\paragraph{Steered Broca answers lie among Broca transcripts, lesioned ones near the model scrambles.}
We place the models' answers in the text space of Appendix~\ref{appsec:data}: each model's intact answer to the task prompt, the answers of its \ourmethod lesions for the Broca and Wernicke targets, and its answers steered toward the two types with the task's own vector, at the setting that the rule above chooses, for all 11 models (Figure~\ref{appfig:tsne-answers}).
For each answer, we count the groups of its 10 nearest texts among the transcripts and model scrambles.
The intact answers lie among control transcripts, which make up 82\% and 92\% of their nearest texts on the Cinderella and the sandwich task.
Answers steered toward Broca's aphasia have 57\% and 43\% Broca transcripts among their nearest texts, against 21\% and 11\% of all texts, and only 11\% and 12\% model scrambles.
Answers lesioned toward Broca's aphasia have 53\% and 37\% model scrambles among their nearest texts, against 13\% and 9\% of all texts, and only 33\% and 15\% Broca transcripts.
The lesions that reach the Broca target are no exception: the answers of the 7 such Cinderella lesions have 57\% model scrambles among their nearest texts.
The Wernicke answers of both methods lie between groups, but Wernicke transcripts, 4\% and 3\% of all texts, make up 26\% and 18\% of the nearest texts of the steered answers and 12\% and 26\% of those of the lesioned answers.
In a space that knows nothing of our labels, steering toward Broca's aphasia thus moves the answer toward Broca's speech, whereas lesioning moves it toward the machine failures that the classifier learns to set apart.
The encoder may, however, place short and fragmented speech near fragmented machine output, which the classifier's fine-tuning separates, so this map checks the classifier's labels rather than replacing them.

\paragraph{Continuations of human transcripts.}
Steering the model's own answer cannot show whether subtracting the aphasia direction moves speech toward control speech, because the unsteered answers are already far from aphasic.
We therefore start the answer with the first 25 words of a human transcript and let the model continue it for 120 new tokens, while we add the aphasia vector at layer $\mathrm{round}(L/4)$ with $r \in \{-2, -1, 0, 1, 2\}$, and we score only the continuation.
Each task uses the same 20 speakers in all 11 models: 10 control speakers and 10 speakers with aphasia, of whom 3 have anomic, 3 Broca's, 2 conduction, and 2 Wernicke's aphasia.
Adding the direction makes the continuations aphasic: of the 208 continuations of control speakers that the classifier scores but does not call aphasic without steering, 206 are long enough to score at $r = 1$, and 153 of these become aphasic, although in \gptxl only 3 of 18 do, and its continuations become model scrambles instead.
Subtracting the direction, however, does not make the continuations of aphasic speakers control-like.
Without steering, the classifier calls only 41 of the 220 continuations of aphasic speakers aphasic.
Of the 40 of these that are long enough to score at $r = -1$, subtracting the direction makes 32 no longer aphasic, but only 10 of them become control speech by the classifier's most probable class, while 22 become model scrambles.
At $r = -2$, subtracting the direction even turns 50 of the 208 control continuations that are scored and not aphasic without steering into aphasic ones, and the classifier calls 105 of the 220 control continuations model scrambles.

\paragraph{Cinderella $\leftrightarrow$ sandwich transfer of aphasia vector.}
The aphasia vector of one task also steers the answers to the other task.
We add the other task's vector at the same 8 layers and positive strengths, scaled so that it has the norm that the task's own vector would have at the same $r$.
It works at 98 of the 176 layers, against 103 for the task's own vector, and in 234 of the 1,408 settings, against 224, although the cosine between the two tasks' aphasia vectors ranges from 0.265 to 0.936 over models and layers.
It works at 54 of 64 layers in the Pythia models, as many as the task's own vector, and at slightly fewer layers than the task's own vector in the other families.

\paragraph{Type-specific vectors.}
The Broca and Wernicke vectors point in nearly the same direction as the aphasia vector, with cosines of 0.838 to 0.989 for the Broca vector and 0.548 to 0.975 for the Wernicke vector over models, tasks, and layers.
To steer toward what sets a type apart from aphasia in general, we remove from the Broca and Wernicke vectors their component along the aphasia vector at every layer and sweep these type-specific vectors as above.
By the classifier, the Wernicke-specific vector works at 94 of the 176 layers, against 25 for the full Wernicke vector, and in all 22 pairs of model and task, including the 10 in which the full vector never works.
Among these 10 are both tasks of \gptxl and of the two Llama-2 models, where the Wernicke-specific vector needs median strengths of 1.5 to 4.
The Broca-specific vector, in contrast, works at only 29 layers, against 78 for the full Broca vector.
The human evaluation uses the full vectors, so its Wernicke answers from \gptxl and the Llama-2 models come from settings that do not pass 0.5, even though the Wernicke-specific vector passes 0.5 in these models.

\begin{table*}[t]
\centering
\small
\setlength{\tabcolsep}{3.5pt}
\begin{tabular}{@{}lrrrrrrrrr@{}}
\toprule
 & \multicolumn{2}{c}{Broca against Wernicke} & \multicolumn{2}{c}{Same type, across tasks} & \multicolumn{2}{c}{Shuffled labels} & & \multicolumn{2}{c}{Type-specific parts} \\
\cmidrule(lr){2-3} \cmidrule(lr){4-5} \cmidrule(lr){6-7} \cmidrule(l){9-10}
Model & Cinderella & Sandwich & Broca & Wernicke & Cinderella & Sandwich & Below & Cinderella & Sandwich \\
\midrule
\gptxl & $0.90$ & $0.81$ & $0.62$ & $0.59$ & $0.99$ & $0.97$ & $96/96$ & $-0.47$ & $-0.43$ \\
\addlinespace[2pt]
\llamaseven & $0.87$ & $0.79$ & $0.66$ & $0.69$ & $0.99$ & $0.97$ & $64/64$ & $-0.48$ & $-0.42$ \\
\llamathirteen & $0.86$ & $0.79$ & $0.66$ & $0.69$ & $0.99$ & $0.97$ & $78/80$ & $-0.49$ & $-0.41$ \\
\addlinespace[2pt]
\pythia{1B} & $0.87$ & $0.80$ & $0.72$ & $0.63$ & $0.99$ & $0.97$ & $32/32$ & $-0.65$ & $-0.43$ \\
\pythia{1.4B} & $0.86$ & $0.83$ & $0.72$ & $0.64$ & $0.99$ & $0.97$ & $48/48$ & $-0.63$ & $-0.43$ \\
\pythia{2.8B} & $0.87$ & $0.82$ & $0.71$ & $0.69$ & $0.99$ & $0.97$ & $64/64$ & $-0.62$ & $-0.47$ \\
\pythia{6.9B} & $0.90$ & $0.81$ & $0.73$ & $0.65$ & $0.99$ & $0.97$ & $64/64$ & $-0.57$ & $-0.45$ \\
\addlinespace[2pt]
\qwen{0.8B} & $0.87$ & $0.81$ & $0.77$ & $0.66$ & $0.99$ & $0.98$ & $48/48$ & $-0.65$ & $-0.54$ \\
\qwen{2B} & $0.88$ & $0.83$ & $0.70$ & $0.64$ & $0.99$ & $0.98$ & $48/48$ & $-0.63$ & $-0.53$ \\
\qwen{4B} & $0.90$ & $0.82$ & $0.72$ & $0.69$ & $0.99$ & $0.98$ & $64/64$ & $-0.52$ & $-0.47$ \\
\qwen{9B} & $0.90$ & $0.83$ & $0.75$ & $0.72$ & $0.99$ & $0.98$ & $64/64$ & $-0.48$ & $-0.50$ \\
\bottomrule
\end{tabular}
\caption{Cosine similarity of steering vectors, as the median over a model's layers.
\emph{Broca against Wernicke}: the Broca and the Wernicke vector of the same task.
\emph{Same type, across tasks}: the Cinderella and the sandwich vector of one type.
\emph{Shuffled labels}: the Broca and the Wernicke vector after shuffling the two labels among the transcripts of both types, keeping the group sizes, averaged over 200 shuffles.
\emph{Below}: layers at which the observed Broca-against-Wernicke cosine lies below the 2.5th percentile of the shuffled cosines, over both tasks.
\emph{Type-specific parts}: the cosine of the Broca and the Wernicke vector after projecting the aphasia vector out of both.}
\label{apptab:subtype-cosine}
\end{table*}

\paragraph{The Broca and Wernicke vectors share most of their length.}
For every model, task, and layer, we compute the cosine similarity of the Broca and the Wernicke vector of the same task and compare it with two references (Table~\ref{apptab:subtype-cosine}).
The first reference is the same type's vector from the other task, which shows how much a vector changes when only the task changes.
The second reference shuffles the Broca and Wernicke labels among the transcripts of the two types, 200 times, keeping the group sizes, and recomputes both vectors.
Within a task, the Broca and the Wernicke vector have a median cosine over layers of 0.79 to 0.90 in the 22 pairs of a model and a task, and both lie close to the aphasia vector, at median cosines of 0.973 for the Broca vector and 0.915 for the Wernicke vector over models, tasks, and layers.
The same type's vectors from the two tasks have median cosines of only 0.72 for Broca and 0.66 for Wernicke, lower than those of the Broca and Wernicke vectors of one task, although this comparison favors the pair from one task, whose two vectors share the control transcripts and the prompt.
Shuffled labels give median cosines of 0.97 to 0.99, and the observed cosine falls below the 2.5th percentile of the shuffled cosines at 670 of the 672 layers, at every layer except the first two of \llamathirteen on the sandwich task.
With the aphasia vector projected out of both, the remaining parts of the Broca and Wernicke vectors have median cosines of $-0.65$ to $-0.41$, so the parts that set the two types apart point away from each other.
The Wernicke part carries its type: the Wernicke-specific vector works in all 22 pairs, as the previous paragraph shows, whereas the Broca-specific vector works in only 15.
The two types thus share most of their deviation from control speech, while each keeps a small part of its own that steering can isolate.

\section{Human Evaluation of Steered Answers}
\label{appsec:human}

We describe the annotation sheets on which two experts labeled the steered answers, defines the agreement measures of Table~\ref{tab:agreement}, and breaks the results of \S\ref{sec:results-steer} down by expert, class, model, prompt, and steering vector.

\paragraph{The annotation sheets.}
We steer seven models, \gptxl, \llamaseven, \llamathirteen, \pythia{1B}, \pythia{1.4B}, \pythia{2.8B}, and \pythia{6.9B}, toward each of the three targets with each of two steering vectors, one computed from the Cinderella transcripts and one from the sandwich transcripts.
Each steered model uses the layer and strength that the sweep of \S\ref{sec:method-steer} chose for its target on the vector's own task, the source task (Appendix~\ref{appsec:steering}), and answers seven prompts once.
The seven prompts are the requests of the two tasks and the five connected-text prompts of the TAB: \say{Tell me about the best trip you ever took}, \say{Describe a happy childhood memory}, \say{Tell me about your first job}, \say{What do you like about where you live?}, and \say{Describe the steps to make a simple meal}.
The models decode as in \S\ref{sec:setup}, except that they write at most 120 new tokens for the TAB prompts.
Seven models, seven prompts, two steering vectors, and three targets give 294 steered answers.
The answers steered toward aphasia form the two-way sheet, on which the experts label each answer as aphasic or not.
The answers steered toward Broca's or Wernicke's aphasia form the five-way sheet, on which the experts give each answer one of the classifier's five classes: control speech, conduction or anomic aphasia, Broca's aphasia, Wernicke's aphasia, or model scrambles.
We leave 22 answers out of the analysis, 7 of the 98 on the two-way sheet and 15 of the 196 on the five-way sheet: 18 answers with fewer than 20 characters, 8 of them empty, which the classifier does not score (\S\ref{sec:setup}), and 4 answers for which the LLM judge returned an empty reply.
The analysis thus covers 272 answers, 91 on the two-way sheet and 181 on the five-way sheet, of which 90 are steered toward Broca's aphasia and 91 toward Wernicke's aphasia.

\paragraph{Expert labels.}
We take expert~1's labels from the annotation sheets, and expert~2's from a tab of the sheets in which expert~2 labeled the answers before the experts compared notes.
The sheets also hold a later version of expert~2's labels, set after the experts compared notes, and on all 46 answers where the two versions differ, the later version equals expert~1's label.
With the later version, the experts would agree on 96\% of the answers, at a Cohen's $\kappa$ of 0.95, rather than on 80\%, at a $\kappa$ of 0.76.
Expert~2's tab has no label for 4 answers, 2 on each sheet, so we compare these answers with expert~1 alone and leave them out of every statistic that needs both experts.
The two-way sheet has no label for model scrambles, so an answer on it counts as a model scramble when the experts' comments mark it as one, which holds for 25 of its 91 answers.
On the five-way sheet, an answer counts as a model scramble when either expert labels it so, which holds for 52 of its 181 answers.

\paragraph{Agreement measures.}
The classifier and the LLM judge label the same answers as the experts.
On the two-way sheet, the classifier labels an answer aphasic when \paph exceeds 0.5, and the judge answers yes or no to whether the speaker has aphasia.
On the five-way sheet, the classifier gives its most probable class, and the judge names one of the five classes (Appendix~\ref{appsec:judge}).
A labeler agrees with the experts on an answer when its label matches the label of either expert, which is the agreement of Table~\ref{tab:agreement}.
As a baseline, the best constant label gives every answer of a sheet the one label that agrees with the experts most often: not aphasic on the two-way sheet and Broca's aphasia on the five-way sheet.
It agrees with the experts on 63\% of the two-way answers, 39\% of the five-way answers, and 47\% of all answers, against 73\%, 62\%, and 65\% for the classifier and 64\%, 51\%, and 55\% for the judge.
Cohen's $\kappa$ corrects the agreement of two labelers for the agreement expected by chance~\citep{cohen1960coefficient}: ${\kappa = (p_o - p_e) / (1 - p_e)}$, where $p_o$ is the share of answers on which the two agree and $p_e$ the share on which two independent labelers with the same label frequencies would agree.
By the scale of \citet{landis1977measurement}, a $\kappa$ from 0.21 to 0.40 is fair agreement, from 0.41 to 0.60 moderate, from 0.61 to 0.80 substantial, and above 0.80 almost perfect.
On all answers, $\kappa$ pools the two sheets and treats their labels as one set of seven.

\begin{table}[t]
\centering
\small
\setlength{\tabcolsep}{14pt}
\begin{tabular}{@{}lrrr@{}}
\toprule
 & Two-way & Five-way & All \\
\midrule
\multicolumn{4}{@{}l}{\textit{Against expert 1}} \\
Classifier & $\pmb{64}$ & $\pmb{48}$ & $\pmb{53}$ \\
\quad Cohen's $\kappa$ & $0.25$ & $0.33$ & $0.44$ \\
LLM judge & $62$ & $41$ & $48$ \\
\quad Cohen's $\kappa$ & $0.22$ & $0.21$ & $0.38$ \\
\midrule
\multicolumn{4}{@{}l}{\textit{Against expert 2}} \\
Classifier & $\pmb{71}$ & $\pmb{57}$ & $\pmb{62}$ \\
\quad Cohen's $\kappa$ & $0.38$ & $0.43$ & $0.54$ \\
LLM judge & $55$ & $48$ & $50$ \\
\quad Cohen's $\kappa$ & $0.09$ & $0.26$ & $0.39$ \\
\midrule
\multicolumn{4}{@{}l}{\textit{On the experts' consensus}} \\
Classifier & $\pmb{71}$ & $\pmb{57}$ & $\pmb{62}$ \\
LLM judge & $60$ & $51$ & $54$ \\
Answers & $78$ & $136$ & $214$ \\
\midrule
\multicolumn{4}{@{}l}{\textit{Classifier against LLM judge}} \\
Agreement & $52$ & $37$ & $42$ \\
\quad Cohen's $\kappa$ & $0.00$ & $0.19$ & $0.31$ \\
\bottomrule
\end{tabular}
\caption{Agreement of the classifier and the LLM judge with each of the two experts and with the experts' consensus, in percent, on the steered answers.
\emph{Two-way}: answers steered toward aphasia, labeled as aphasic or not.
\emph{Five-way}: answers steered toward Broca's or Wernicke's aphasia, labeled with the classifier's five classes.
\emph{All}: the answers of both sheets together, on which Cohen's $\kappa$ treats the labels of the two sheets as one set of seven.
\emph{Consensus}: the answers to which both experts give the same label.
Comparisons with expert 2 leave out the 4 answers that expert 2 did not label.
Bold marks the higher agreement of the classifier and the LLM judge.}
\label{apptab:human-agreement}
\end{table}

\paragraph{Agreement with each expert.}
Table~\ref{apptab:human-agreement} compares the classifier and the judge with each expert separately and with the experts' consensus, the answers to which both experts give the same label.
The classifier agrees more with expert~2 than with expert~1, on 62\% against 53\% of all answers, at $\kappa = 0.54$ and $0.44$, whereas the judge agrees with the two about equally, on 50\% and 48\%, at $\kappa = 0.39$ and $0.38$.
By the scale of \citet{landis1977measurement}, the classifier's agreement with each expert is moderate and the judge's is fair, whereas the experts' agreement with each other, at $\kappa = 0.76$, is substantial.
On the 214 answers on which the experts agree, the classifier gives their label to 62\% and the judge to 54\%.
The classifier and the judge agree with each other on only 42\% of all answers, at $\kappa = 0.31$, and on the two-way sheet at $\kappa = 0.00$, which is no better than chance.

\paragraph{Model scrambles.}
The classifier recognizes model scrambles much better on the two-way sheet than on the five-way sheet.
On the two-way sheet, the classifier's label shows only whether an answer is aphasic, so we use its most probable class: it puts 27 answers in the model-scramble class, 20 of which the experts' comments mark as model scrambles, a precision of 74\%, and it finds 20 of the 25 marked answers, a recall of 80\%.
On the five-way sheet, its precision is 52\%, 16 of 31, and its recall 31\%, 16 of 52, and over both sheets, its precision is 62\% and its recall 47\%.
The judge can name model scrambles only on the five-way sheet, where it calls 4 answers model scrambles, 2 of them correctly, a precision of 50\% and a recall of 4\%.
On the two-way sheet, the judge calls 20 of the 25 model scrambles aphasic, whereas the classifier's label calls 5 of them aphasic.
Of the 52 five-way answers that either expert calls model scrambles, the classifier labels 30 as Broca's aphasia, and the judge labels 28 as Wernicke's aphasia and 19 as Broca's aphasia.
Both experts call 22 of these answers model scrambles, and the classifier finds 8 of them and labels the other 14 as Broca's aphasia.
On the source task, the classifier calls all 14 answers steered toward Broca's aphasia Broca-like, because its own score chose their steering settings, whereas the experts call 9 of them model scrambles.
The experts disagree with each other mostly about model scrambles.
Of their 43 disagreements on the five-way sheet, 29 involve model scrambles, and in 27 of these, expert~1 labels a model scramble where expert~2 labels a type of aphasia: Broca's aphasia 18 times, Wernicke's aphasia 8 times, and conduction or anomic aphasia once.
Overall, expert~1 labels 50 of the 181 five-way answers as model scrambles, and expert~2 labels 24 of the 179 that expert~2 labeled.
On the two-way sheet, the experts disagree on 11 of the 89 answers that both labeled.

\begin{table}[t]
\centering
\small
\setlength{\tabcolsep}{2.5pt}
\begin{tabular}{@{}>{\raggedright\arraybackslash}p{1.65cm}rrrrrrr@{}}
\toprule
 & & \multicolumn{3}{c}{Classifier} & \multicolumn{3}{c}{LLM judge} \\
\cmidrule(lr){3-5} \cmidrule(l){6-8}
Class & Experts & Calls & Prec. & Rec. & Calls & Prec. & Rec. \\
\midrule
Control & $16$ & $18$ & $50$ & $\pmb{56}$ & $14$ & $\pmb{57}$ & $50$ \\
Conduction or anomic & $22$ & $28$ & $\pmb{36}$ & $\pmb{45}$ & $12$ & $17$ & $9$ \\
Broca & $70$ & $68$ & $\pmb{74}$ & $\pmb{71}$ & $40$ & $57$ & $33$ \\
Wernicke & $64$ & $36$ & $\pmb{75}$ & $42$ & $111$ & $51$ & $\pmb{89}$ \\
Model scrambles & $52$ & $31$ & $\pmb{52}$ & $\pmb{31}$ & $4$ & $50$ & $4$ \\
\bottomrule
\end{tabular}
\caption{Precision and recall of each class, in percent, for the classifier and the LLM judge on the 181 answers steered toward Broca's or Wernicke's aphasia.
\emph{Experts}: the answers that either expert puts in the class; an answer on which the experts disagree counts in both of their classes.
\emph{Calls}: the answers to which the classifier or the judge gives the class.
\emph{Prec.}: precision, the share of the calls that either expert shares.
\emph{Rec.}: recall, the share of the experts' answers in the class that the classifier or the judge also puts there.
Bold marks the higher value of the two.}
\label{apptab:human-perclass}
\end{table}

\begin{table}[t]
\centering
\small
\setlength{\tabcolsep}{4pt}
\begin{tabular}{@{}>{\raggedright\arraybackslash}p{1.75cm}rrrrrr@{}}
\toprule
 & \multicolumn{5}{c}{Classifier} & \\
\cmidrule(lr){2-6}
Expert 2 & Ctrl. & Cond. & Broca & Wern. & MS & Total \\
\midrule
Control & $\pmb{8}$ & $3$ & $0$ & $0$ & $2$ & $13$ \\
Conduction or anomic & $6$ & $\pmb{9}$ & $0$ & $3$ & $1$ & $19$ \\
Broca & $0$ & $1$ & $\pmb{49}$ & $6$ & $8$ & $64$ \\
Wernicke & $4$ & $14$ & $3$ & $\pmb{27}$ & $11$ & $59$ \\
Model scrambles & $0$ & $0$ & $15$ & $0$ & $\pmb{9}$ & $24$ \\
\midrule
Total & $18$ & $27$ & $67$ & $36$ & $31$ & $179$ \\
\bottomrule
\end{tabular}
\caption{Labels of the classifier against those of expert 2 on the 179 answers steered toward Broca's or Wernicke's aphasia that expert 2 labeled.
Rows give expert 2's label and columns the classifier's, and the bold diagonal counts the 102 answers on which the two agree.
\emph{Ctrl.}: control speech; \emph{Cond.}: conduction or anomic aphasia; \emph{Wern.}: Wernicke's aphasia; \emph{MS}: model scrambles.}
\label{apptab:human-confusion}
\end{table}

\paragraph{Agreement by class.}
On the five-way sheet, the classifier is more precise than the judge for every class except control speech, and it recalls more answers of every class except Wernicke's aphasia (Table~\ref{apptab:human-perclass}).
The judge recalls 57 of the 64 answers that either expert labels as Wernicke's aphasia, 89\%, but it gives that label to 111 of the 181 answers, and it recalls only 23 of the 70 Broca answers, 33\%.
Against expert~2, the classifier's most frequent errors are 15 model scrambles that it labels as Broca's aphasia, 14 Wernicke answers that it labels as conduction or anomic aphasia, and 11 Wernicke answers that it labels as model scrambles (Table~\ref{apptab:human-confusion}).
Against expert~1, who calls more answers model scrambles, the classifier labels 29 of expert~1's 50 model scrambles as Broca's aphasia, and it agrees with expert~1 on 87 of 181 answers, 48\%, against 102 of 179, 57\%, with expert~2.

\begin{table*}[t]
\centering
\small
\setlength{\tabcolsep}{6pt}
\begin{tabular}{@{}lrrrrrrrr@{}}
\toprule
 & \multicolumn{2}{c}{Aphasia} & \multicolumn{2}{c}{Broca} & \multicolumn{2}{c}{Wernicke} & \multicolumn{2}{c}{All three} \\
\cmidrule(lr){2-3} \cmidrule(lr){4-5} \cmidrule(lr){6-7} \cmidrule(l){8-9}
Model & w/o MS & with MS & w/o MS & with MS & w/o MS & with MS & w/o MS & with MS \\
\midrule
\gptxl & $6/7$ & $9/14$ & $7/7$ & $7/14$ & $5/13$ & $5/14$ & $18/27$ & $21/42$ \\
\llamaseven & $2/2$ & $2/9$ & $6/6$ & $7/12$ & $2/7$ & $3/12$ & $10/15$ & $12/33$ \\
\llamathirteen & $4/9$ & $4/13$ & $5/9$ & $7/13$ & $1/6$ & $1/9$ & $10/24$ & $12/35$ \\
\pythia{1B} & $7/12$ & $8/14$ & $0/4$ & $6/14$ & $11/13$ & $12/14$ & $18/29$ & $26/42$ \\
\pythia{1.4B} & $9/13$ & $9/13$ & $8/8$ & $13/14$ & $9/14$ & $9/14$ & $26/35$ & $31/41$ \\
\pythia{2.8B} & $5/11$ & $5/14$ & $9/12$ & $11/14$ & $9/11$ & $10/14$ & $23/34$ & $26/42$ \\
\pythia{6.9B} & $6/12$ & $8/14$ & $1/7$ & $1/9$ & $11/12$ & $13/14$ & $18/31$ & $22/37$ \\
\midrule
All models & $39/66$ & $45/91$ & $36/53$ & $52/90$ & $48/76$ & $53/91$ & $123/195$ & $150/272$ \\
\bottomrule
\end{tabular}
\caption{Steering success by model: the number of steered answers that either expert labels with the steering target, out of the model's labeled answers for that target, without and with the answers that the experts call model scrambles (MS).
Each model has at most 14 answers per target, to 7 prompts with 2 steering vectors, one from the Cinderella and one from the sandwich transcripts, and loses the answers that are too short to score or that the LLM judge left without a reply.
An answer that one expert labels with the target and the other calls a model scramble counts as a success with MS and is left out without MS, so a model can succeed more often with MS.
Over all models, either expert labels the target for 59\%, 68\%, and 63\% of the answers steered toward aphasia, Broca's aphasia, and Wernicke's aphasia without MS.}
\label{apptab:human-bymodel}
\end{table*}

\begin{table*}[t]
\centering
\small
\setlength{\tabcolsep}{5pt}
\begin{tabular}{@{}>{\raggedright\arraybackslash}p{0.28\textwidth}rrrrrrrr@{}}
\toprule
 & \multicolumn{2}{c}{Aphasia} & \multicolumn{2}{c}{Broca} & \multicolumn{2}{c}{Wernicke} & \multicolumn{2}{c}{All three} \\
\cmidrule(lr){2-3} \cmidrule(lr){4-5} \cmidrule(lr){6-7} \cmidrule(l){8-9}
Prompt & w/o MS & with MS & w/o MS & with MS & w/o MS & with MS & w/o MS & with MS \\
\midrule
Tell me as much of the story of Cinderella as you can. & $9/9$ & $11/12$ & $6/7$ & $8/14$ & $7/10$ & $9/13$ & $22/26$ & $28/39$ \\
Tell me how you would make a peanut butter and jelly sandwich. & $7/9$ & $7/13$ & $1/3$ & $6/14$ & $8/11$ & $8/14$ & $16/23$ & $21/41$ \\
Tell me about the best trip you ever took. & $3/11$ & $4/14$ & $7/11$ & $8/13$ & $5/9$ & $7/12$ & $15/31$ & $19/39$ \\
Describe a happy childhood memory. & $6/10$ & $7/12$ & $9/12$ & $10/13$ & $9/14$ & $9/14$ & $24/36$ & $26/39$ \\
Tell me about your first job. & $7/9$ & $7/14$ & $6/9$ & $7/12$ & $5/9$ & $6/12$ & $18/27$ & $20/38$ \\
What do you like about where you live? & $2/8$ & $4/13$ & $3/6$ & $6/12$ & $6/11$ & $6/12$ & $11/25$ & $16/37$ \\
Describe the steps to make a simple meal. & $5/10$ & $5/13$ & $4/5$ & $7/12$ & $8/12$ & $8/14$ & $17/27$ & $20/39$ \\
\midrule
All prompts & $39/66$ & $45/91$ & $36/53$ & $52/90$ & $48/76$ & $53/91$ & $123/195$ & $150/272$ \\
\bottomrule
\end{tabular}
\caption{Steering success by prompt: the number of steered answers that either expert labels with the steering target, out of the labeled answers to the prompt for that target, without and with the answers that the experts call model scrambles (MS).
The first two prompts are those of the Cinderella and sandwich tasks, and the other five are the connected-text prompts of the Text Aphasia Battery.
Each prompt has at most 14 answers per target, from 7 models with 2 steering vectors, one from the Cinderella and one from the sandwich transcripts.
For the Cinderella and sandwich prompts, 7 of these 14 come from the vectors of the same task, whose answers chose the steering layer and strength.
An answer that one expert labels with the target and the other calls a model scramble counts as a success with MS and is left out without MS.}
\label{apptab:human-byprompt}
\end{table*}

\paragraph{Steering success by model, prompt, and vector.}
Steering toward Wernicke's aphasia reaches the target mostly in the Pythia models, for 40 of their 50 answers that are not model scrambles, against 8 of 26 in \gptxl and the Llama-2 models, for which no Wernicke setting passed 0.5 on the source task (Table~\ref{apptab:human-bymodel}).
Steering toward Broca's aphasia, in contrast, fails in \pythia{1B} and \pythia{6.9B}, for 0 of 4 and 1 of 7 such answers, and 10 of the 14 Broca answers of \pythia{1B} are model scrambles.
Over all seven models, steering toward Broca's aphasia breaks the most answers: the experts call 37 of its 90 answers model scrambles, against 25 of 91 for aphasia and 15 of 91 for Wernicke's aphasia.
By prompt, steering reaches the target most often on the Cinderella prompt, for 22 of 26 answers that are not model scrambles, and least often on the prompts about the best trip and about where the speaker lives, for 15 of 31 and 11 of 25 (Table~\ref{apptab:human-byprompt}).
Steering vectors computed from the sandwich transcripts reach the target more often than those computed from the Cinderella transcripts.
Without model scrambles, either expert labels the target for 79\%, 84\%, and 74\% of the answers steered toward aphasia, Broca's aphasia, and Wernicke's aphasia with sandwich vectors, against 45\%, 54\%, and 54\% with Cinderella vectors.
With model scrambles, the gap narrows to 56\%, 60\%, and 66\% against 43\%, 55\%, and 51\%, and toward aphasia, sandwich vectors produce more model scrambles, 17 of 45 answers against 8 of 46.

\paragraph{Steering carries its type to the prompts that did not choose its setting.}
The sweep chose each steering setting on the source task alone, so the answers to the other narrative and to the five TAB prompts test whether the targeted type carries over.
Without model scrambles, either expert labels the target for 5 of the 5 Broca answers and 8 of the 12 Wernicke answers to the source task, and for 31 of the 48 Broca answers, 65\%, and 40 of the 64 Wernicke answers, 63\%, to the other six prompts.
With model scrambles, the labels on these six prompts follow the steering vector: either expert labels 44 of the 76 answers steered toward Broca's aphasia as Broca's aphasia and only 9 as Wernicke's aphasia, and 44 of the 77 answers steered toward Wernicke's aphasia as Wernicke's aphasia and only 13 as Broca's aphasia.
The two vectors thus produce different types on prompts that played no part in choosing them.

\begin{table*}[t]
\centering
\small
\setlength{\tabcolsep}{4pt}
\begin{tabular}{@{}>{\raggedright\arraybackslash}p{0.17\textwidth}>{\raggedright\arraybackslash}p{0.1\textwidth}>{\raggedright\arraybackslash}p{0.11\textwidth}>{\raggedright\arraybackslash}p{0.11\textwidth}>{\raggedright\arraybackslash}p{0.44\textwidth}@{}}
\toprule
Model and vector & Prompt & Classifier & LLM judge & Answer \\
\midrule
\multicolumn{5}{@{}l}{\textit{Steered toward aphasia, labeled aphasic by both experts}} \\
\gptxl, Cinderella vector & Cinderella$^{\dagger}$ & Aphasic ($0.87$) & Aphasic & \textit{I don,t know that I like to tell it. It's a good thing I didn. / I was in my. . .} \\
\midrule
\multicolumn{5}{@{}l}{\textit{Steered toward Broca's aphasia, labeled Broca by both experts}} \\
\llamathirteen, Cinderella vector & Simple meal & Model scrambles ($0.39$) & Broca & \textit{Sure! Here are the steps to make a simple meal: / First, choose a recipe you like and ingredients list need . go shopping groc yes. Next, preheat oven or stove top cooking no. Then, chop vegetables car yes. After that, cook meat dish yes. Finally, serve hot enjoy eat.} \\
\llamaseven, Cinderella vector & First job & Broca ($0.71$) & Broca & \textit{sure! my job is good. okay.} \\
\midrule
\multicolumn{5}{@{}l}{\textit{Steered toward Wernicke's aphasia, labeled Wernicke by both experts}} \\
\llamaseven, Cinderella vector & First job & Wernicke ($0.60$) & Conduction or anomic & \textit{sure, my first job was a lot of fun. I like you know what it's good for. It’s good for you.} \\
\pythia{1B}, Cinderella vector & Cinderella$^{\dagger}$ & Wernicke ($0.82$) & Wernicke & \textit{I think it's a very good one, and that is what we are doing now in this country. We have been trying to do something for years. It was not so long ago when they were saying about the other thing. But there is no reason why we should be afraid because if he will say yes or no? Yes. No. And then also with him. So yeah. Yeah.} \\
\midrule
\multicolumn{5}{@{}l}{\textit{Steered toward Broca's aphasia, called model scrambles by both experts}} \\
\llamaseven, Cinderella vector & Cinderella$^{\dagger}$ & Broca ($0.96$) & Broca & \textit{sure! The story of Cinder. / Cinderelllll. / Okay, let's go. / okay okay. / Sure. / okay. / okay. / okay. / okay. / okay. / okay. / okay. ok. / ok. / ok. / ok. / ok. / ok. / ok. / ok. / ok. / ok. / ok. / ok. / ok. / ok. / ok. / ok. / ok. / ok. / ok. / ok. / ok. / ok. / ok. / ok. / ok. / ok. / ok.} \\
\pythia{6.9B}, sandwich vector & Cinderella & Broca ($0.94$) & Broca & \textit{Question \textup{[171 line breaks]} yes / no} \\
\bottomrule
\end{tabular}
\caption{Steered answers, verbatim, with the labels of the classifier and the LLM judge.
In the first five, both experts label the answer with the steering target; in the last two, both experts call the answer a model scramble, while the classifier labels it Broca.
\emph{Vector}: the task whose transcripts gave the steering vector, and whose answers chose the steering layer and strength; $^{\dagger}$ marks an answer to that task, the source task.
\emph{Prompt}: the Cinderella or sandwich task, or the topic of a connected-text prompt of the Text Aphasia Battery, \say{Describe the steps to make a simple meal} or \say{Tell me about your first job}.
The classifier's probability for its label follows the label; for the answer steered toward aphasia it is \paph.
A slash marks one or more line breaks, and the bracketed text replaces 171 line breaks.}
\label{apptab:human-examples}
\end{table*}

\paragraph{Examples.}
Table~\ref{apptab:human-examples} shows seven steered answers.
In the first five, both experts label the answer with the steering target.
The two answers of \llamaseven to the prompt about its first job show how the targets differ: steered toward Broca's aphasia, it answers in six words, and steered toward Wernicke's aphasia, it answers fluently but with little content.
In the last two, both experts call the answer a model scramble, while the classifier and the judge both label it as Broca's aphasia: one ends in a loop of \say{okay.} and \say{ok.}, and the other is the word \say{Question}, 171 line breaks, \say{yes}, and \say{no}.
The first of the two is an answer to the source task, so the classifier's score for it, $\pbroca = 0.96$, is the score with which the sweep chose the steering setting that produced it.

\section{Steering Back}
\label{appsec:rescue}

We describe the steering-back test of \S\ref{sec:method-steer} and reports its results.
The test asks whether removing a lesion's own direction from the lesioned model, or adding the direction with a negative strength, undoes the lesion, and whether random directions do the same.
Random directions undo the lesions almost as often as the own direction, so the direction of a type does not reverse a lesion specifically.
Most undone answers are model scrambles rather than the intact model's speech, so steering back undoes the lesions without healing them.

\paragraph{Random directions serve as controls.}
We test the 54 \ourmethod lesions that reach their target (Figure~\ref{fig:grown-heads}).
An answer reaches the target when the classifier's target probability exceeds 0.5 or when the target is its most probable class, where any of the three aphasia classes counts for the aphasia target.
A lesion's own direction is the steering vector $\vvec_\ell$ of its target (Equation~\eqref{eq:steering-vector}), computed on the task where the search found the lesion.
For the aphasia target, the own direction is the aphasia vector of Appendix~\ref{appsec:steering}, which pools the transcripts of the four aphasic groups.
The controls are three Gaussian random directions, and the same three serve the lesions of all three targets of a model and task.
For each lesion, the model answers the prompt of that task under 135 conditions, with the greedy decoding and the scoring of \S\ref{sec:setup}.
Seven answers are references: those of the intact model, of the intact model with the own direction removed, of the lesioned model, of the lesioned model with the own direction removed, and of the lesioned model with each of the three random directions removed.
The other 128 answers come from pushes, one for each combination of four directions, eight layers, and four strengths.
A push adds the own direction or one of the three random directions to the lesioned model after one of eight layers spread across the model's depth, with a strength of $r = -0.5$, $-1$, $-2$, or $-4$.
We call a push with the own direction an own push, and a push with a random direction a random push.
If a lesion acts through its own direction, steering back along that direction should undo it more often than steering back along a random direction.

\paragraph{Pushes match the directions in size, but removal does not.}
Following \citet{arditi2024refusal}, removal projects a direction's unit vector out of the residual stream after every block (\S\ref{sec:method-steer}).
Removal takes out whatever part of the residual stream lies along the direction, so the amount that it takes out depends on the direction, and we did not record this amount for either kind of direction.
A push adds a multiple of the direction after one layer $\ell$, scaled so that the added vector has the norm $|r|\, \bar h_\ell$, with $r$ and $\bar h_\ell$ as in \S\ref{sec:method-steer}.
The added vector thus has the same norm for every direction, so pushes compare the own and the random directions at equal size.
A removal or a push undoes the lesion when its answer no longer reaches the target, that is, when its target probability is 0.5 or below and the target is not its most probable class.
An answer shorter than 20 characters does not count by the rules of \S\ref{sec:setup}, so we do not score it, and such an answer does not undo the lesion.

\begin{figure}[t]
    \centering
    \includegraphics[width=\columnwidth]{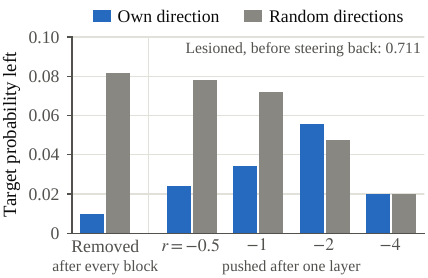}
    \caption{Mean target probability that steering back leaves in the 54 \ourmethod lesions, for the lesion's own direction and for three random directions.
    \emph{Removed}: removal of the direction from the residual stream after every block.
    $r$: the strength of a push, which adds the direction after one of eight layers; own and random pushes add vectors of the same norm.
    Each mean pools the lesions, the layers of a push, and the random directions, over the answers long enough to score.
    Before steering back, the lesioned answers have a mean target probability of 0.711, and a lesion is undone when its answer no longer reaches the target: its target probability falls to 0.5 or below and the target is not its most probable class.}
    \label{appfig:rescue}
\end{figure}

\begin{table}[t]
\centering
\small
\setlength{\tabcolsep}{3pt}
\begin{tabular}{@{}lrrrr@{\hspace{12pt}}rr@{}}
\toprule
 & & \multicolumn{3}{c}{Removal undoes} & \multicolumn{2}{c}{Probability left} \\
\cmidrule(lr){3-5}\cmidrule(l){6-7}
 & & & \multicolumn{2}{c}{Random} & & \\
\cmidrule(lr){4-5}
 & Sets & Own & $\geq 1$ & All $3$ & Own & Random \\
\midrule
\multicolumn{7}{@{}l}{\textit{By model family}} \\
GPT-2 & $5$ & $5$ & $5$ & $4$ & $0.000$ & $0.052$ \\
Llama-2-chat & $11$ & $10$\rlap{$^{\dagger}$} & $11$ & $11$ & $0.029$ & $0.028$ \\
Pythia & $22$ & $22$ & $22$ & $16$ & $0.005$ & $0.133$ \\
Qwen3.5 & $16$ & $16$ & $16$ & $14$ & $0.006$ & $0.056$ \\
\midrule
\multicolumn{7}{@{}l}{\textit{By target}} \\
Aphasia & $22$ & $22$ & $22$ & $19$ & $0.007$ & $0.094$ \\
Broca & $15$ & $15$ & $15$ & $13$ & $0.004$ & $0.038$ \\
Wernicke & $17$ & $16$\rlap{$^{\dagger}$} & $17$ & $13$ & $0.019$ & $0.103$ \\
\midrule
All & $54$ & $53$\rlap{$^{\dagger}$} & $54$ & $45$ & $0.010$ & $0.081$ \\
\bottomrule
\end{tabular}
\caption{Removing a direction after every block of the 54 \ourmethod lesions, by model family and by target.
\emph{Removal undoes}: the number of lesions whose answer no longer reaches the target, a target probability of 0.5 or below with the target not the most probable class, when the own direction is removed, and when at least one, or all three, of three random directions are removed.
\emph{Probability left}: the mean target probability after removal, against 0.711 before it.
$^{\dagger}$With its own direction removed, the remaining lesion, of \llamathirteen on the sandwich task with the Wernicke target, gives an answer too short to score.}
\label{apptab:rescue}
\end{table}

\paragraph{Own and random directions both undo the lesions.}
Before steering back, the lesioned answers have a mean target probability of 0.711, and the intact answers have a mean of 0.006.
Removing the own direction undoes 53 of the 54 lesions, and the remaining lesion, the \llamathirteen sandwich Wernicke lesion, gives an answer too short to score (Table~\ref{apptab:rescue}).
The three random removals undo 51, 50, and 50 of the 54 lesions.
At least one of them undoes each of the 54 lesions, and all three undo 45 of them.
The own direction leaves less of the target probability: after removal, a mean of 0.010, against 0.081 for the random directions (Figure~\ref{appfig:rescue}).
Removal of the own direction leaves less than random removal in every model family and target except the Llama-2 models, where the two leave 0.029 and 0.028.
Pushes, which match the directions in size, keep this order only at weak strengths, with 0.024 against 0.078 at $r = -0.5$ and 0.034 against 0.072 at $r = -1$, and lose it at strong ones, with 0.055 against 0.047 at $r = -2$ and 0.020 for both at $r = -4$.
At every strength and at every one of the eight layers, most pushes undo the lesion, own pushes slightly more often than random ones: 92 to 97\% against 89 to 94\% across the strengths, and 87 to 99\% against 83 to 96\% across the layers.
Over all strengths and layers, 1,643 of the 1,728 own pushes and 4,720 of the 5,184 random pushes undo the lesion, or 95\% and 91\%.
A random direction thus undoes a lesion almost as often as the lesion's own direction, so undoing a lesion does not show that the lesion acts through the direction of its type.

\paragraph{The undone answers are not the intact answers.}
Undoing a lesion asks only that the answer no longer reach the target, so an undone answer need not be the intact model's speech.
No removal or push reproduces the intact answer: none of the 54 own removals, 162 random removals, 1,728 own pushes, and 5,184 random pushes yields the intact text.
Removing the own direction undoes the lesions mostly by breaking the answers.
The classifier's label, its most probable class, is model scrambles for 52 of the 53 undone answers and control speech for none of them.
The remaining answer, of the 276-head \llamaseven Cinderella Wernicke lesion, stays aphasic, and its label moves from Wernicke's aphasia to conduction or anomic aphasia.
To check such labels without the classifier, we call an answer fluent when it passes two tests.
First, compressing the answer with zlib must leave at least a quarter of its length, which rules out answers that loop.
Second, at least 80\% of its non-space characters must belong to words of an American English word list.
This sense of fluent concerns broken text, not the clinical fluency of Appendix~\ref{appsec:aphasia}.
The intact answers of all 54 lesions are fluent by this check, but 35 of the 53 undone answers of the own removal are not.
The check matters because the label alone overstates breakage: the classifier also labels the intact answers of 13 of the 54 lesions model scrambles, although all of them are fluent.
Random removals break the answers less often: 111 of the 151 random removals that undo a lesion give a fluent answer, and 34 give a fluent answer that the classifier labels control speech.
An undone lesion is thus not a healed one: no answer returns to the intact text, and removing the own direction mostly breaks the answer.

\paragraph{Few pushes restore fluent control speech.}
Of the 1,728 own pushes, 167, or 9.7\%, undo the lesion with a fluent answer that the classifier labels control speech, and such pushes exist in 38 of the 54 lesions.
Of the 5,184 random pushes, 271 do so, or 5.2\%, in 32 lesions.
Most pushes that undo a lesion end in model scrambles instead: 1,411 of the 1,643 own pushes and 4,290 of the 4,720 random pushes.
Stronger pushes less often undo the lesion with an answer that the classifier labels control speech: 22\% of the own and 15\% of the random pushes do so at $r = -0.5$, and none of the 432 own and 1,296 random pushes at $r = -4$.
Table~\ref{apptab:rescue-examples} shows two lesions that a weak own push undoes with fluent control speech, which we chose because their answers read well.
In both, the steered-back answer is a fluent answer to the task, but not the intact answer.
At the same layer and strength, all three random pushes also undo both lesions, and the classifier labels five of their six answers model scrambles.
Removing the own direction instead turns both lesioned answers into model scrambles, with a probability of 1.00 for that class.
The own direction thus restores control speech almost twice as often as a random direction, but for fewer than one push in ten.

\begin{table*}[t]
\centering
\small
\begin{tabular}{@{}>{\raggedright\arraybackslash}p{0.31\textwidth}>{\raggedright\arraybackslash}p{0.31\textwidth}>{\raggedright\arraybackslash}p{0.31\textwidth}@{}}
\toprule
Intact & Lesioned & Lesioned and steered back \\
\midrule
\multicolumn{3}{@{}l}{\gptxl, Broca target, lesion of 146 heads; own direction pushed after layer 24 at $r = -0.5$} \\
$\pbroca = 0.000$, control & $\pbroca = 0.884$, Broca & $\pbroca = 0.000$, control \\
\textit{I'd start with the bread, cut it into slices about 1/2 inch thick, then spread some of your favorite jam on each slice. Then add in some sliced peanuts (or other nuts) and top that off with a scoop of PB\&J. You can also use peanut butter instead of\ldots}
& \textit{I'd love to have some peanut butter, but not the kind of peanut butter. / I'm sure do like peanut butter. / But I don't know what it is. / It's just peanut butter. / And I can't tell you what it is. / Peanut Butter. / / That's right? / Yes. / No. / What are you talking\ldots}
& \textit{I'd use the same recipe as shown in this picture, but with only half of the peanut butter and half of the jelly. Then I'd put it on top side of the bread and then I'd spread it out to form a circle around the bread and I'd cut off about 1/4th of an\ldots} \\
\midrule
\multicolumn{3}{@{}l}{\llamaseven, aphasia target, lesion of 60 heads; own direction pushed after layer 8 at $r = -0.5$} \\
$\paph = 0.006$, control & $\paph = 0.931$, Broca & $\paph = 0.006$, control \\
\textit{Sure! Here's my step-by-step guide on how to make the perfect peanut butter and jelly sandwich: / / Ingredients: / / * 2 slices of bread (white or whole wheat, whichever you prefer) / * Peanut butter (creamy or crunchy, your choice) / * Jelly or jam (your\ldots}
& \textit{\textup{[emoji]}peanut Butt the jigglesand!}
& \textit{Sure! Here's one way to make a peanut butter: / / / / 1. Start by spreading a layer of bread on the bottom of your pan. You can use any type of bread you like, such as white or whole wheat. Just be sure it is flat enough for the filling. / 2. Next, add a\ldots} \\
\bottomrule
\end{tabular}
\caption{Two \ourmethod lesions on the sandwich task that a weak push with the lesion's own direction undoes.
Each row gives the answers of the intact model, of the lesioned model, and of the lesioned model steered back by the push, each under its target probability and the classifier's most probable class.
We chose the two lesions because their answers read well, so they are not typical: in 16 of the 54 lesions, no push with the own direction undoes the lesion with a fluent answer that the classifier labels control speech, where a fluent answer does not loop and has at least 80\% of its non-space characters in dictionary words.
Answers are verbatim up to the last whole word within their first 250 characters; \ldots{} marks the omitted rest, a slash marks a line break, and [emoji] replaces an emoji.}
\label{apptab:rescue-examples}
\end{table*}

\paragraph{Removing the own direction also breaks intact answers.}
Removing the own direction from the intact model changes the intact answer for all 54 lesions and makes 20 of these answers fail the fluency check.
In three settings, all of Qwen3.5 with the aphasia target, removing the aphasia direction even makes the intact model's answer reach the target.
The probability of aphasia rises from 0.001 to 0.548 for \qwen{9B} on the Cinderella task and from 0.006 to 0.848 on the sandwich task.
For \qwen{4B} on the sandwich task, it rises from 0.005 to 0.450, and conduction or anomic aphasia becomes the most probable class.
Removal thus damages the language of intact models too, so its undoing of the lesions does not show that the lesions act through the own direction.

\paragraph{Random directions also undo the lesions of the layer search.}
For comparison, we ran the same test on nine aphasia lesions of \gptxl and the Pythia models that the layer search of Appendix~\ref{appsec:lesion-methods} found.
These lesions have one to seven heads, and the \gptxl lesion also knocks out seven blocks of 256 MLP neurons.
Removing the own direction undoes all nine lesions, whereas at least one random removal undoes eight of them and all three random removals undo six.
In one lesion, of \pythia{1B} on the sandwich task, the removal of the own direction undoes the lesion but no random removal does: the three random removals leave a mean target probability of 0.92.
Over the nine lesions, removal leaves a mean target probability of 0.001 for the own direction and 0.244 for the random directions.
Random removals undo these lesions 21 of 27 times, and they undo the \ourmethod aphasia lesions of the same nine pairs of a model and a task about as often, 22 of 27 times.
In three of these pairs, both searches found the same single head.
Random directions thus undo most lesions of either search, so the weak specificity of steering back is not a property of \ourmethod alone.

\section{LLM Judge Prompts}
\label{appsec:judge}

One LLM judge, DeepSeek-V4.1-Flash, plays two roles in this paper.
It labels the steered answers that the experts also label, given only the names of the classes (\S\ref{sec:method-steer}), and it rates the answers of intact and lesioned models to the TAB's five connected-text prompts for 19 aphasia features (\S\ref{sec:method-lesion}).
This appendix gives the judge's settings, its three prompts verbatim, and the rules that turn its replies into labels.
In the listings, a gray arrow marks where a long line of a prompt wraps on the page.

\paragraph{Judge and settings.}
We query DeepSeek-V4.1-Flash~\citep{deepseekai2026deepseekv41flash} through DeepSeek's application programming interface (API), under the model name \texttt{deepseek-flash}, at temperature 0.
Each request holds one answer and nothing about how the answer was produced, such as the steering target or the lesion.
DeepSeek-V4.1-Flash is a reasoning model, and it is not deterministic at temperature 0: the same request can return a different reply.
We therefore send each distinct request once and keep its first reply.
We store every reply under the full content of its request, which is the model name, the messages, the temperature, and, for the TAB prompt, the reply format.
An analysis that sends a request again reads the stored reply instead of asking the judge, so all analyses use the same verdicts.
We parse only the final reply, not the reasoning that the API returns with it.

\paragraph{Prompts for the steered answers.}
The judge labels each answer steered toward aphasia with the two-way prompt of Listing~\ref{lst:judge-binary}, and each answer steered toward Broca's or Wernicke's aphasia with the five-way prompt of Listing~\ref{lst:judge-fiveway}.
Both prompts give the class names and nothing else: no definitions and no examples.
The classifier learned its classes from labeled examples, and a judge given hand-written definitions or examples would judge partly by those, so with the class names alone the judge relies on what it knows about aphasia.
In both prompts, \texttt{\{question\}} is the request that the model answered, for example \say{Tell me about your first job.}, and \texttt{\{answer\}} is the steered answer.
For the TAB's prompts, the question leaves out the battery's instruction \say{Respond to the following prompt in 3-5 full sentences:}, which the model received with the prompt.
The judge receives the filled prompt as the only message of the request, with no system message.

\par\medskip\noindent
\begin{minipage}{\columnwidth}
\begin{lstlisting}[basicstyle=\small\ttfamily, breaklines=true, columns=fullflexible, frame=single,
  framesep=3pt, xleftmargin=3.4pt, xrightmargin=3.4pt, breakatwhitespace=true, upquote=true,
  language={}, numbers=none, showstringspaces=false, literate={—}{{\textemdash}}1,
  postbreak=\mbox{\textcolor{gray}{$\hookrightarrow$}\space},
  caption={Two-way prompt of the LLM judge, for the answers steered toward aphasia. The judge receives it as a single user message, with the request that the model answered in place of \texttt{\{question\}} and the model's answer in place of \texttt{\{answer\}}.},
  label=lst:judge-binary]
Here is a question and a speaker's answer to it, transcribed.

Question: {question}
Answer: {answer}

Does the speaker have aphasia? Reply with one word: yes or no.
\end{lstlisting}
\end{minipage}
\par\medskip

\par\medskip\noindent
\begin{minipage}{\columnwidth}
\begin{lstlisting}[basicstyle=\small\ttfamily, breaklines=true, columns=fullflexible, frame=single,
  framesep=3pt, xleftmargin=3.4pt, xrightmargin=3.4pt, breakatwhitespace=true, upquote=true,
  language={}, numbers=none, showstringspaces=false, literate={—}{{\textemdash}}1,
  postbreak=\mbox{\textcolor{gray}{$\hookrightarrow$}\space},
  caption={Five-way prompt of the LLM judge, for the answers steered toward Broca's or Wernicke's aphasia, sent in the same way as the two-way prompt of Listing~\ref{lst:judge-binary}. The options are the classifier's five classes, and the last option stands for model scrambles.},
  label=lst:judge-fiveway]
Here is a question and a speaker's answer to it, transcribed.

Question: {question}
Answer: {answer}

Which one of these describes the speaker?
Control (no aphasia)
Anomic or conduction aphasia
Broca's aphasia
Wernicke's aphasia
Not human speech (garbled machine text)

Reply with exactly one of these five options.
\end{lstlisting}
\end{minipage}
\par\medskip

\paragraph{Prompt for the TAB's connected text.}
The TAB prompt follows the battery's protocol for scoring connected text automatically~\citep{roll2025tab}.
Its system message, in Listing~\ref{lst:judge-tab-system}, is the TAB paper's instruction to its judge, word for word.
Its user message, in Listing~\ref{lst:judge-tab-user}, defines the 19 features with an example of each, shows two transcripts annotated with all 19 features, and then gives the answer to rate.
There, \texttt{\{prompt\}} is one of the TAB's five connected-text prompts, again without the instruction on length, and \texttt{\{transcript\}} is the model's answer.
The request also asks the API for a reply in JavaScript Object Notation (JSON).
The features are the battery's 19, in its order, but we wrote the definitions, the examples, and the two annotated transcripts ourselves, from the feature names and from standard usage in aphasiology.
The prompt thus reconstructs the battery's protocol rather than copying it, and its definitions and examples differ from those that the TAB paper prints.
Both annotated transcripts are synthetic, so the judge never sees text from an AphasiaBank transcript.
The TAB's authors validated their protocol against expert raters with Gemini 2.5 Flash as the judge, whereas we did not compare the TAB ratings of DeepSeek-V4.1-Flash with human raters, so the TAB results of Table~\ref{tab:beyond} rest on the judge alone.

\paragraph{From replies to labels.}
A two-way reply gets the label of its first whole word \texttt{yes} or \texttt{no}, ignoring case: aphasic for \texttt{yes}, and not aphasic for \texttt{no}.
A five-way reply gets the class of the keyword that appears first in it, ignoring case: \texttt{control} for control speech, \texttt{anomic} or \texttt{conduction} for conduction or anomic aphasia, \texttt{broca} for Broca's aphasia, \texttt{wernicke} for Wernicke's aphasia, and \texttt{not human} or \texttt{garbled} for model scrambles.
A reply without any of these words gets no label.
The judge labeled 294 steered answers, 98 steered toward aphasia and 196 toward Broca's or Wernicke's aphasia, and all but 4 of its replies were exactly one of the options, up to capitalization.
For those 4 answers, 1 steered toward aphasia and 3 toward Broca's or Wernicke's aphasia, the judge's reasoning reached 65,536 tokens and its reply was empty.
Neither these 4 answers nor the 18 answers that were too short for the classifier are among the 272 answers of Table~\ref{tab:agreement}.
We read a TAB reply as a JSON object, from its first opening brace to its last closing brace, and give each feature its value in the object, 0 or 1; a feature that the object leaves out counts as absent.
Every reply behind Table~\ref{tab:beyond} was a JSON object with exactly the 19 features, each 0 or 1.
Table~\ref{tab:beyond} counts the features that a reply marks present, out of 19, so the count includes the feature of overall communication impairment.

\par\medskip\noindent
\begin{minipage}{\columnwidth}
\begin{lstlisting}[basicstyle=\small\ttfamily, breaklines=true, columns=fullflexible, frame=single,
  framesep=3pt, xleftmargin=3.4pt, xrightmargin=3.4pt, breakatwhitespace=true, upquote=true,
  language={}, numbers=none, showstringspaces=false, literate={—}{{\textemdash}}1,
  postbreak=\mbox{\textcolor{gray}{$\hookrightarrow$}\space},
  caption={System message of the TAB judge prompt, which is the instruction that the TAB paper gives its judge~\citep{roll2025tab}.},
  label=lst:judge-tab-system]
You are an automated system for analyzing transcripts of speech for aphasic features. Your task is to read a transcript passage and determine whether each of the following features is present (1) or not present (0). Your output must be a JSON file with each feature as a key and its value as either 0 or 1.
\end{lstlisting}
\end{minipage}
\par\medskip

\par\begingroup
\dimen0=15\baselineskip
\vskip 0pt plus \dimen0 \penalty -100 \vskip 0pt plus -\dimen0
\vskip \dimen0 \penalty 9999 \vskip -\dimen0 \vskip 0pt
\endgroup
\begin{lstlisting}[basicstyle=\small\ttfamily, breaklines=true, columns=fullflexible, frame=single,
  framesep=3pt, xleftmargin=3.4pt, xrightmargin=3.4pt, breakatwhitespace=true, upquote=true,
  language={}, numbers=none, showstringspaces=false, literate={—}{{\textemdash}}1,
  postbreak=\mbox{\textcolor{gray}{$\hookrightarrow$}\space},
  caption={User message of the TAB judge prompt, sent after the system message of Listing~\ref{lst:judge-tab-system} for each connected-text answer, with the TAB prompt in place of \texttt{\{prompt\}} and the model's answer in place of \texttt{\{transcript\}}. We wrote the definitions, the examples, and the two annotated transcripts, which differ from those of the TAB paper.},
  label=lst:judge-tab-user]
Features to score:
- anomia: Word-finding difficulty: pauses, circumlocutions, vague fillers ('thing', 'stuff'), or an explicit failure to retrieve a word. Example: "we went to the... the place with the, um, the water thing"
- abandoned_utterances: An utterance is started and given up before it is completed. Example: "and then we drove to the—. It was a nice day."
- empty_speech: Fluent output that carries little information: generic words and vague statements in place of content. Example: "It was good. We did things there. It was really something, you know."
- semantic_paraphasias: A real word substituted for the intended one from a related meaning (fork for spoon). Example: "I poured the milk into the fork"
- phonemic_paraphasias: Sound-level errors that produce a near-word: substituted, added, or transposed sounds. Example: "we watched the tevelision"
- neologisms: Invented non-words used as if they were words. Example: "she brought the flimber to the table"
- jargon: Fluent but largely incomprehensible speech made of neologisms and paraphasias. Example: "the plorbing came to the sandwitcher and it was gasping"
- perseverations: Inappropriate repetition of a word, phrase or idea already produced, carried over where it no longer fits. Example: "I like the beach. The beach is nice. My job was at the beach. The beach."
- stereotypies_and_automatisms: Recurring stock phrases or automatic sequences used instead of content ('you know', 'and so on', counting). Example: "you know, you know, and so on, you know"
- short_simplified_utterances: Telegraphic output: very short utterances with reduced syntax. Example: "Trip. Beach. Nice. Sun."
- omission_of_bound_morphemes: Missing inflections such as -ed, -s, -ing. Example: "yesterday I walk to the store and buy two apple"
- omission_of_function_words: Missing articles, prepositions, auxiliaries, or pronouns. Example: "went store. bought bread. came home."
- paragrammatism: Grammatical errors of commission in otherwise fluent speech: wrong tense, agreement, or word order. Example: "the trip were going by us to have had yesterday"
- false_starts: An utterance begins, stops, and restarts in a different form. Example: "My first— when I first— the job I had first was..."
- retracing: Words or phrases repeated in order to self-correct or to continue. Example: "we went to the, we went to the beach"
- conduite_d_approche: Successive approximations toward a target word. Example: "cat- catas- catastro- catastrophe"
- meaning_unclear: The intended message cannot be recovered from the utterance. Example: "it was the one for the when they had it over"
- off_topic: The response does not address the prompt. Example: (asked about a trip) "The stock market has been volatile recently."
- overall_communication_impairment: Global judgment: taken as a whole, communication is impaired. Example: (any transcript where the features above materially reduce communicative success)

Annotated example 1:
Prompt: Tell me about your first job.
Transcript: My first job was at a small bakery near my school. I started at five in the morning, helped shape the dough, and served customers once we opened. It was tiring but I loved the smell of fresh bread. I stayed there for two summers before going to college.
Labels: {"anomia": 0, "abandoned_utterances": 0, "empty_speech": 0, "semantic_paraphasias": 0, "phonemic_paraphasias": 0, "neologisms": 0, "jargon": 0, "perseverations": 0, "stereotypies_and_automatisms": 0, "short_simplified_utterances": 0, "omission_of_bound_morphemes": 0, "omission_of_function_words": 0, "paragrammatism": 0, "false_starts": 0, "retracing": 0, "conduite_d_approche": 0, "meaning_unclear": 0, "off_topic": 0, "overall_communication_impairment": 0}

Annotated example 2:
Prompt: Tell me about your first job.
Transcript: Job... um, first job. Bakery. Bakery, no, the— the bread place. Five o'clock, morning. Make the... the thing, the dough. Dough. People come, I give bread. Two summer. Tired but good. Bread smell good, good.
Labels: {"anomia": 1, "abandoned_utterances": 0, "empty_speech": 0, "semantic_paraphasias": 0, "phonemic_paraphasias": 0, "neologisms": 0, "jargon": 0, "perseverations": 1, "stereotypies_and_automatisms": 0, "short_simplified_utterances": 1, "omission_of_bound_morphemes": 1, "omission_of_function_words": 1, "paragrammatism": 0, "false_starts": 1, "retracing": 1, "conduite_d_approche": 0, "meaning_unclear": 0, "off_topic": 0, "overall_communication_impairment": 1}

Now score this transcript. Output only the JSON object with all 19 keys.
Prompt: {prompt}
Transcript: {transcript}
\end{lstlisting}



\end{document}